\documentclass[10pt,twocolumn,letterpaper]{article}

\usepackage[pagenumbers]{cvpr} 
\usepackage{adjustbox}
\usepackage{graphicx}
\usepackage{amsmath}
\usepackage{amssymb}
\usepackage{multirow}
\usepackage{booktabs}
\usepackage{soul}
\usepackage{comment}
\usepackage{rotating}
\usepackage{comment}

\usepackage[square,numbers,sort&compress]{natbib}
\usepackage[inline]{enumitem} 
\usepackage{algpseudocode}
\usepackage{array}
\usepackage{multirow}
\usepackage{booktabs}
\usepackage{algorithm}
\usepackage{subcaption}
\usepackage[normalem]{ulem}
\usepackage{xparse}
\usepackage{pifont}
\usepackage{bm}
\usepackage{listings}
\usepackage{mwe} \usepackage{makecell}
\usepackage{color, colortbl}

\newcommand{\ImNetDataset}{ImageNet\xspace}
\newcommand{\ImNet}{ImageNet-1K\xspace}
\newcommand{\ImNetFull}{ImageNet-22K\xspace}

\newcommand{\casualconv}{Casual Conversations\xspace}
\newcommand{\utk}{UTK-Faces\xspace}
\newcommand{\Places}{Places205\xspace}
\newcommand{\inat}{iNaturalist18\xspace}
\newcommand{\iwild}{iWildCam-WILDS\xspace}
\newcommand{\dollarstreet}{DollarStreet\xspace}
\newcommand{\hateful}{HatefulMemes\xspace}

\newcommand{\VOCseven}{VOC07\xspace}

\newcommand{\dsprite}{dSprites\xspace}
\newcommand{\copydays}{Copydays\xspace}

\newcommand{\regnet}{RegNet\xspace}
\newcommand{\vit}{ViT\xspace}

\newcommand{\multigrain}{Multigrain\xspace}
\newcommand{\supervised}{Supervised\xspace}
\newcommand{\seer}{SEER\xspace}
\newcommand{\swav}{SwAV\xspace}
\newcommand{\swavbigrn}{SwAV-RN50w5\xspace}
\newcommand{\swavbigregnet}{SwAV-RG128Gf\xspace}
\newcommand{\dino}{DINO\xspace}
\newcommand{\moco}{MoCov3\xspace}
\newcommand{\byol}{BYOL\xspace}
\newcommand{\byolbig}{BYOL-RN200w2\xspace}
\newcommand{\simclr}{SCLRv2\xspace}
\newcommand{\simclrbig}{SimCLRv2-RN152w3+SK\xspace}

\newcommand{\syncbn}{\texttt{SyncBatchNorm}\xspace}
\newcommand{\batchnorm}{\texttt{BatchNorm}\xspace}

\newlength\savewidth

\newlength\thinwidth\newcommand\thinline{\noalign{\global\savewidth\arrayrulewidth
  \global\arrayrulewidth 0.5pt}\hline\noalign{\global\arrayrulewidth\savewidth}}

\usepackage[pagebackref,breaklinks,colorlinks]{hyperref}

\usepackage[capitalize]{cleveref}
\crefname{section}{Sec.}{Secs.}
\Crefname{section}{Section}{Sections}
\Crefname{table}{Table}{Tables}
\crefname{table}{Tab.}{Tabs.}

\def\cvprPaperID{*****} \def\confName{CVPR}
\def\confYear{2022}

\begin{document}
\title{
Vision Models Are More Robust And Fair \\
When Pretrained On Uncurated Images Without Supervision\\
}

\author{
Priya Goyal$^1$
~~
Quentin Duval$^1$
~~
Isaac Seessel$^1$
~~
Mathilde Caron$^{1, 2}$
~~
Ishan Misra$^1$
~~
Levent Sagun$^1$ \\
~~
Armand Joulin$^1$
~~
Piotr Bojanowski$^1$
\\ 
\\
$^1$ Meta AI Research ~~~ $^2$ Inria
\\
\small \url{https://github.com/facebookresearch/vissl/tree/main/projects/SEER}
}

\twocolumn[{\renewcommand\twocolumn[1][]{#1}\maketitle
\begin{center}
  \centering
  \captionsetup{type=figure}
  \includegraphics[width=\linewidth]{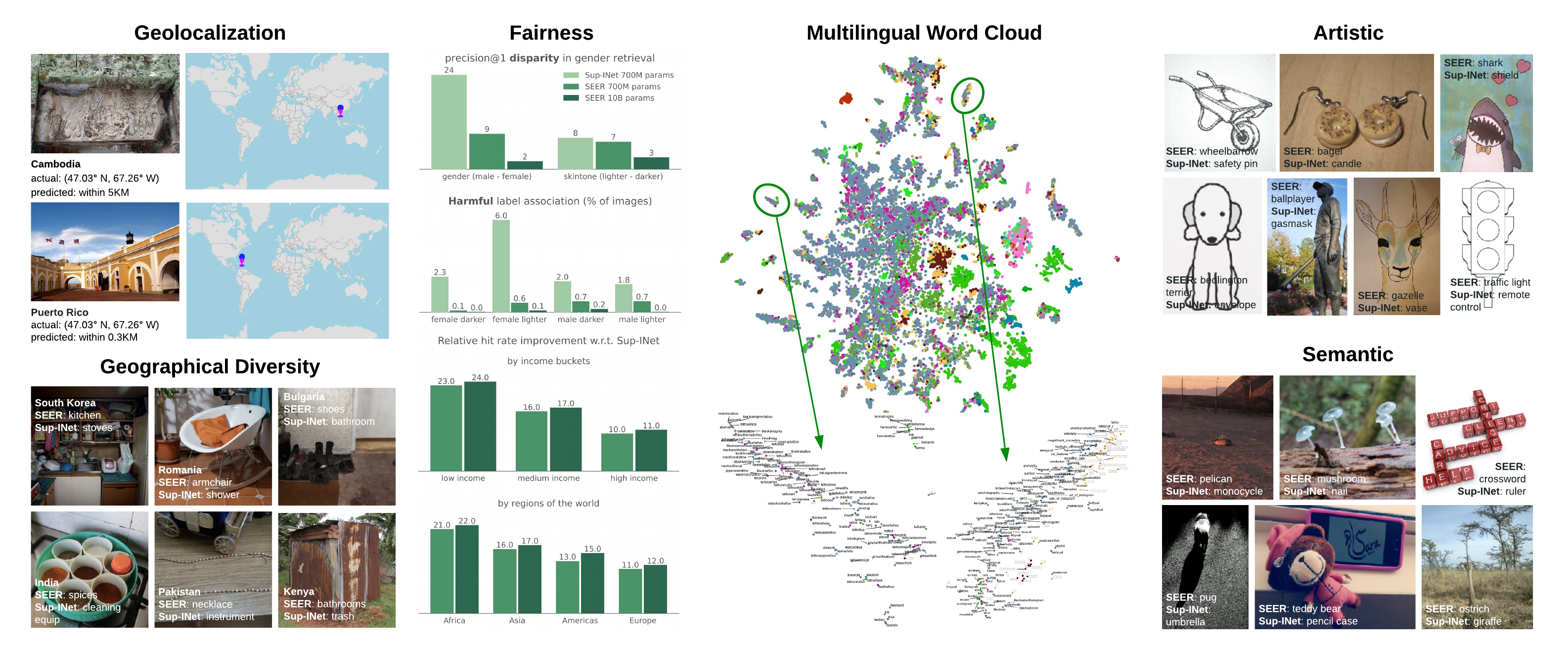}
  \captionof{figure}{
    Self-supervised training on diverse, real, and unfiltered internet data leads to interesting properties emerging like geolocalization, fairness, multilingual hashtag embeddings, artistic and better semantic information. See supplemental material for license information.
  }
  \label{fig:front_page}
\end{center}
}]

\begin{abstract}
Discriminative self-supervised learning allows training models on any random group of internet images, and possibly recover salient information that helps differentiate between the images.
Applied to ImageNet, this leads to object-centric features that perform on par with supervised features on most object-centric downstream tasks.
In this work, we question if using this ability, we can learn \texttt{any} salient and more representative information present in diverse unbounded set of images from across the globe.
To do so, we train models on billions of random images without any data pre-processing or prior assumptions about what we want the model to learn. 
We scale our model size to dense 10 billion parameters to avoid underfitting on a large data size. 
We extensively study and validate our model performance on over 50 benchmarks including fairness, robustness to distribution shift, geographical diversity, fine grained recognition, image copy detection and many image classification datasets.  
The resulting model, not only captures well semantic information, it also captures information about artistic style and learns salient information such as geolocations and multilingual word embeddings based on visual content only.
More importantly, we discover that 
such model is more robust, more fair, less harmful and less biased than supervised models or models trained on object-centric datasets such as ImageNet.
\end{abstract}

\section{Introduction}
In the span of a few years, self-supervised learning has surpassed supervised methods as a way to pretrain neural networks~\cite{chen2020simple, he2020momentum, caron2020unsupervised, grill2020bootstrap}.
At the core of this success lies discriminative approaches that learn by differentiate between images~\cite{dosovitskiy2014disc, wu2018unsupervised} or clusters of images~\cite{caron2018deep, asano2019self}.
Despite little assumptions made by these methods on the underlying factors of variations in the data, they produces features that are general enough to be re-used as they are in a variety of supervised tasks.
While this has been widely studied in the context of object-centric benchmarks, like ImageNet~\cite{russakovsky2015imagenet} or COCO~\cite{lin2014microsoft}, we conjecture that this property is more general and could allow to recover any factor of variation in a given distribution of images.
In other words, this property can be leveraged to \textit{``discover'' properties in uncurated datasets of images}. 

These properties that a self-supervised model may discover depend on the factors of variation contained in the training data~\cite{bouchacourt2021grounding}. 
For instance, learning features on object-centric dataset will produce features that have object-centric properties~\cite{caron2021emerging}, while training them in the wild may contain information that are related to people's general interests.
While some of these signals may be related to metadata -- e.g., hashtags, GPS coordinate --  or semantic information about scenes or objects, other factors may be related to human-centric properties -- e.g., fairness, artistic style -- that are harder to annotate automatically.
In this work, we are interested in probing which of the properties emerge in visual features trained with no supervision on as many images from across the world as possible.

A difficulty with training models on images in the wild is the absence of control on the distribution of images, e.g., the data likely has concepts that are dis-proportionally represented compared to others.
This means that an under-parameterized network may underfit and only learn the most predominant concepts.
For instance, studies~\cite{goyal2021self} show that even a $1$ billion parameter model saturates after $32$M images, and do not extract more information when trained on billion of images.
Even without these difficulties, learning the diversity of concepts in images from billions of people around the world requires significantly larger models than what is deployed for training on ImageNet scale. 

In this work, we question the limits of what can be learned on such data by further increasing the capacity of pretrained models to $10$billion dense parameters.
We address some of the engineering challenges and complexity of training at this scale and thoroughly evaluate the resulting model on in-domain problems as well as on out-of-domain benchmarks.
Unsurprisingly, the resulting network learn features that are superior to smaller models trained on the same data on standard benchmarks.
More interestingly though, on in-domain benchmarks, we observe that some properties of the features captured by the larger model was far less present in smaller model.
In particular, one of our key empirical findings is that \emph{self-supervised learning on random internet data leads to models that are more fair, less biased and less harmful}.
Second, we observe that our model is also able to leverage the diversity of concepts in the dataset to train \emph{more robust features}, leading to better out-of-distribution generalization.
We thoroughly study this finding on a variety of benchmarks to understand what may explain this property.

\section{Related Work}
\begin{figure*}[t]
  \centering
        \includegraphics[width=0.45\linewidth]{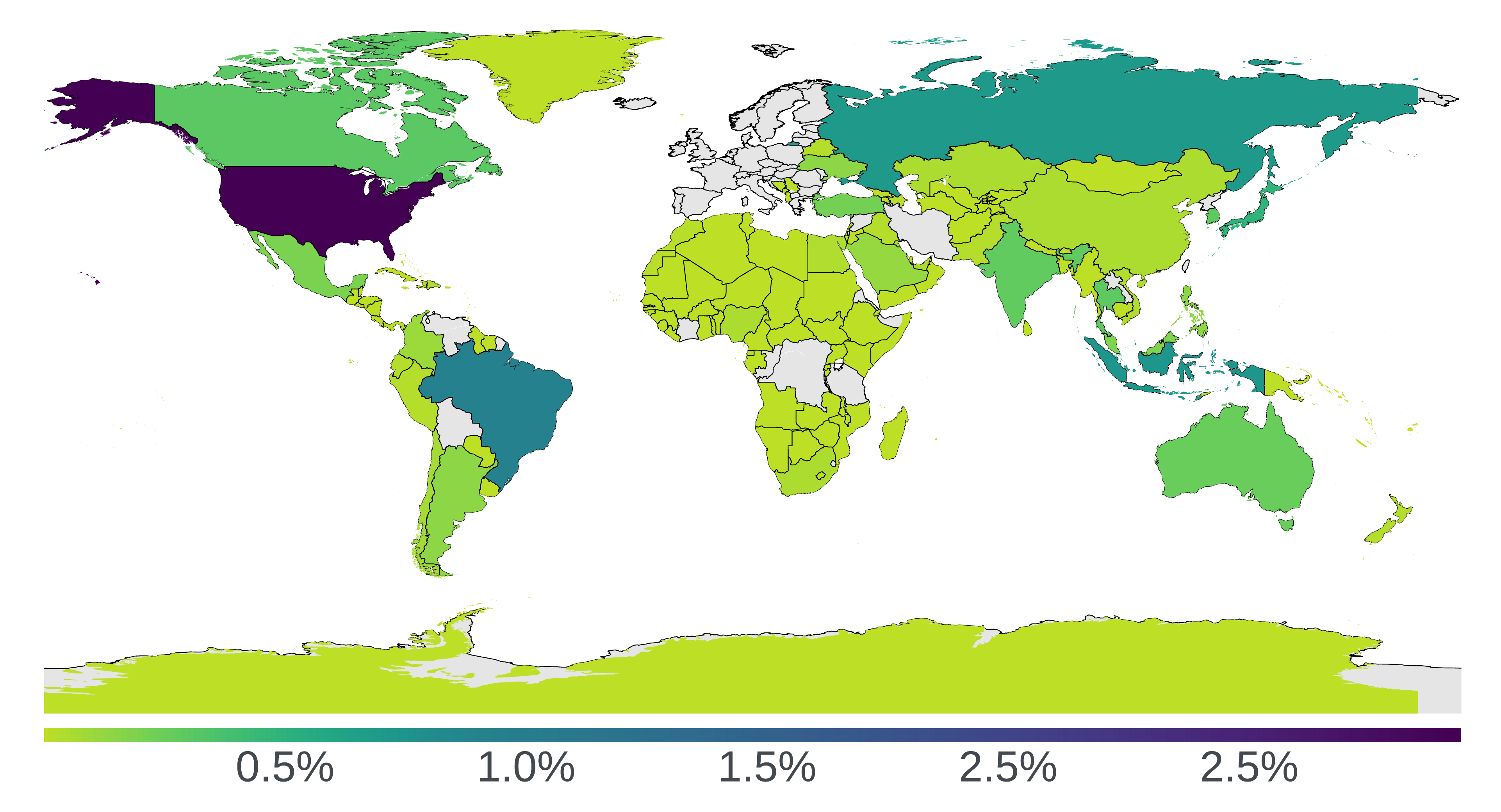}
       \hspace{0.5in}
  \includegraphics[width=0.45\linewidth]{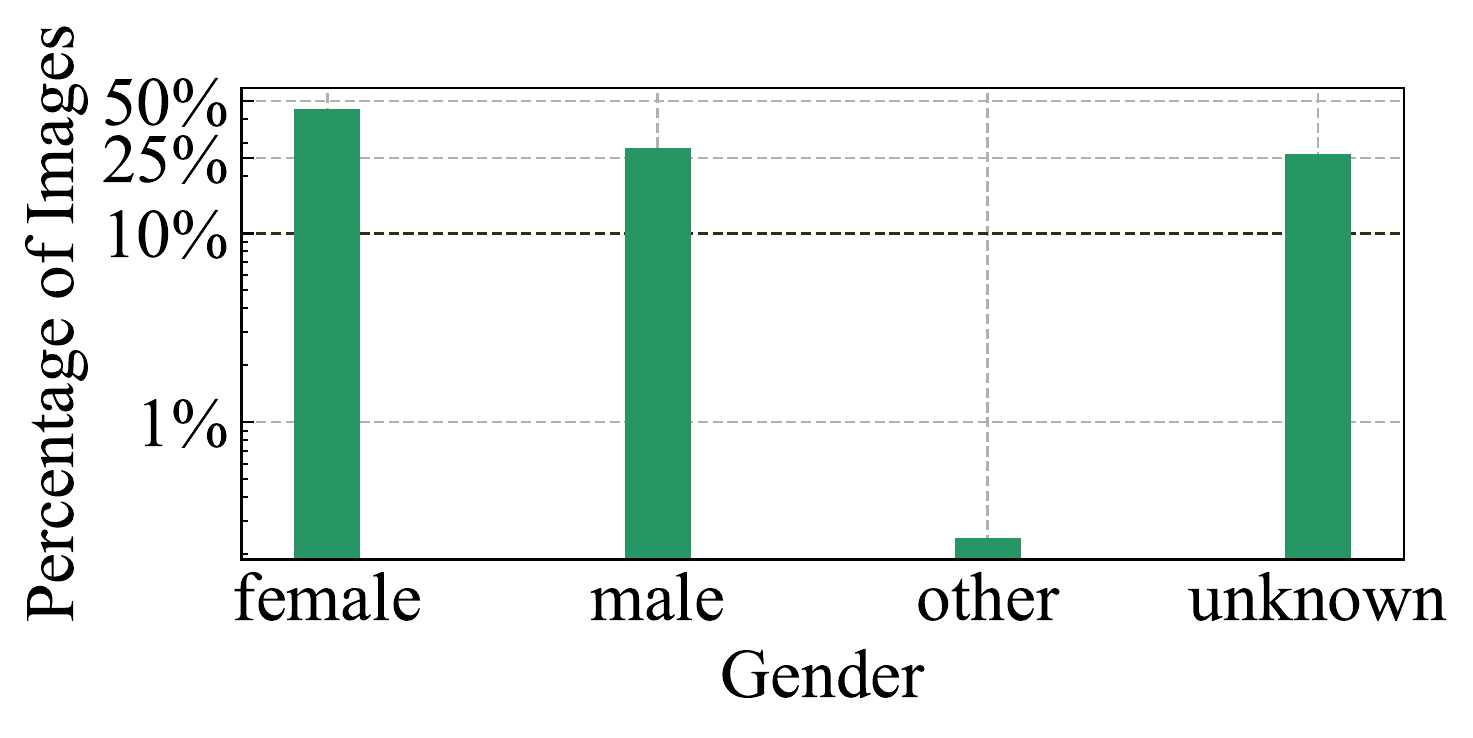}
  \caption{
  \label{fig:ig10m_data_distribution} \textbf{Geographical and Gender data distribution \textit{found} in SEER Pretraining Data}: we train our model on random group of a billion public Instagram images. We do \emph{not} perform any data sampling or curation to achieve a certain data distribution. We instead discover that the random group of images naturally represent the geographic and demographic diversity of the world.
    \textbf{(left):} Geographical Data distribution of images found in the pre-training data.
    \textbf{(right):} Gender distribution found in the same images.
Both distributions correspond to a random subset of $10M$ images from our $1$ billion pre-training dataset. Percentages (\%) denote fraction of $10$M images from each country and gender found in the dataset.
  }
\end{figure*}
\paragraph{Unsupervised Training of Visual Features.}
Unsupervised feature learning has a long history in computer vision, and many approaches have been explored in this space.
Initially, methods using a reconstruction loss have been explored with the use of autoencoders~\cite{ranzato2007unsupervised,vincent2008extracting}.
More recently, a similar paradigm has been used in the context of masked-patch-prediction models~\cite{bao2021beit,xie2021simmim,he2021masked}, showing that scalable pre-training can be achieved.
Alternatively, many creative pretext tasks have also been proposed, showing that good features can be trained that way~\cite{doersch2015unsupervised,agrawal2015learning,jenni2018self,kim2018learning,larsson2016learning,mahendran2018cross,misra2016shuffle,misra2020self,pathak2017learning,pathak2016context,wang2015unsupervised,wang2017transitive,zhang2017split}.
A popular trend was using instance discrimination~\cite{bojanowski2017unsupervised,chen2020simple,chen2020improved,dosovitskiy2016discriminative,hadsell2006dimensionality,he2020momentum,wu2018unsupervised} as a training task.
In this setup, each sample in the dataset is also it's own class.
Several other interesting papers proposed to learn joint embeddings, ``pulling together'' different views of the same image~\cite{grill2020bootstrap,bardes2021vicreg,zbontar2021barlow}.
Finally, a large body of work considered grouping instances and using clustering~\cite{asano2019self,caron2018deep,coates2011analysis,gidaris2020obow,huang2019unsupervised,junnan2021prototypical,xie2016unsupervised,yang2016joint,zhuang2019local} or soft versions thereof~\cite{caron2020unsupervised,caron2021emerging} as training tasks.
Many of those works have shown excellent performance on numerous downstream tasks, often showing that unsupervised features can surpass supervised ones.
In this paper, we use the model proposed by Caron \etal~\cite{caron2020unsupervised}, using soft assignments of images to prototypes.

\paragraph{Uncurated Data.}
Most works on unsupervised learning of features learn the models on supervised datasets like ImageNet~\cite{russakovsky2015imagenet}.
Some previous works have explored unsupervised training on images~\cite{caron2019unsupervised,doersch2015unsupervised,goyal2019scaling} and videos~\cite{miech2020end} taken ``in the wild''.
The conclusions of these works were mixed but these studies were conducted at a relatively small scale, both in model and data size.
There are now evidences that self-supervised pretraining benefits greatly from large models~\cite{caron2020unsupervised,chen2020big, henaff2019data,goyal2021self}.
Our work builds upon these findings to explore if we can learn good visual representations by training significantly larger models on random, uncurated and unlabeled images.

\paragraph{Scaling Architectures.}
Many works have shown the benefits of training large models on the quality of the resulting features~\cite{radosavovic2020designing,tan2019efficientnet,xie2017aggregated}. 
Training large models is especially important when pretraining on a large dataset, where a model with limited capacity will underfit~\cite{mahajan2018exploring}.  
This becomes even more important when training with unsupervised learning.
In that case, the network has to learn features that capture many aspects of the data, without being guided by a narrow output space defined by the manual annotation. 
To scale architecture size, all combinations of increasing width and depth have been explored in the self-supervised learning literature.
Kolesnikov \etal~\cite{kolesnikov2019revisiting} demonstrated the importance of wider networks for learning  high-quality visual features with self-supervision, 
Further, Chen \etal~\cite{chen2020big} achieved impressive performance with deeper and wider configurations. 
The largest models trained for each algorithm vary a lot, with architecture that are both deeper and wider, such as ResNet-50-w5, ResNet-200-w2 or ResNet-152-w3.
More generally, a large body of work is dedicated to building efficient models with large capacity~\cite{bello2021revisiting,tan2019efficientnet,touvron2020fixing,xie2017aggregated}.
Of particular interest, the RegNet model family~\cite{radosavovic2020designing} achieves competitive performance on standard image benchmarks.
while offering an efficient runtime and memory usage making them a good candidate for training at scale. 
In our work, we build up on the existing work and explore all $4$ dimensions (depth, width, input resolution, compound) of scaling the \regnet architecture in Sec.~\ref{sec:approach_scaling_architecture}.

\paragraph{Large-Scale Benchmarking of Computer Vision Models.}
Training high-quality visual representations that work well on a wide range of downstream tasks has been a core interest in the computer vision community. 
Recent advances in self-supervised learning~\cite{caron2020unsupervised,chen2020simple,he2020momentum,goyal2019scaling,goyal2021self} have shown that high quality visual features can be trained without labels.
They surpass the performance of supervised learning on many computer vision tasks including object detection, image classification and low-shot learning. 

The most widely used evaluation, initially proposed by Zhang \etal~\cite{zhang2016colorful}, consists in training linear classifiers on top of frozen features on ImageNet.
While widely adopted, this evaluation has been criticized for being somewhat artificial.
A finer study has proposed by Sariyildiz \etal, probing the performance of models when transferring to more distant concepts in ImageNet-22k~\cite{sariyildiz2021concept}.
Many recent works, following Chen \etal~\cite{chen2020simple} demonstrate performance on other image classification datasets such as Oxford Flowers~\cite{flowers2008}, Oxford Pets~\cite{parkhi12a}, MNIST~\cite{lecun1998gradient} or CIFAR~\cite{Krizhevsky2009LearningML}.
These benchmarks are saturated with near perfect accuracy, and hence offer limited insight about the quality of a method.

Several works~\cite{zhai2020largescale,radford2021learning,kolesnikov2019big} proposed a collection of more than $30$ datasets to measure the generalization of weakly / fully-supervised models~\cite{mahajan2018exploring,yan2020clusterfit,dosovitskiy2021image}. 
Our work builds up on these studies and aims at validating the generalization of our self-supervised trained model on a large set of evaluation tasks.
To this end we use more than $50$ computer vision tasks that allow to capture the model's performance on various applications of computer vision.
We argue that measuring model generalization on out-of-domain tasks is important as models can be used ``off-the-self'' for applications that are hard to anticipate.

\paragraph{Fairness of computer vision models.} Several concerns have surfaced around the societal impact of computer vision models~\cite{dentongebru2020}, to name a few:
mis-classification of people’s membership in social groups (e.g., gender)~\cite{pinar2021,keyes2018}, computer vision systems that reinforce harmful stereotypes~\cite{doi:10.1177/2378023120967171,bhargava2019exposing} and the gender biases towards darker-skinned people~\cite{buolamwini2018gender}. 
Further, studies~\cite{yang2020towards} show that training on ImageNet might lead to potential biases and harms in models, that are then transferred to the downstream tasks that model is applied on. 
Dulhanty and Wong~\cite{dulhanty2019auditing} studied the demographics on ImageNet, showing that males aged 15 to 29 make up the largest subgroup. 
Stock and Cisse~\cite{stock2018convnets} have shown that models trained on ImageNet exhibit mis-classifications consistent with racial stereotypes. 
De Vreis \etal~\cite{de2019does} showed that the ImageNet trained models lack geographical fairness/diversity and work poorly on images from non-Western countries. 
Recently, effort has been made by Yang~\etal~\cite{yang2020towards} to reduce these biases by removing 2,702 \texttt{synsets} (out of 2,800 total) from the \texttt{person subtree} used in ImageNet. 
Motivated by the importance of building socially responsible models, we follow recent works~\cite{goyal2022fairness} to systematically study the fairness, harms and biases of our models trained using self-supervised learning on random group of internet images.

\section{Approach}

\begin{figure}[t]
        \centering
\includegraphics[width=0.8\linewidth]{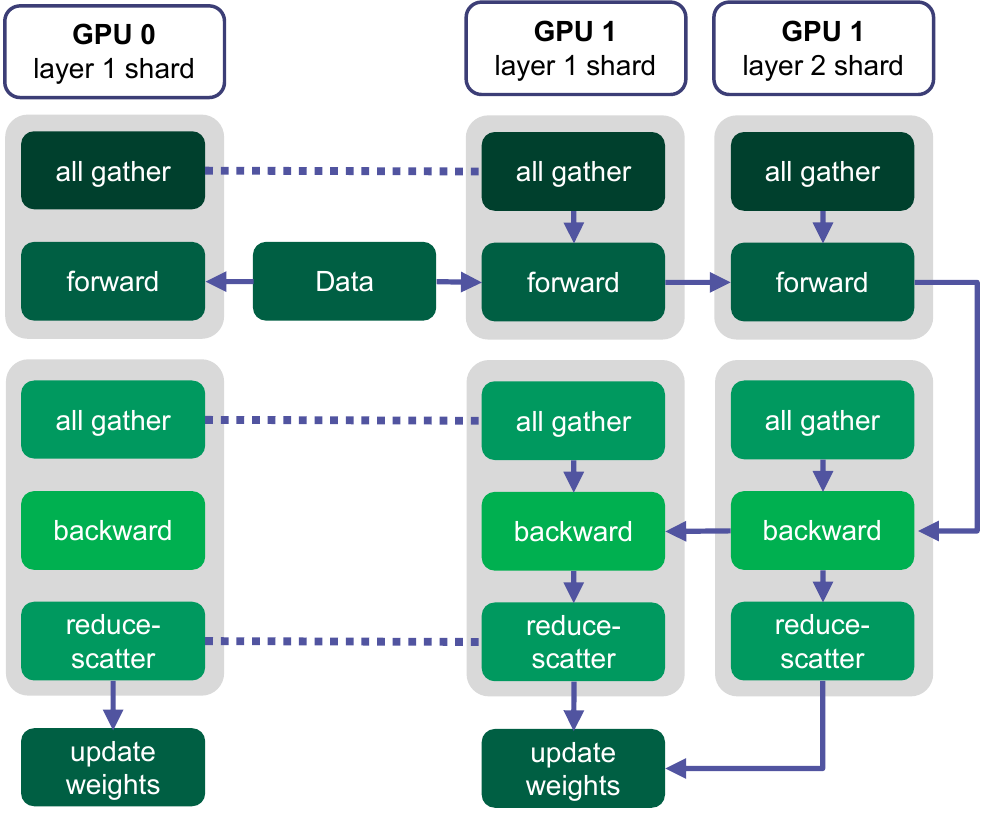}
        
        \caption{
          \textbf{Overview of FSDP data-parallel training}. Each model layer is sharded across all data-parallel workers. Dotted lines represent synchronisation points between GPUs and arrows represent dependencies. We overlap computation and communication to \textit{improve the efficiency} of \texttt{FSDP} training by, for example, scheduling all-gathers and forwards on different CUDA streams.
        }
        \label{fig:fsdp_diagram_and_activation_checkpoint} 
\end{figure}

\subsection{Self-supervised objective}
We train our model using SwAV~\cite{caron2020unsupervised}, and provide a short description of this algorithm here. Given two data augmentations of an image, that we refer to as $s$ and $t$, we compute their \emph{codes} $\mathbf{q}_s$ and $\mathbf{q}_t$.
SwAV trains a network by learning to predict the codes from the other view by minimizing the following loss function:
\begin{equation}
  \ell(\mathbf{z}_t, \mathbf{q}_s)
  + \ell(\mathbf{z}_s, \mathbf{q}_t),
\end{equation}
where $\mathbf{z}_s$ and $\mathbf{z}_t$ are the outputs of the network for augmentations $s$ and $t$.
The codes are typically predicted using a linear model $C$, and the loss $\ell$ then takes the following form:
\begin{equation}
  \ell(\mathbf{z}, \mathbf{q}) = - \sum_k \mathbf{q}^{(k)} \log \frac{\exp \left ( \frac{1}{\tau} \mathbf{z}^\top \mathbf{c}_k \right ) }{\sum_{k'} \exp \left ( \frac{1}{\tau} \mathbf{z}^\top \mathbf{c}_{k'} \right ) },
\end{equation}
where the $\mathbf{c}_k$ are \emph{prototypes}.
We obtain the codes by matching the features against prototypes using the Sinkhorn algorithm.
We defer the reader to~\cite{caron2020unsupervised} for more details.
The objective function can be minimized with stochastic gradient descent methods.

\subsection{Pre-training Data}
In this work, we are interested in training high-quality visual representations on a large collection of random, unfiltered, unlabeled internet images. 
To this end, we train our models on a subset of \textit{randomly} selected $1$ billion public and non-EU (to conform to GDPR) Instagram (IG) images. 
We do not apply any other pre-filtering and also do not curate the data distribution. 
Our dataset is unfiltered but we monitor the resulting geographical and gender distribution on a subset of randomly selected $10$M images in Fig.~\ref{fig:ig10m_data_distribution}.
As shown on the left panel, we find ~$192$ different countries represented in our pre-training data. 
Similarly, we observe that our data represents images from various genders, as shown on the right panel. 
We also quantitatively measure the fairness of our model in Sec.~\ref{sec:fairness_section}.

\begin{figure}[t]
\centering
\begin{tabular}{c@{}c}
        \includegraphics[width=0.48\linewidth]{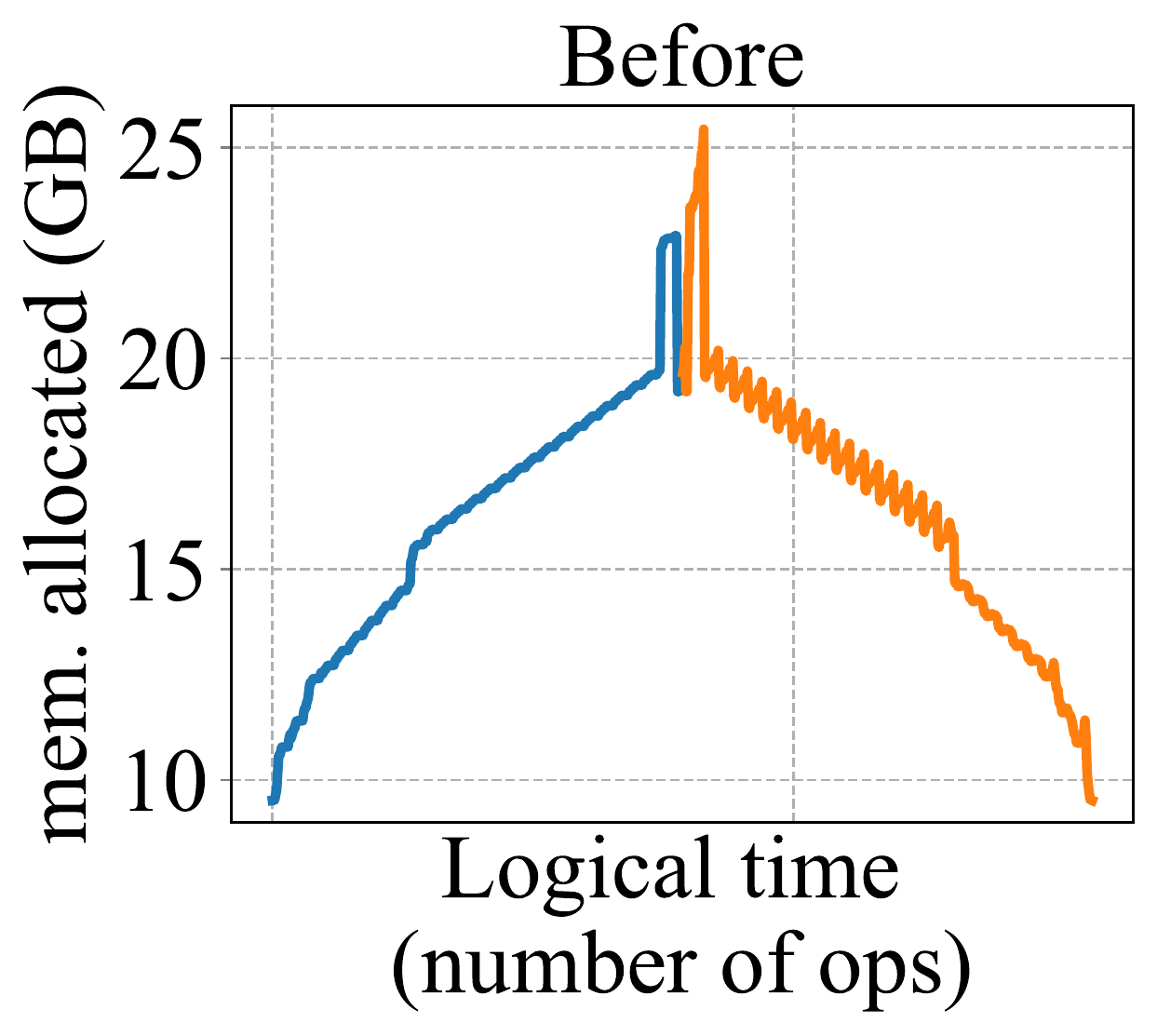}&
        \includegraphics[width=0.48\linewidth]{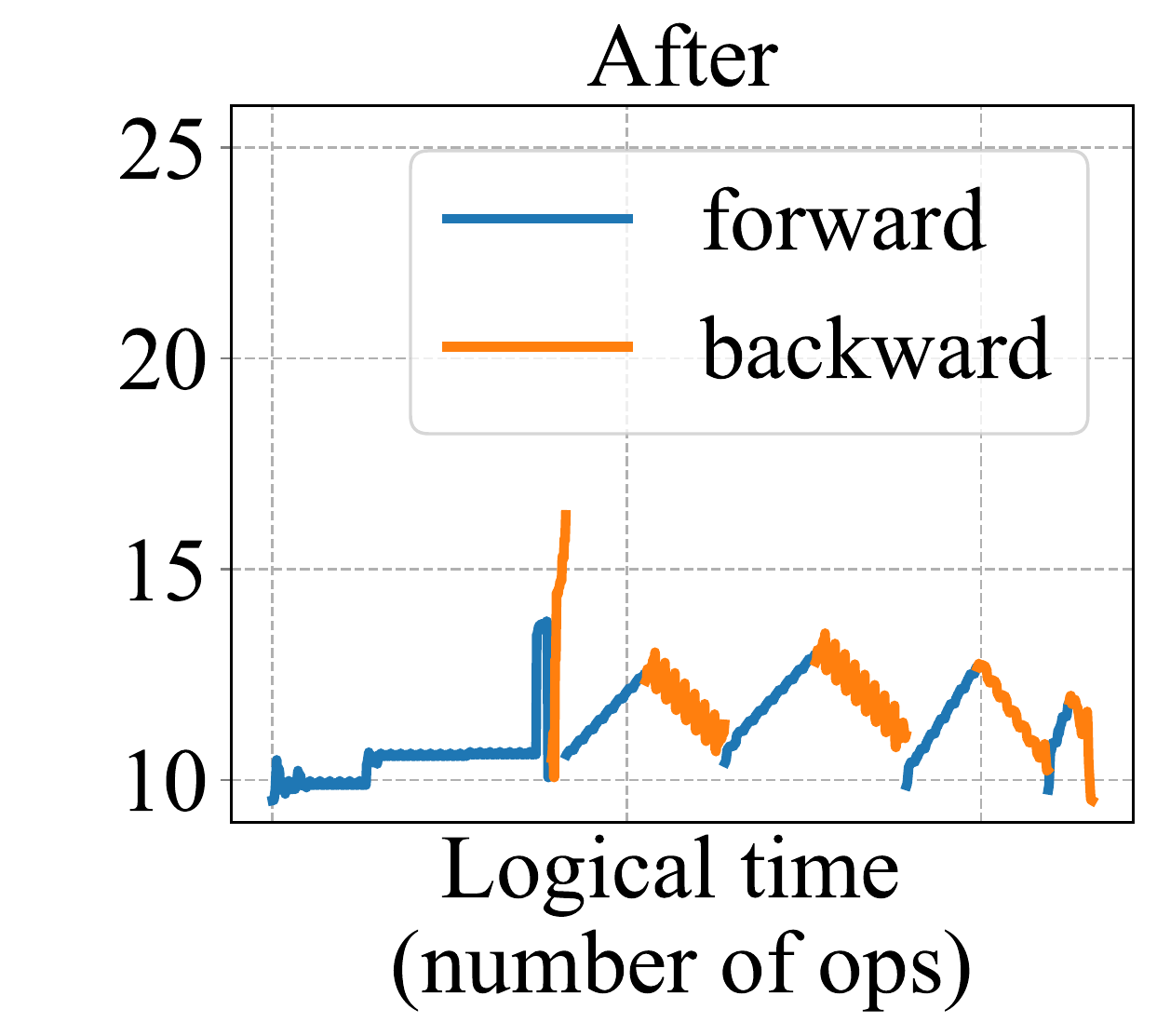}
\end{tabular}
        \caption{
          \textbf{Impact of activation checkpointing.} Memory profile of the 10B model on $8$ GPUs and $8$ images as input before and after inserting dynamic activation checkpointing.
The peak memory usage is reduced from $26$GB to $16$GB. The computation time increases by 15\% for same number of images but allows to increase batch size, hence increasing computational efficiency.
        }
        \label{fig:fsdp_diagram_and_activation_checkpoint2} 
\end{figure}

\subsection{Scaling the model architecture}

\paragraph{Scaling Axes.}
Self-supervised learning requires no annotations/labels for training models which means we can train large models from trillions of images at internet scale. Following previous works~\cite{goyal2021self,caron2020unsupervised} which demonstrated the possibility to train high-quality visual features from billions of internet images using self-supervised learning, we consider $3$ axes of scaling: 1) data size, 2) model size, and 3) data and model size. 

In our work, we are interested in scaling along second axis i.e. model size first. The reasoning behind this choice is two-folds: a) training on large data requires large enough model in order to take advantage of the data scale and discover properties present in the dataset, and b) model size appears to be a strong lever for low-shot learning ~\cite{goyal2021self} and we are interested in pushing these limits further.

\paragraph{Choosing and Scaling Model Architecture.} 
\label{sec:approach_scaling_architecture}
Towards our goal of scaling model size and pushing the limits further in self-supervised learning, we target training a \textbf{10B parameters dense model},  which, to the best of our knowledge, is the largest \textit{\textbf{dense}}\footnote{where every input is processed by every parameter, as defined in ~\cite{riquelme2021scaling}} computer vision model (contrary to the model in ~\cite{riquelme2021scaling} which is a ``sparse'' model). Following studies~\cite{goyal2021self}, we explore \regnet~\cite{radosavovic2020designing} (a ConvNet) architecture which has demonstrated promising model size scaling without any signs of saturation in performance. Further, since the largest model defined in \regnet family is a $1.5$B parameters model, we explore several strategies to increasing the architecture size to $10$B parameters. 

To increase the model size, we explore four dimensions: width, depth, resolution and compound scaling for the \regnet model family. Additionally, we explore a variant \regnet-Z~\cite{dollar2021fast} of this model family. Appendix Table~\ref{tab:model_scaling_variations} summarizes the variants. We trained each variant on $100M$ images using the same experimental setup and for each variant training, we evaluated model performance on the downstream task of linear classification on \ImNet. Our observations are as follows: \textit{(i)} the less wider but deeper models didn't change model performance on downstream task compared to the base model. However, such models lead to faster training time, \textit{(ii)} high input resolution models increase model runtime without increasing model parameters and yielded only modest increases in accuracy, \textit{(iii)} wider and deeper model with more FLOPs (than base model) improved performance on downstream task, and \textit{(iv)} RegNet-Z model architecture are more intensive and not efficient for scaling parameters.

Following these findings, we decided to keep the resolution fixed and increase the width (and/or depth) of the base model to scale to $10$ billion parameters model. We note that for better training speed, we ultimately kept the depth same and increased the width. Our full model details are described in Appendix~\ref{sec:appendix_models_seer}.

\paragraph{Fully Sharded Data Parallel.}
We train our models using PyTorch on NVIDIA A100 GPUs and our biggest model with 10B parameters requires 40GB of GPU memory (with additional 40GB required for optimizer state during pre-training). On a single V100\_32G GPU or more recent 40GB A100, such a model can not fit and hence the DDP (Distributed Data Parallel) training can not be used. We instead resort to model sharding and use the Fully Sharded Data Parallel \texttt{FSDP}~\cite{rajbhandari2019zero,xu2020automatic} training which shards the model such that each layer of the model is sharded across different data parallel workers (GPUs). The computation of minibatch is still local to each GPU worker. \texttt{FSDP} decomposes the \texttt{all-reduce} operations in DDP into separate \texttt{reduce-scatter} and \texttt{all-gather} operations. During the \texttt{reduce-scatter} phase, the gradients are summed in equal blocks among ranks on each GPU based on their rank index. During the \texttt{all-gather} phase, the sharded portion of aggregated gradients available on each GPU are made available to all GPUs. During the forward pass, the parameters of the layer to be computed are temporarily assembled before they are re-sharded. For training efficiency, the communication and computation are overlapped: un-sharding the next layer parameters (via \texttt{all-gather}) while computing the current layer. We illustrate the communication and compute optimizations in Fig.~\ref{fig:fsdp_diagram_and_activation_checkpoint}. For our model training, we leverage the \texttt{FSDP} implementation from Fairscale \footnote{\url{https://github.com/facebookresearch/fairscale}} and adapt it for our model.

\paragraph{Activation Checkpointing Automation.}
In our model trainings, we use Activation Checkpointing~\cite{activationchen2016training} which is the technique of trading compute for memory. It works by discarding all model activations during the \texttt{forward} pass except the layers that have been configured to be ``checkpointed''. During the \texttt{backward} pass (backpropagation), the \texttt{forward} pass on a part of the model (between two checkpointing layers) is re-computed. While this technique can help increase the batch size (leading to more compute to be overlapped with communication which leads to more efficient training), one downside is that manual configuration / tuning of which layers should be ``checkpointed'' is needed. This can be time-consuming and often hard to find the optimal checkpointing state. Further, for the models that are hard to fit in memory, it can become very difficult to perform manual tuning. 

To address this, we implemented a Dynamic Programming algorithm\footnote{We implemented it in open sourced library \url{https://github.com/facebookresearch/vissl}.} to find the best checkpoint positions for a given model rather than manual tuning. The algorithm is as follows: \textit{(Step 1)} we first collect the amount of activation's memory allocation produced at each layer using automatic tooling, in an array $m$, \textit{(Step 2)} we optimally split with dynamic programming this array in consecutive sub-arrays delimited by K points $p_i$ where $0 \le i \le K$ and $p_i \le p_{i+1}$ such that:  \begin{equation}
    \underset{p_i}{\operatorname{argmin}} \max \sum_{j=p_{i-1}}^{p_i} m_j
\end{equation} \textit{(Step 3)} the points $p_i$ such that $0 < i < K$ are our activation checkpoints points, minimizing the maximum amount of cumulative activation memory for $K-2$ activation checkpoints, and \textit{(Step 4)} we iterate this algorithm increasing $K$ until we manage to fit our desired batch size on the GPU memory.

In practice, when applied to our $10$billion parameter model, the algorithm selected $4$ activation checkpoints locations. We further adapted the checkpoint positions for any further trade-offs not accounted in the algorithm. The impact on memory reduction is shown in Figure~\ref{fig:fsdp_diagram_and_activation_checkpoint2}.

\paragraph{Optimizing training speed.}
To optimize the training speed of the model, we use several optimizations. We use mixed-precision for training and perform the \texttt{forward} pass computations in \texttt{FP16}. Since computation happens in \texttt{FP16}, for the un-sharding of parameters via \texttt{all-gather} operation (which performs communication of parameters over the network), we exchange \texttt{FP16} weights instead of \texttt{FP32}. This speeds-up the training by communicating model parameters faster. We note that for certain special layers such as \syncbn, we still use \texttt{FP32} as otherwise the training becomes unstable. Further, we use LARC optimizer~\cite{you2017large} from NVIDIA Apex library\footnote{\url{https://github.com/NVIDIA/apex}} for large batch size training. Since the model parameters are sharded, we adapted the LARC implementation to compute the distributed norms of parameters but without \texttt{all-gather} of model weights. We share more details on this in Appendix~\ref{sec:appendix_larc_adaptation}. Additionally, we add the activation checkpointing in the order \texttt{FSDP(checkpointing(model layer))} instead of the other way around. This is because activation checkpointing re-computes the \texttt{forward} pass on part of the model during back-propagation and doing a \texttt{forward} pass on \texttt{FSDP} wrapped layer requires ``un-sharding'' of layer which involves communication of weights across all GPUs. Hence, \texttt{FSDP(checkpointing(model layer))} ensures that we do not trigger excessive ``un-sharding'' / communication cost across GPUs.

\subsection{Pretraining the SEER model}
We use open source VISSL library~\cite{goyal2021vissl} for our model training and implement \texttt{FSDP} and activation checkpointing integration for \regnet-Y model architecture. We generate a wider RegNetY-$10$B parameters architecture with the configuration: \texttt{w\_0} = $1744$, \texttt{w\_a} = $620.83$, \texttt{w\_m} = $2.52$, \texttt{depth} = ($2$, $17$, $7$, $1$), \texttt{group\_width} = $1010$. We use a $3$-layer multi-layer perceptron (MLP) projection head of dimensions $20280\times8192$, $8192\times8192$ and $8192\times256$. We do \textit{not} use \texttt{BatchNorm} layers in the head. We use \syncbn in the model trunk and synchronize \batchnorm stats globally across all GPU workers. Following ~\cite{goyal2021self}, we use \swav algorithm with same data augmentations and $6$ crops per image of resolutions $2\times160+4\times96$\footnote{We use lower resolution $160$ instead of $224$ for the bigger crop for better training speed. Our experiments (for smaller model sizes) yielded marginal difference in performance on downstream task between the two crop sizes}. For the SwAV objective, we use $16,000$ prototypes, temperature $\tau$ set to 0.1, sinkhorn regularization parameter (epsilon) to $0.03$ and perform $10$ iterations of sinkhorn algorithm. We train our model with stochastic gradient descent (SGD) momentum of $0.9$ using a large batch size of $7,936$ different images distributed over $496$ NVIDIA A100 GPUs results in $16$ different images per GPU. We use a weight decay of $1e-5$, LARS optimizer~\cite{you2017large}, activation checkpointing~\cite{activationchen2016training} and \texttt{FSDP} for training the model. We use learning rate warmup~\cite{goyal2017accurate} and linearly ramp up learning rate from $0.15$ to $9.3$ for the first $5,500$ iterations. After warmup, we use cosine learning rate schedule and decay the learning rate to final value $0.0093$. We train on $1$ billion images in total leading to $126$K training iterations. We share details about other smaller variants of SEER model in Appendix Table~\ref{tab:seer_model_variants}.

\paragraph{Reliable model training and evaluations.} To pre-train the large dense $10$Billion parameters dense model, pre-training reliability is crucial. Further, whereas we pretrain the model on $496$ GPUs using \texttt{FSDP} model sharding, we want to use and evaluate the model on \textit{many} downstream tasks but using much fewer GPUs (e.g. $8$ GPUs). We implemented an efficient model state dictionary checkpointing technique that helps us achieve reliable pre-training on $496$ GPUs and scalable model evaluations on $8$ GPUs. We discuss more details on this in Appendix~\ref{sec:appendix_model_state_dict}.

\section{Experiments}
We extensively validate the performance of our model on over 50 benchmarks tasks. In Sec.~\ref{sec:fairness_section}, we evaluate and compare the performance of our model on $4$ different fairness benchmarks including $3$ fairness indicators. In Sec.~\ref{sec:transfer_learning_section}, we further study the performance on many downstream tasks in computer vision including out-of-domain robustness in Sec.~\ref{sec:ood_section}, fine-grained image recognition in Sec.~\ref{sec:inat_section}, image copy detection in Sec.~\ref{sec:copy_detection_results} and finally test the feature representation quality via linear probe on over $25$ computer vision datasets in Sec.~\ref{sec:summarize_representation_learning_main}.

\begin{table*}[t]
  \centering

  \begin{adjustbox}{max width=\textwidth}
  \begin{tabular}{@{}lll c cc c cc c cccc c cccc@{}}
  \toprule
  
  &&& & \multicolumn{2}{c}{\textbf{Gender}} && \multicolumn{2}{c}{\textbf{Skintone}} && \multicolumn{4}{c}{\textbf{Gender Skintone}} &&  \multicolumn{4}{c}{\textbf{Age Groups}} \\
  \cmidrule{5-6} \cmidrule{8-9} \cmidrule{11-14} \cmidrule{16-19} 
    \textbf{Model} & \textbf{Data} & \textbf{Arch.}  && \textbf{female} & \textbf{male} && \textbf{darker} & \textbf{lighter} && \makecell{\textbf{female} \\\textbf{darker}} & \makecell{\textbf{female}\\\textbf{lighter}} & \makecell{\textbf{male}\\\textbf{darker}} & \makecell{\textbf{male}\\\textbf{lighter}} && \makecell{\textbf{18-30}} & \makecell{\textbf{30-45}} & \makecell{\textbf{45-70}} & \makecell{\textbf{70+}} \\
    
    \midrule 
    \multicolumn{19}{@{}l}{\textit{Supervised pretraining on ImageNet}} \\
    \supervised  & INet-1K & RG-128Gf && 67.5 & 91.8 && 73.6 & 82.1 && 58.2 & 75.1 & 92.7 & 91.1 && 78.5 & 76.7 & 80.1 & 75.8 \\

    \midrule 
    \multicolumn{19}{@{}l}{\textit{Self-supervised pretraining on ImageNet}} \\
    \swav       & INet-1K & RG-128Gf  && 62.1 & 93.0 && 69.7 & 80.8 && 50.3 & 71.6 & 93.7 & 92.5 && 76.6 & 74.6 & 76.7 & 69.4 \\
    
    \midrule
    \multicolumn{19}{@{}l}{\textit{Pretrained on random internet images}} \\
        
        \seer(ours)        & IG-1B & RG-128Gf  && 86.7 &  \textbf{96.1} &&  86.8 &  94.2 && 78.2 &  93.7 &  \textbf{97.5} & 94.9 && 89.6 &  90.5 &  92.6 &  88.7 \\

    \seer(ours)        & IG-1B & RG-10B  && \textbf{93.9} & \underline{95.8}  && \textbf{92.9}  & \textbf{96.2}  && \textbf{90.3} & \textbf{96.8} & \underline{96.1}  & \textbf{95.4} && \textbf{93.2}   & \textbf{95.0}  & \textbf{95.6}  & \textbf{96.7} \\
    
  \bottomrule
  \end{tabular}
  \end{adjustbox}
    \caption{
    Fairness Indicator1 result \textbf{\texttt{Precision@1} metric for \underline{Gender Retrieval} for different \textit{gender}, \textit{skintone} and \textit{age groups}} of several models on the \casualconv Dataset as described in Sec.~\ref{sec:indicator1_description}. This benchmark tests if model embeddings work well in recognizing gender based social membership for everyone. This benchmark involves similarity search in the embedding space of raw pre-trained models. The \texttt{Database} is image features on \utk and \texttt{Queries} is image features on \casualconv. For each models, features are extracted on both datasets and cosine-similarity search is used for same-attribute (gender) retrieval. \underline{Higher number is better}. We observe that our model obtains the best precision and it increases with model size.
  }
  \label{tab:indicator1_results}
\end{table*}
\begin{table}[t]
  \centering
   \setlength{\tabcolsep}{0.3em}\scalebox{0.81}{
  \begin{tabular}{@{}lll c ccc@{}}
  \toprule
    &&& & \multicolumn{3}{c}{\textbf{\texttt{P@1} difference}} \\
  \cmidrule{5-7} 
    \textbf{Model} & \textbf{Data} & \textbf{Arch.} && gender && skintone \\
                   & &    && (male - female) && (ligher - darker) \\
  \midrule 
    \supervised & INet-1K & RG-128Gf  && $\phantom{0}24\%$ && $\phantom{0}8\%$ \\

  \midrule   \multicolumn{7}{@{}l}{\textit{Self-supervised pretraining on ImageNet}} \\
  \swav       & INet-1K & RG-128Gf  && $\phantom{0}31\%$ && $\phantom{0}11\%$ \\

  \midrule
  \multicolumn{7}{@{}l}{\textit{Pretrained on random internet images}} \\
    \seer (ours)      & IG-1B & RG-128Gf  && $\phantom{0}9\%$ && $\phantom{0}7\%$ \\
    \seer (ours)       & IG-1B & RG-10B  && $\phantom{0}2\%$ && $\phantom{0}3\%$ \\
  \bottomrule
  \end{tabular}
  }
  \caption{
    \textbf{\underline{Disparity} in Gender retrieval performance} between subgroups for different gender and skintone corresponding to the retrieval performance in Table~\ref{tab:indicator1_results}. \underline{Lower number is better} and indicates lower disparity or in other words, the model works equally well for male / female genders and lighter/darker skintone. Our biggest model achieves lowest disparity and overall higher precision.
  }
  \label{tab:indicator1_disparity_comparison}
\end{table}

\subsection{Fairness}
\label{sec:fairness_section}

The ubiquitous use of computer vision models in many applications has also raised questions about their societal implications.
This necessitates the need to properly measure and quantify what harms and biases a model has with respect to societal groups of various membership types (\eg age, gender, race, skintone \etc). 
\seer models demonstrate strong performance on a broad range of publicly available computer vision benchmark tasks.
As models improve in performance on such tasks, the likelihood of using a model ``off-the-shelf'' for downstream applications increases and the nature and context of such applications is hard to anticipate. 
Motivated by this, we probe the fairness of \seer models. 

We follow the protocols \etal~\cite{goyal2022fairness} to probe the performance of our larger SEER models on three different fairness indicators:
\textit{(i)} disparities in learned representations of people’s membership in social groups Sec.~\ref{sec:indicator1_description},
\textit{(ii)} harmful mislabeling of images of people in Sec.~\ref{sec:label_association},
\textit{(iii)} geographical disparity in object recognition in Sec.~\ref{sec:geo_fairness_dollar_street}. Further, we also test on multimodal (image and text) hate speech detection for different types of hate-speech in Sec.~\ref{sec:hateful_memes_section}. 

We note that our motivation behind these fairness probes is not to validate the use of any given model.
As noted in ~\cite{goyal2022fairness}, for a given model, the choice of what fairness probes to measure depends on the application and use context.
This choice must be thoroughly assessed by the stakeholder so as to answer why those probes are chosen, what kind of assumptions are embedded in this choice, and what specific questions do the system designers aim to answer \cite{paml, kalluri2020don}. 
Therefore, we ask practitioners and developers to \textit{not} treat these results as a validation of use of a model.

\subsubsection{Indicator1: Same Attribute Retrieval}
\label{sec:indicator1_description}
We directly apply the benchmark protocol (including data preparation) as proposed in ~\cite{goyal2022fairness}. 
In this experiment, we perform \textit{similarity search}, which requires a set of \texttt{Queries} and a \texttt{Database}. 
For \texttt{Queries}, we use  the \texttt{mini test} split of \casualconv~\cite{hazirbas2021towards} which has $2,982$ videos (two videos per participant with one dark and one bright lighting video when possible). 
The dataset provides \textit{self-identified} age (from $18$ to $85$) and gender (`male', `female', `other' and `n/a') labels along with annotated Fitzpatrick skintone~\cite{FitzPatrick75}. 
For each video, when possible, we use the middle frame and use the face crops from each image. 
As \texttt{Database}, we use the \utk~\cite{zhifei2017cvpr} dataset which has $24,108$ face images annotated with apparent age and gender labels. 
Following Buolamwini \etal~\cite{buolamwini2018gender}, we group the Fitzpatrick scale into two types: Lighter (Type I to Type III) and Darker (Type IV to Type VI). 
As a result, we obtain four gender-skintone subgroups \texttt{[female, male]} $\times$ \texttt{[lighter, darker]} and four age subgroups $18-30$, $30-45$, $45-70$, $70+$. 
We extract features on \casualconv and \utk and for each query, retrieve the closest image in the Database based on cosine similarity metric. 
We perform similarity search for the gender attribute and measure \texttt{P@1} for different sub-groups: gender, skintone and age groups. 

\begin{table*}[t]
  \centering
  
  \begin{adjustbox}{max width=\textwidth}
    \begin{tabular}{@{}llll   rrrr c rrrr@{}}
      \toprule
      &&&& \multicolumn{4}{c}{\textbf{Gender Skintone}} &&  \multicolumn{4}{c}{\textbf{Age Groups}} \\
      \cmidrule{5-8} \cmidrule{10-13}
      \textbf{Model} & \textbf{Data}       & \textbf{Arch.} & \textbf{Assoc.} &  \makecell{female\\darker} & \makecell{female\\lighter} & \makecell{male\\darker} & \makecell{male\\lighter} &&  \makecell{18-30} & \makecell{30-45} & \makecell{45-70} & \makecell{70+} \\
      \midrule                 \supervised & INet-1K & RG-128Gf & \texttt{Non-Human}      & 2.3  & 6.0  & 2.0  & 1.8  && 2.1  & 2.4   & 5.4 & 4.9 \\
                  &   &      & \texttt{Crime}          & 1.2  & 0.2  & 0.7  & 0.4  && 0.6  & 0.9   & 0.1 & 3.2 \\
                  &   &      & \texttt{Human}          & 37.4 & 18.5 & 29.5 & 17.5 && 26.9 & 25.7  & 22.8 & 21.0 \\
                  &   &      & \texttt{Possibly-Human} & 24.3 & 41.4 & 50.1 & 54.0 && 43.9 & 43.7  & 39.7 & 22.7 \\
      \midrule       \multicolumn{12}{@{}l}{\textit{Self-Supervised pretraining on ImageNet}} \\
            \swav       & INet-1K & RG-128Gf & \texttt{Non-Human}      & 0.1  & 0.2  & 0.3  & 0.1  && 0.1  & 0.2  & 0.2  & 0.1 \\
                  &   &      & \texttt{Crime}          & 0.1  & 0.1  & 0.3  & 0.1  && 0.1  & 0.3  & 0.1  & 0.1 \\
                  &   &     & \texttt{Human}          & 58.7 & 58.2 & 32.2 & 43.1 && 46.6 & 44.7 & 57.9 & 46.8 \\
                  &   &    & \texttt{Possibly-Human} & 66.9 & 66.4 & 82.5 & 70.4 && 70.8 & 73.4 & 69.1 & 53.2 \\
      \midrule
      \multicolumn{12}{@{}l}{\textit{Pretrained on random internet images}} \\
                                              \seer (ours)      & IG-1B & RG-128Gf & \texttt{Non-Human}      & 0.1  & 0.6  & 0.7  & 0.7  && 0.8  & 0.1  & 0.5  & 3.2 \\
                  & &        & \texttt{Crime}          & 0.1  & 0.1  & 0.2  & 0.1  && 0.1  & 0.1  & 0.2  & 0.1 \\
                  & &        & \texttt{Human}          & 78.7 & 73.3 & 40.0 & 43.3 && 58.4 & 57.4 & 66.1 & 67.7 \\
                  & &        & \texttt{Possibly-Human} & 23.8 & 21.8 & 56.4 & 40.6 && 38.7 & 38.6 & 24.8 & 6.45 \\
        \cmidrule{3-13}
      \seer (ours)       & IG-1B & RG-10B & \texttt{Non-Human}      & 0  & 0.1   & 0.2  & 0    && 0.1  & 0    &   0.1  &  0 \\
                  &  &       & \texttt{Crime}          & 0    & 0  & 0.2  & 0.1  && 0   & 0.1  &  0  & 1.6 \\
                  &  &      & \texttt{Human}          & \textbf{93.0} & \textbf{87.3} & \textbf{57.2} & \textbf{59.8} && \textbf{73.3} & \textbf{72.7} &  \textbf{82.4} & \textbf{79.0} \\
                  &  &     & \texttt{Possibly-Human} & 20.2 & 27.9 & 72.6 & 65.1 && 44.9 & 48.3 & 39.5 & 22.6 \\

      \bottomrule
    \end{tabular}
      \end{adjustbox}
  \caption{
    \textbf{Label Association Fairness Indicator2} results of several models on the \casualconv Dataset as described in Sec.~\ref{sec:label_association}. This indicator helps measure magnitude of Harmful (\texttt{Non-Human, Crime}) label predictions for images of people. \underline{Lower [\texttt{Non-Human, Crime}] is better} and \underline{Higher [\texttt{Human}] is better}.
    Since self-supervised models don't predict labels, all models need to be adapted to image classification task. We full-finetune \textit{all} models on same subset of \ImNetFull dataset. We then, for each gender and skintone perform inference of transferred models on the \casualconv Dataset and measure \textbf{percentage of images associated with different labels} at confidence \textbf{threshold $0.1$} following~\cite{goyal2022fairness}. We observe that our model makes the least Harmful predictions and most Human predictions on images of people.
  }
  \label{tab:indicator2_results}
\end{table*}

\begin{table}[t]
  \centering
  \begin{adjustbox}{max width=\linewidth}
    \begin{tabular}{@{}lll@{}}
      \toprule
      \textbf{Association} & \textbf{Type} & \textbf{Labels in the \ImNetDataset taxonomy} \\
      
      \midrule
      \texttt{Non-Human} & Harmful & \makecell[l]{swine, slug, snake, monkey, lemur, \\
      chimpanzee, baboon, animal, bonobo, \\
      mandrill, rat, dog, capuchin, gorilla, \\
      mountain gorilla, ape, \\
      great ape, orangutan.} \\
      \midrule
      \texttt{Crime} & Harmful & prison \\
      \midrule
      \texttt{Human} & Non-Harmful & face, people\\
      \midrule
      \texttt{Possibly-Human} & Non-Harmful & makeup, khimar, beard \\
      \bottomrule
    \end{tabular}
    
  \end{adjustbox}
  \caption{
    \textbf{Label association types for ImageNet taxonomy} for computing Harmful and Non-Harmful label associations (Sec.~\ref{sec:label_association}).
  }
  \label{tab:label_association}
\end{table}

This first indicator allows to measure the disparity in the learned representations of people by directly using the raw model embeddings.
If a model has higher \texttt{P@1} for "male" than for "female", this indicator tells how much the model falsely recognizes a true female population as male \ie does mis-gendering. 
We show the results of this indicator in Table~\ref{tab:indicator1_results} and further measure the disparity between different genders and skintones in Table~\ref{tab:indicator1_disparity_comparison}. 
We make sevaral observations.
\textit{First}, models pretrained on \ImNet have a higher disparity.
\textit{Second}, SEER models have the lowest disparity between different genders and skintones.
\textit{Finally}, we observe that for SEER models, as the model size increases, the disparity decreases.
That means the model embeddings seem to recognize different genders and skintones more fairly.
We hypothesize that this is because SEER is pretrained on a very diverse dataset (see Fig.~\ref{fig:ig10m_data_distribution}) and the size of the model allows to better extract the salient information present in the image leading to better visual features.
The baseline models are trained on \ImNet whose disparity has been empirically confirmed in previous work~\cite{yang2020towards}. 

\begin{table*}[t]
  \centering
  \begin{adjustbox}{max width=\textwidth}
    \begin{tabular}{@{}lll c ccc c cccc@{}}
      \toprule
      && && \multicolumn{3}{c}{\textbf{Income buckets}} &&  \multicolumn{4}{c}{\textbf{Regions}}   \\
      \cmidrule{5-7} \cmidrule{9-12} 
      \textbf{Model} & \textbf{Data} & \textbf{Arch.} && low & medium & high && Africa & Asia & Americas & Europe \\
      \midrule
      \multicolumn{12}{@{}l}{\textit{Supervised pretraining on ImageNet}} \\
      \supervised & INet-1K & RG-128Gf && 48.3 & 67.2 & 77.9 && 54.2 & 65.3 & 70.7 & 76.2 \\
      \midrule
      \multicolumn{11}{@{}l}{\textit{Pretrained on random internet images}} \\
      \seer (ours)     & IG-1B & RG-128Gf && 59.5 & 77.8 & 86.0 && \textbf{66.0} & 75.9 & 79.5 & 84.6 \\
      \seer (ours)      & IG-1B & RG-10B && \textbf{59.7} & \textbf{78.5} & \textbf{86.6} && \underline{65.9} & \textbf{76.3} & \textbf{81.1} & \textbf{85.6} \\
      \midrule
      \multicolumn{12}{@{}l}{\textit{Relative improvement of pretraining on random internet images over ImageNet}}\\
      \multicolumn{2}{@{}l}{SEER RG-128Gf \textit{vs} Sup. RG-128Gf} &&& +23\% & +16\% & +10\% && +21\% & +16\% & +13\% & +11\% \\
      \multicolumn{2}{@{}l}{SEER RG-10B \textit{vs} Sup. RG-128Gf} &&& \textbf{+24\%} & \textbf{+17\%} & \textbf{+11\%} && \textbf{+22\%} & \textbf{+17\%} & \textbf{+15\%} & \textbf{+12\%} \\
          \bottomrule
    \end{tabular}
  \end{adjustbox}
  \caption{
    \textbf{Geographical Fairness Indicator3 results} and \textbf{diversity analysis of object recognition} accuracy for \textbf{different income} households and \textbf{regions} of the world as described in Sec.~\ref{sec:geo_fairness_dollar_street}. This indicator allows measuring how good the model is at detecting objects all over the world and in various households of varying income brackets. \underline{Higher number is better}. We observe that our model achieves better object recognition accuracy for all income brackets and regions of the world. Moreover, the object recognition accuracy improves the most for low- and medium-income brackets and for non-America/non-Europe regions of the world.
  }
  \label{tab:dollar_street_income_region_analysis}
\end{table*}

\subsubsection{Indicator2: Label Association}
\label{sec:label_association}

We use the \casualconv dataset as described in Sec.~\ref{sec:indicator1_description} which has $2,982$ images of faces of people.
For the unsupervised models, since they do not predict labels by design, we first adapt the model by finetuning it on a subset of \ImNetFull~\cite{goyal2022fairness}. 
For fair comparison, we apply the same finetuning steps to all models. 
Afterwards, for each image in the \casualconv dataset, we perform model inference and record top-5 label predictions along with the confidence scores. 
For each image, we study the type of label predicted where the labels are grouped in various association types as described in Table~\ref{tab:label_association}.

This indicator allows to measure the harmful predictions of a model, in particular when mis-labeling images of people. 
These harms can be bigger if the type of predicted labels are derogatory or reinforce harmful stereotypes~\cite{doi:10.1177/2378023120967171,bhargava2019exposing}. 
As proposed in the benchmark~\cite{goyal2022fairness}, we study the predictions with a confidence threshold of $0.1$~\cite{stock2018convnets}. 
This is in contrast to reporting the top-5 predicted labels, irrespective of confidence. 
We compare our models with two baselines and report the results of this study in Table~\ref{tab:indicator2_results}.

On one hand we see that the supervised model trained on ImageNet makes the most \texttt{Non-Human} predictions for all gender, skintone and age-groups. 
Within this, the models predict \texttt{Non-Human} labels most often for "female" and age group "45-70".
Moreover, the supervised ImageNet model also makes the most \texttt{Crime} predictions for all gender, skintone and age-groups. 
The disparity is greatest for "male-darker".
On the other hand, \seer models make the most \texttt{Human} predictions.
For a given face crop image, this model will more likely predict one of the \texttt{[face, people]} labels for all gender, skintone and age groups.
we note that the \texttt{Human} label prediction is least for "male" skintone with a disparity of ~30\% between "male" and "female".
Also, we observe that as the SEER model size increases, the association of the \texttt{Human} labels increases significantly ($+10\%$ from RG-128Gf to RG-10B across genders and skintones).
We hypothesize that since SEER is trained on the human-centric Instagram data (while ImageNet is object centric), it has learned better and fairer representations of people. 
Further, since the Instagram data represents content from "female" more (see Figure~\ref{fig:ig10m_data_distribution}), the dataset makes more human-centric predictions for female.

\subsubsection{Indicator3: Geographical Fairness}
\label{sec:geo_fairness_dollar_street}

We use the \dollarstreet dataset~\cite{de2019does} and benchmark protocol~\cite{goyal2022fairness} for evaluating the disparity in object recognition accuracy in different parts of the world. 
The dataset is composed of $16,073$ images from $289$ households of varying income levels, representing $94$ concepts across $54$ countries over $4$ regions of the world. 
The data distribution per country and per region is shown in Appendix Fig.~\ref{fig:dollar_street_data_distribution}.

\begin{figure}[h]
  \centering
    \includegraphics[width=\linewidth]{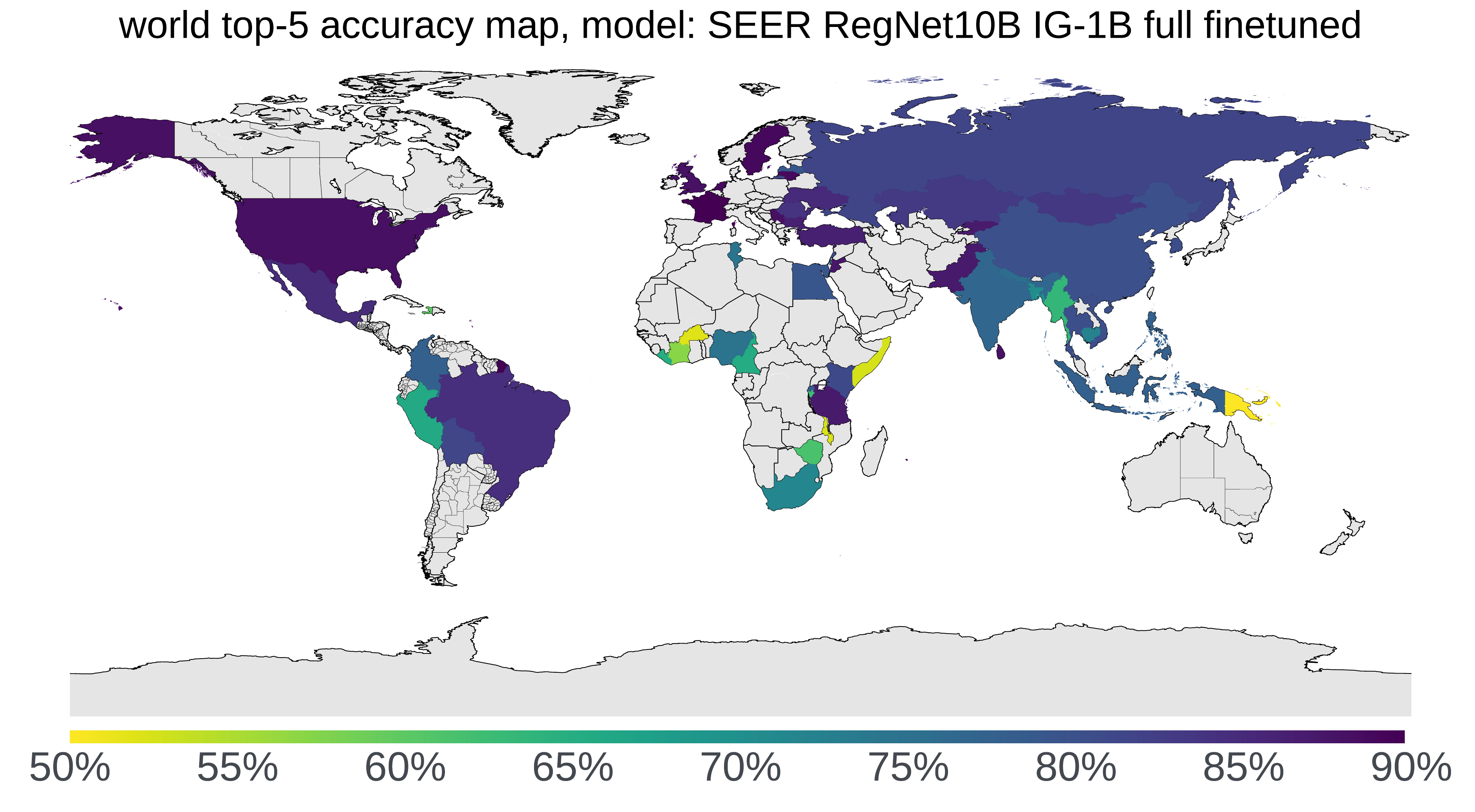}
  \caption{
    \textbf{Geographical fairness} object recognition top-5 accuracy \textbf{per country} of SEER RG-10B model on Dollar Street Dataset. This dataset comprises of $54$ countries and we show accuracy per country.
      }
  \label{fig:seer_rg256_dollar_street_accuracy} \end{figure}

As in the previous Indicator2, since self-supervised models do not have capability to predict labels, we finetune the models on a subset of \ImNetFull.
We use the manual mapping of \dollarstreet classes to \ImNetFull classes proposed in previous work~\cite{goyal2022fairness}.
Using this mapping, we retain from \ImNetFull a subset of $127$K images spanning $108$ concepts. 
For fair comparison, we finetune all the models on this data. 
Once the models are adapted, we run inference on the $16$K images from \dollarstreet and record the model predictions. 
We measure performance by computing the \texttt{top-5} accuracy. 
For analysing fairness, we follow~\cite{goyal2022fairness} and aggregate the model predictions by household and split per income level and per region.
We report this analysis in Table~\ref{tab:dollar_street_income_region_analysis}.

This indicator measures if a model is capable of recognizing concepts across different income households across different regions of the world. 
From the analysis in Table~\ref{tab:dollar_street_income_region_analysis}, we observe that the improvement of SEER models over the supervised baseline is smallest for high income households and the American / European regions.
At the same time, the relative improvement in accuracy is significant for the other groups ($+23\%$ for low-income households and $+21\%$ for the African region). As the model size increases to 10B parameters, the trend holds.
As for the previous experiments, we hypothesize that the performance of SEER follows this pattern because of the diversity of our pre-training data.
As shown in Fig.~\ref{fig:ig10m_data_distribution}, the pre-training data distribution is geographically diverse compared to datasets such as ImageNet, which mostly contain data from Western countries.

\begin{figure*}[t]
  \centering
  \includegraphics[width=\textwidth]{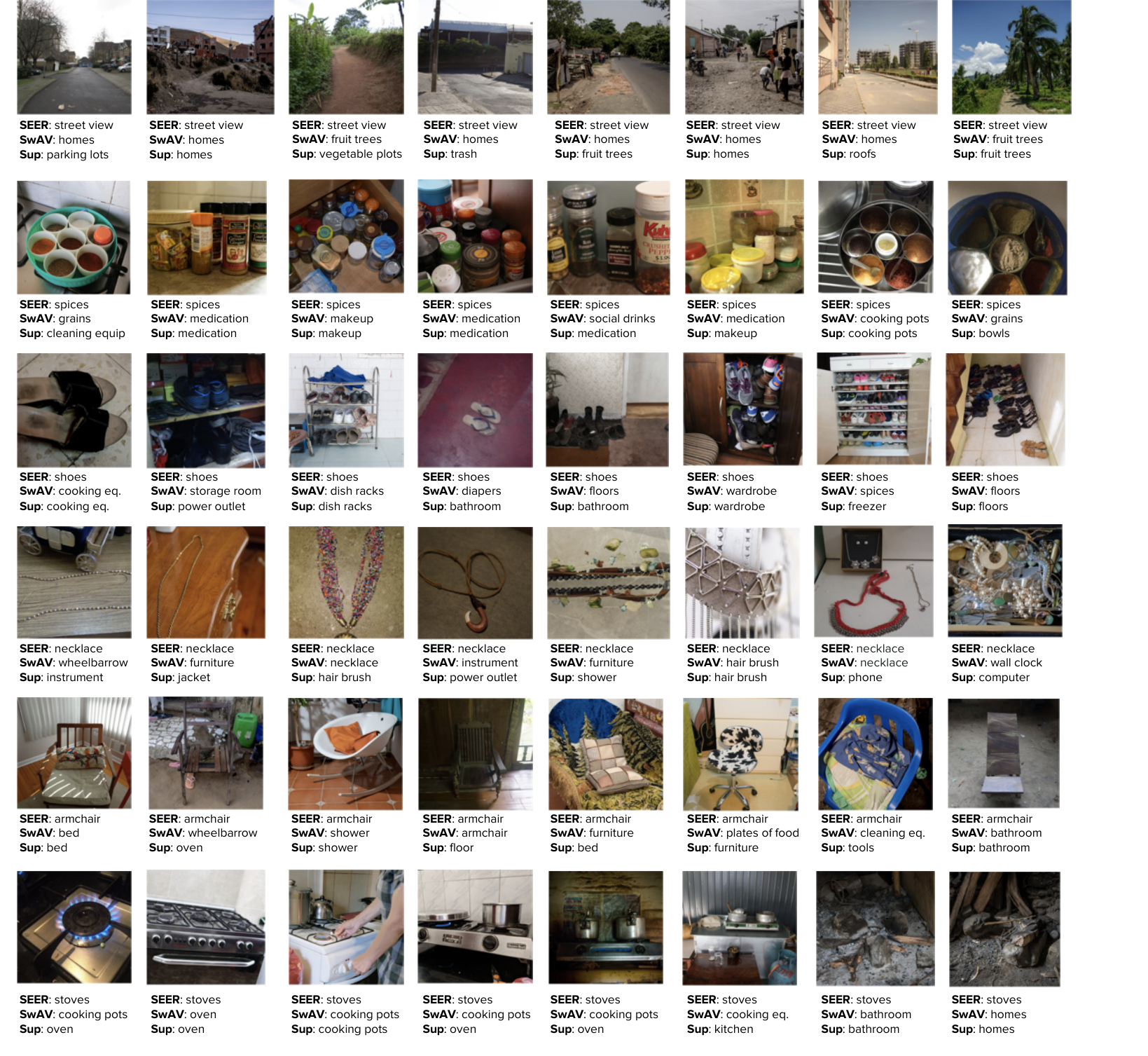}
  \caption{
    \textbf{Qualitative analysis of Geographical fairness} on \dollarstreet dataset as described in Sec.~\ref{sec:geo_fairness_dollar_street}. For a fixed architecture, here RG-128Gf, we show a few example images of improvement where our \seer model outperforms self-supervised and supervised pre-training on \ImNet. The examples are from various households of varying income levels and regions of the world. See supplemental material for license information.
  }
  \label{fig:dollar_street_samples} \end{figure*}

\subsubsection{Hate Speech Detection: \hateful}
\label{sec:hateful_memes_section}

For this experiment, we use the \hateful Challenge Dataset~\cite{kiela2021hateful}.
This is a multi-modal dataset consisting of $10,000$ images with associated text annotated with types of hate speech.
The hate speech categories are: inciting violence, dehumanizing, inferiority, contempt, mocking, slurs, exclusion and no hate-speech.
Those are further split into different protected categories (race, religion, gender, disability, nationality and `pc\_empty` for no protected category). 
The \texttt{train} split contains $8,500$ memes and the \texttt{dev} set contains $500$ memes. 
The distribution of different protected categories and types of hate-speech is shown in Appendix Figure~\ref{fig:hateful_memes_distribution}.

\begin{table}[t]
  \centering
\begin{adjustbox}{max width=\textwidth}
  \begin{tabular}{@{}lll c c@{}}
    \toprule
    && && \textbf{\hateful} \\
    \textbf{Model} & \textbf{Data} & \textbf{Arch.} && \textbf{ROC AUC} \\
    \midrule 
    \multicolumn{5}{@{}l}{\textit{Supervised pretraining on ImageNet}} \\
    \supervised & INet-1K & RN-152    && 70.1 \\
    \supervised & INet-1K & RG-128Gf  && 68.1 \\
    \midrule 
    \multicolumn{5}{@{}l}{\textit{Self-supervised pretraining on ImageNet}} \\
    \swav       & INet-1K & RG-128Gf  && 66.8 \\
    \midrule
    \multicolumn{5}{@{}l}{\textit{Pretrained on random internet images}} \\
      \seer (ours)      & IG-1B & RG-128Gf  && 72.2 \\
        \seer (ours)      & IG-1B & RG-10B  && \textbf{73.4} \\
  \bottomrule
  \end{tabular}
  \end{adjustbox}
  \caption{
    \textbf{Hate Speech Detection Performance} of several models on the \textbf{\hateful} \texttt{dev} set as described in Sec.~\ref{sec:hateful_memes_section}. For each model, we run evaluation using three different seeds and report the average performance over all seeds. We observe that our model achieves the best performance in hate speech detection and the performance improves as the model size increases.
  }
  \label{tab:hateful_memes_results}
\end{table}

We use our model as an image encoder and extract the visual features for all images in the \hateful dataset.
For all models, we extract the features before the final pooling layer in order to preserve the spatial information. 
We use BERT-Base~\cite{devlin2019bert} as the text encoder. 
We concatenate the image features with the BERT text features and train an MLP head on top. 
We use the AdamW optimizer~\cite{loshchilov2019decoupled} with epsilon $1e-8$, a learning rate of $8e-5$.
We use a linear learning rate warmup for $2,000$ iterations followed by step decays value by $1e-5$ every $500$ iterations.
We train~\footnote{We use open source library \url{https://github.com/facebookresearch/mmf}.} for a total of $22,000$ iterations with a batch size of $64$. 
We report the best \texttt{ROC AUC} metric on the \texttt{dev} set during training.

For each model, we run the evaluation with three seeds ($100$, $200$ and $300$) and report the average \texttt{ROC AUC} on the \texttt{dev} set in Table~\ref{tab:hateful_memes_results}. 
We observe that our \seer models outperform supervised ImageNet trained models by more than $2$ pts.
Interestingly, the same self-supervised learning algorithm applied on ImageNet (\swav) does not yield good performance.
We further note that as the model size increases to $10$B parameters, the performance increases.
We hypothesize that, similar to the fairness indicators in Sec.~\ref{sec:fairness_section}, the diversity and the human-centric nature of the pre-training data leads to better hate-speech detection performance.

\subsection{Transfer Learning on computer vision tasks}
\label{sec:transfer_learning_section}

In previous Sec.~\ref{sec:fairness_section}, we extensively analysed the SEER models for societal implications models can have by, for instance, mislabeling photos of people with harmful labels (derogatory, stereotypes), disparity in learned representation of people's social membership (\eg mis-gendering), hate speech detection and fairness in object recognition capability for various income households across the globe and we observed promising results for our models across the board.

In this section, we analyze the quality of visual representations learned by model on a broad range of computer vision tasks as there's no general agreement on what qualifies for universal or ``ideal'' visual representation~\cite{locatello2020}. To this end, we benchmark the \textit{robustness of models to distribution shift} in Sec.~\ref{sec:ood_section}, \textit{fine-grained recognition} performance on challenging datasets such as \inat~\cite{van2018inaturalist} in Sec.~\ref{sec:inat_section} including the application of model in wildlife conservation efforts, image retrieval (copy detection) in Sec.~\ref{sec:copy_detection_results}, and \textit{representation learning via linear-probe} to test image classification performance on more than $25$ standard object and scene datasets including \ImNet~\cite{olga2015imagenet} (object centric), \Places~\cite{places2014} (scene centric) and PASCAL \VOCseven~\cite{everingham2010pascal} (multi-label) in Sec.~\ref{sec:summarize_representation_learning_main}. We also compare our model performance with state-of-the-art supervised and self-supervised learning on \ImNet on computer vision datasets capturing variety of applications such as OCR, activity recognition in videos, scene recognition, medical and satellite images, structured datasets (to test localization) in and show full results in Appendix~\ref{sec:appendix_reprensentation_learning}.

\subsubsection{Baselines}
\label{sec:models_baseline}
For all benchmarks in this section, we compare performance of our models with supervised learning and state-of-the-art self-supervised learning approaches on ImageNet. For each self-supervised learning approach, we chose the largest publicly available pre-trained model checkpoint and ran evaluations with them. Concretely, the models we compare with are \textit{ConvNets} including \simclrbig~\cite{chen2020big} ($795$M params), \byolbig~\cite{grill2020bootstrap} ($250$M params), \swavbigrn ($585$M params) and \swavbigregnet~\cite{caron2020unsupervised} ($693$M params) and more recent \textit{Vision Transformers}~\cite{dosovitskiy2021image} including \moco \vit-B/$16$~\cite{chen2021empirical} ($85$M params) \dino \vit-B/16~\cite{caron2021emerging}. For SEER models, we trained several model sizes from 40M parameters to 10B parameters as described in Appendix Table~\ref{tab:seer_model_variants}.

\begin{table*}[t]
  \centering
  \begin{adjustbox}{max width=\textwidth}
    \begin{tabular}{@{}llll c ccccccc@{}}
      \toprule
      \textbf{Model} & \textbf{Arch.} & \textbf{Pretrain} & \textbf{Param} && \makecell{\textbf{INet} \textbf{val}} & \textbf{INet-A} & \textbf{INet-R} & \textbf{INet-Sketch} & \textbf{INet-ReaL} & \textbf{INet-v2} & \textbf{ObjectNet}\\
      
      \midrule       Supervised & RG-128Gf & INet-1K & 693M && 82.1 & 21.6 & 41.0 & 27.7 & 87.0 & 71.3 & 44.1 \\
      
      \midrule
      \multicolumn{11}{@{}l}{\textit{Self-supervised pretraining on full ImageNet}} \\

      DINO      & ViT-B/16 & INet-1K & 85M && 81.4 & 21.4 & 46.1 &  33.3 & 86.4 & 70.1 & 39.4 \\
      SimCLR-v2 & RN152w3+SK & INet-1K & 794M && 83.5 & 35.2 &  46.7 & 34.7 & 87.7 & 73.0 & 48.0 \\
      
      BYOL & RN200w2 & INet-1K & 250M && 83.5 & 43.0 & 47.1 & 35.5 & 88.1 & 73.1 & 50.7 \\
      SwAV & RN50w5 & INet-1K & 585M && 81.8 & 26.5 & 39.6 & 26.9 & 86.8 & 70.0 & 43.9 \\
      SwAV & RG-128Gf & INet-1K & 693M && 82.9 & 28.0 & 42.8 & 32.0 & 87.4 & 71.8 & 44.7 \\
      SwAV & RG-128Gf & INet-22k & 693M && 83.9 & 37.8 & 47.8 & 37.9 & 88.7 & 73.3 & 50.0 \\
      
      \midrule
      \multicolumn{11}{@{}l}{\textit{Pretrained on random internet images}} \\
      \seer (ours) & RG-128Gf & IG-1B & 693M && 84.5 & 43.6 & 51.0 &  40.2 &  89.3 &  74.7 &  54.3 \\
      \seer (ours) & RG-10B & IG-1B & 10B && \textbf{85.8} & \textbf{52.7} & \textbf{56.1} & \textbf{45.6} & \textbf{89.8} & \textbf{76.2} & \textbf{60.2} \\
      
          \bottomrule
    \end{tabular}
  \end{adjustbox}
  \caption{
    \textbf{Out-of-domain performance} of all models on various dataset with \textit{distribution shift} as described in Sec.~\ref{sec:ood_section}. The models are finetuned on ImageNet and resulting models are evaluated (inference only) on the target datasets. The best numbers for each dataset are in bold. Our model outperforms supervised and self-supervised models trained on ImageNet.
  }
  \label{tab:ood_results_imagenet}
\end{table*}

\begin{figure*}[t]
  \centering
    \includegraphics{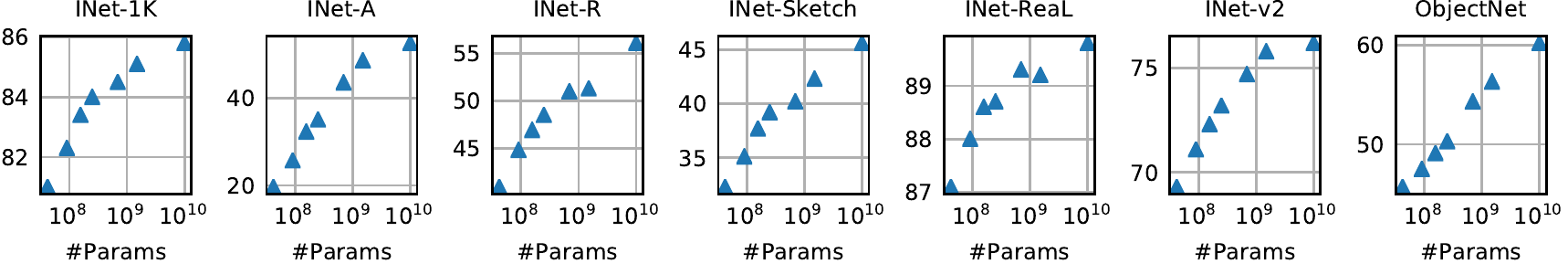}
  \caption{
    \textbf{Influence of SEER model scale on out-of-domain data generalization} and robustness to distribution shift. Across all the datasets, we observe that as the model size increases, the performance increases significantly.
  }
  \label{fig:ood_scaling_trend}
\end{figure*}

\subsubsection{Out-of-domain Generalization and Robustness}
\label{sec:ood_section}
For most ``off-the-shelf'' models in computer vision, it is hard to anticipate the exact application of models and impossible to train a model on precisely the data distribution that the model will be applied to. Inevitably, the model will encounter out-of-domain data on which the model performance can vary widely. For instance, even though deep learning models have surpassed human performance on ImageNet dataset~\cite{he2016deep}, recent works~\cite{dodge2017study,alcorn2019strike} have demonstrated that these models still make simple mistakes and have lower accuracy (than ImageNet and human) on new benchmarks~\cite{objectnet2019,recht2019imagenet}. Therefore, understanding the out-of-domain generalization of models is important. Motivated by this, we probe the performance of our models on out-of-domain datasets. To measure the generalization capabilities of our model, we report the performance of the finetuned model on several alternative test sets.

\paragraph{Datasets.}
A recent comprehensive study~\cite{miller2021accuracy} analyzed the out-of-domain generalization and robustness of ImageNet models on several datasets (which have distribution shifts) and found that across all datasets, the accuracy of models dropped well below the expectation set by the ImageNet validation set. A few datasets tested are: ImageNet-Adversarial~\cite{hendrycks2021natural} (contains natural adversarial images), ImageNet-R~\cite{hendrycks2021many} (renditions), ImageNet-Sketch~\cite{wangh2019learning} (sketches), ImageNet-Real~\cite{beyer2020we} (corrected labels in original dataset), ImageNet-V2~\cite{recht2019imagenet} (new test set for ImageNet benchmark), ObjectNet~\cite{objectnet2019}. Each of these datasets have the subset or same labels as the original \ImNet and we use these dataset for our models benchmarking.

\paragraph{Evaluation Protocol.} We use our pre-trained SEER model trunk for initialization and attach a linear classifier head on top. We full-finetune the model weights on ImageNet task for $15$ epochs using SGD momentum $0.9$, weight decay $1e-4$, learning rate of $0.04$ for batch size $256$ and finetune on $128$ NVIDIA GPUs by scaling learning rate following Goyal \etal~\cite{goyal2017accurate}. We use step learning rate schedule with gamma of $0.1$ and decay at steps [$8$, $12$]. After finetuning the model, we evaluate the finetuned model on all $5$ datasets by performing \textit{inference} only and report the top-$1$ accuracy of several models (including our baseline models) on all datasets including the ImageNet validation set in Table~\ref{tab:ood_results_imagenet}.

\begin{figure*}[t]
  \centering
  \includegraphics[width=\linewidth]{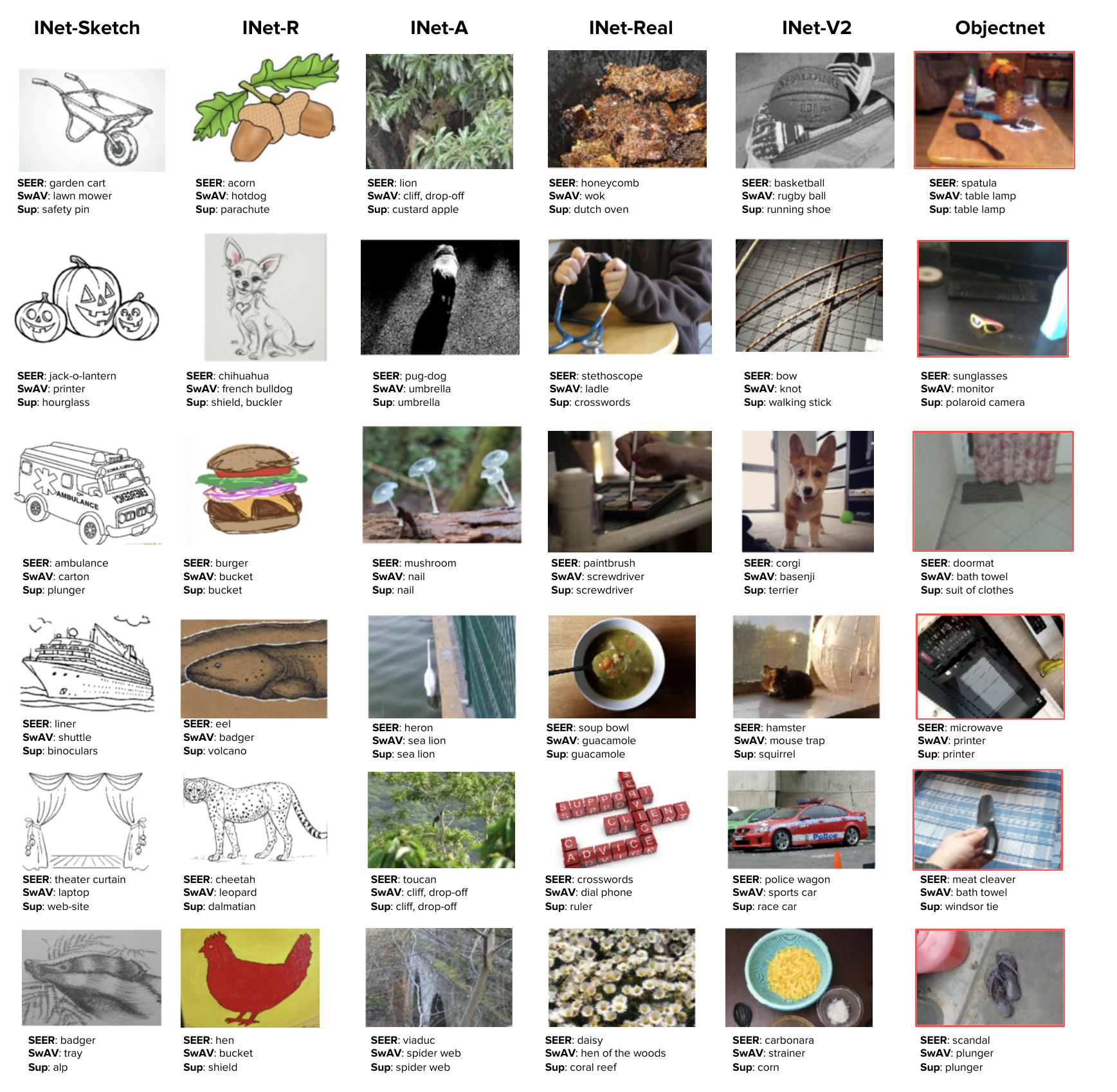}
  \caption{
    \textbf{Qualitative analysis of out-of-domain performance and robustness} as detailed in Sec.~\ref{sec:ood_section}. For the same architecture (RG-128Gf), we show few example images of improvements between our model (SEER) and ImageNet trained supervised and self-supervised (SwAV) model. We note that on out-of-domain data, self-supervised models generalize better than supervised models. Further, SEER model significantly outperforms self-supervised pre-training on ImageNet.
  }
  \label{fig:ood_visualisation}
\end{figure*}

\paragraph{Results.}
Table~\ref{tab:ood_results_imagenet} shows performance of our SEER model and comparison to its smaller versions and the baseline models in Sec.~\ref{sec:models_baseline}. We observe several interesting trends from this comparison. 
\textit{(i)} self-supervised pretraining objective are more robust and generalize better to out-of-domain data distribution compared to the supervised training objective.
\textit{(ii)} our SEER model, trained on Instagram data achieves better generalization than the self-supervised models trained on ImageNet.
\textit{(iii)} as the size of SEER models increases, the out-of-domain generalization improves significantly as evident in Figure~\ref{fig:ood_scaling_trend}. 

We further investigate our third observation for SEER models by evaluating the whole family of SEER models (outlined in Appendix Table~\ref{tab:seer_model_variants}) we trained. The trend of influence of model scale on performance is demonstrated in Figure~\ref{fig:ood_scaling_trend}. We note that the influence of scale on generalization holds true for all datasets and we observe a a log-linear scaling trend in performance improvement with model as, for all the test sets considered. Further, on some datasets such as adversarial ImageNet-A dataset, performance nearly doubles from $19.6\%$ to $52.7\%$.
The gains are most shy on ImageNet-ReaL (only +$2.7$\%) dataset which essentially is same as the ImageNet validation set but with the relabeling to improve on mistakes in the original annotation process. We qualitatively investigate the performance improvement on all these datasets and compare qualitative results for ImageNet trained models vs SEER in Figure~\ref{fig:ood_visualisation}.

\paragraph{Disentangling Factors of Influence using dSprites}
We investigate what are the factors that contribute to the better performance of SEER models on out-of-domain generalization. We hypothesize that the \textit{location} and \textit{orientation} of objects are two common factors of variation in the out-of-domain datasets. For this, we evaluate SEER models on \textit{dSprites}~\cite{dsprites17} dataset which contains simple black/white shapes rendered in $2$D and offers two tasks: location and orientation prediction. This dataset has $664$K images and is an image classification task with $16$ different locations and orientations each.

We use linear probe to evaluate SEER and baseline models on \textit{dSprites} dataset. We initialize models with respective model weight and attach an MLP classifier head on top. While keeping the model trunk fixed, we train the linear classifier head for $28$ epochs using SGD momentum 0.9, weight decay $0.0005$, learning rate of $0.01$ for a batchsize of $256$ and step learning rate schedule with gamma factor $0.1$ with decay at steps [$8$, $16$, $24$]. We share the results in Table~\ref{tab:dsprites_results} and observe that the our SEER models achieve equal or slightly better performance than baseline models on both the tasks. We thus reason that the pretraining data domain for our model instead contributes to better out-of-domain performance. 

\begin{table}[t]
  \centering
  \setlength{\tabcolsep}{0.4em}\scalebox{0.88}{
  \begin{tabular}{@{}lll c cc@{}}
  \toprule
  \textbf{Model} & \textbf{Arch.}  & \textbf{Param.} && \textbf{Orientation} & \textbf{Location} \\
  
  \midrule   \supervised & RG-128Gf  & 693M && 75.8 & 95.1 \\
  \supervised & \vit-B/16 & 85M && 33.3 & 24.3 \\

  \midrule
  \multicolumn{6}{@{}l}{\textit{Self-supervised pretraining on ImageNet}} \\
  
  \dino & \vit-B/8 &   85M && 32.6 & 24.5 \\
   \byol & RN200w2  & 250M && \textbf{80.9} & 94.6 \\
   \swav & RN50w5  & 585M && 79.5 & 91.7 \\
   \swav & RG-128Gf &  693M && 75.2 & 95.7 \\
   SimCLRv2  & RN152w3+SK  & 794M && 54.1 & 64.6 \\
  
  \midrule
  \multicolumn{6}{@{}l}{\textit{Pretrained on random internet images}} \\
  \seer (ours) & RG-128Gf  &  693M   && 75.9 & 95.5 \\
  \seer (ours) & RG-10B    & 10B    && \textbf{80.9} & \textbf{96.3} \\
  \bottomrule
  \end{tabular}}
   \caption{
  \textbf{Disentangling factors influencing model performance on Out-of-Domain} benchmarks. We show model performance comparison on \textbf{\dsprite dataset} which contains simple black/white shapes rendered in $2$D, with two tasks: \textit{location} and \textit{orientation} prediction. On both tasks, our model achieves equal or better performance compared to the baseline models. We hypothesize that training on more diverse dataset instead contributes to better out-of-domain performance of our model.
  }
  \label{tab:dsprites_results}
\end{table}

\begin{table*}[t]
  \centering
\begin{adjustbox}{max width=\linewidth}
  \begin{tabular}{@{}llc c cc c cc@{}}
  \toprule
    & & && \multicolumn{2}{c}{\textbf{\inat}} && \multicolumn{2}{c}{\textbf{\iwild}} \\
    \cmidrule{5-6} \cmidrule{8-9}
    \textbf{Model} & \textbf{Arch.} & \textbf{Pretrain} && \textbf{linear} & \textbf{finetuned} && \textbf{linear} & \textbf{finetuned} \\
  
  \midrule   \multicolumn{9}{@{}l}{\textit{Supervised pretraining on ImageNet}} \\
  Supervised~\cite{dosovitskiy2021image}  & ViT-B/16    & INet-1K                            && 40.7 & 79.8 && -- & -- \\
  Supervised~\cite{dosovitskiy2021image}  & ViT-L/16    & INet-1K                            && -- & 81.7 && -- & -- \\
  Supervised~\cite{goyal2021self}  & RG-128Gf    & INet-1K      && 47.2 & 78.7 && 73.32 & 76.9 \\
  DeiT~\cite{touvron2020training}   & ViT-B/16    & INet-1K                            && -- & 79.5 && -- & -- \\
  ERM~\cite{miller2021accuracy}         & \makecell{PNASNet-5-Large} & -- && -- & -- && -- & 77.3 \\ 
  
  \midrule   \multicolumn{9}{@{}l}{\textit{Self-supervised pretraining on full ImageNet}} \\
  
  \dino~\cite{caron2021emerging}     & ViT-B/16  & INet-1K          && 50.1 & 72.6 && -- & -- \\
  
  \simclr    & RN152w3+SK  & INet-1K       && 43.0 & 74.1 && 67.9 & 75.5\\
  \byol      & RN200w2 & INet-1K           && 45.7 & 76.1 && 73.4 & 75.8 \\
  
  \swav      & RN50w5  & INet-1K           && 48.6 & 76.0 && 73.6 & 75.7\\
  \swav      & RG-128Gf     & INet-1K      && 47.5 & 79.7 && 73.6 & 76.1 \\

  \midrule
  \multicolumn{9}{@{}l}{\textit{Pretrained on random internet images}} \\
  \seer (ours) & RG-128Gf & IG-1B && 47.2 & 82.6 && 75.7 & 78.1 \\
  \seer (ours) & RG-10B   & IG-1B && \textbf{53.0} & \textbf{84.7} && \textbf{76.4} & \textbf{78.9} \\
  
  \bottomrule
  \end{tabular}
  \end{adjustbox}
  \caption{
    \textbf{Fine-grained recognition} image classification performance of models measured via linear probe and full-finetuning on \textbf{\inat and \iwild datasets} as described in Sec.~\ref{sec:inat_section}. For \iwild dataset, we report performance on \texttt{testID} split. We observe that for both linear and full-finetuning probes, our model achieves the best performance. Further, as the size of our model increases, the performance consistently increases. Qualitative analysis of performance is presented on \inat in Figure~\ref{fig:inat18_viz} and on \iwild in Figure~\ref{fig:iwilds_viz}.
  }
  \label{tab:inat_results}
\end{table*}

\begin{figure*}[t]
  \centering
    \includegraphics[width=\linewidth]{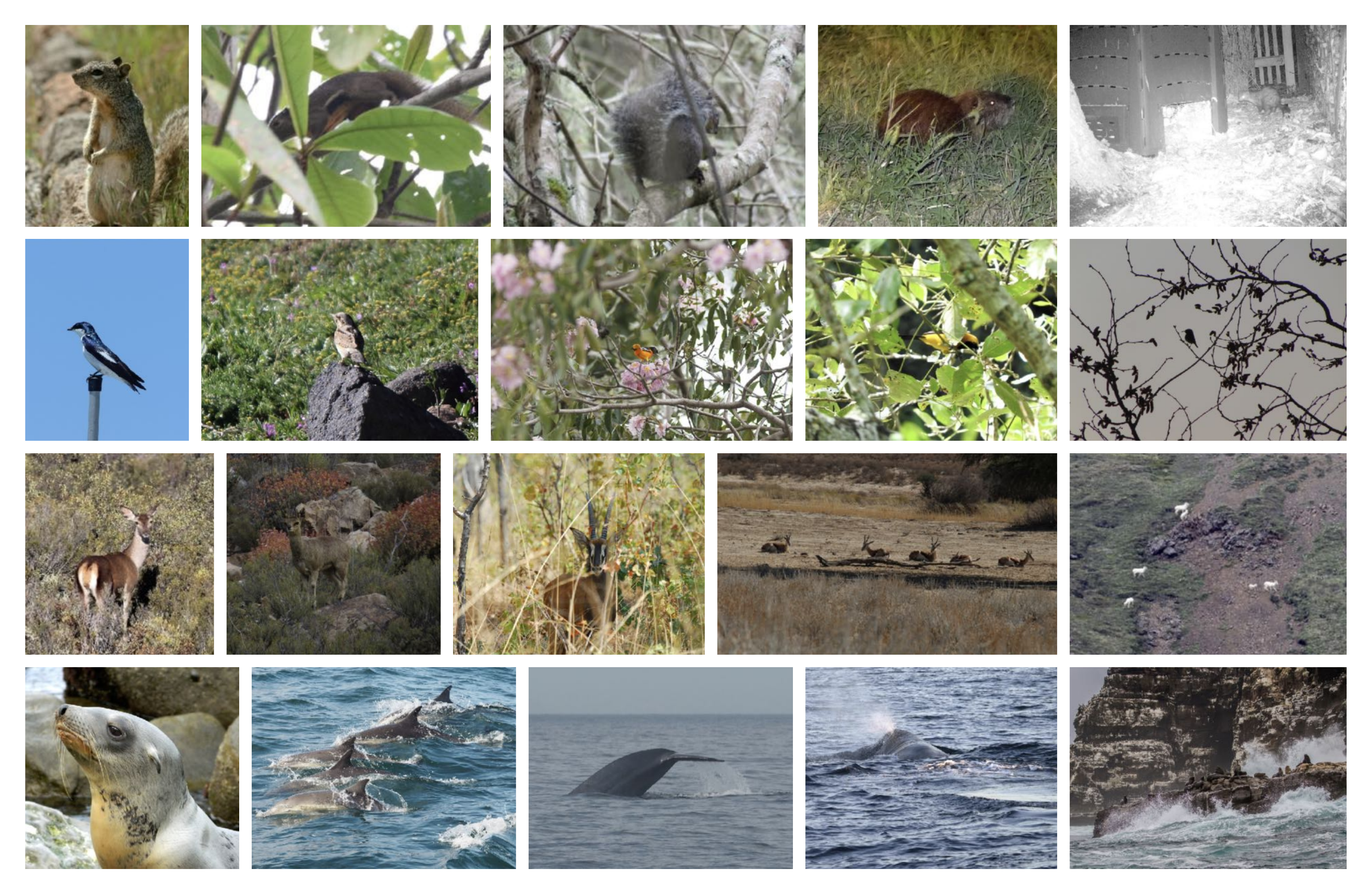}
  \caption{
    \textbf{Qualitative analysis of fine-grained recognition performance} as described in Sec.~\ref{sec:inat_section}. We show few example images from \textbf{\inat} where \seer model demonstrates better performance than pre-training on \ImNet. Each row represents a different category of animals, small mammals, birds, larger mammals and sea mammals. \seer is better at identifying a wide variety of animal species across different view points, lightning conditions, obstructions and zooms.
  }
  \label{fig:inat18_viz}
\end{figure*}

\begin{figure*}[t]
  \centering
    \includegraphics[width=\linewidth]{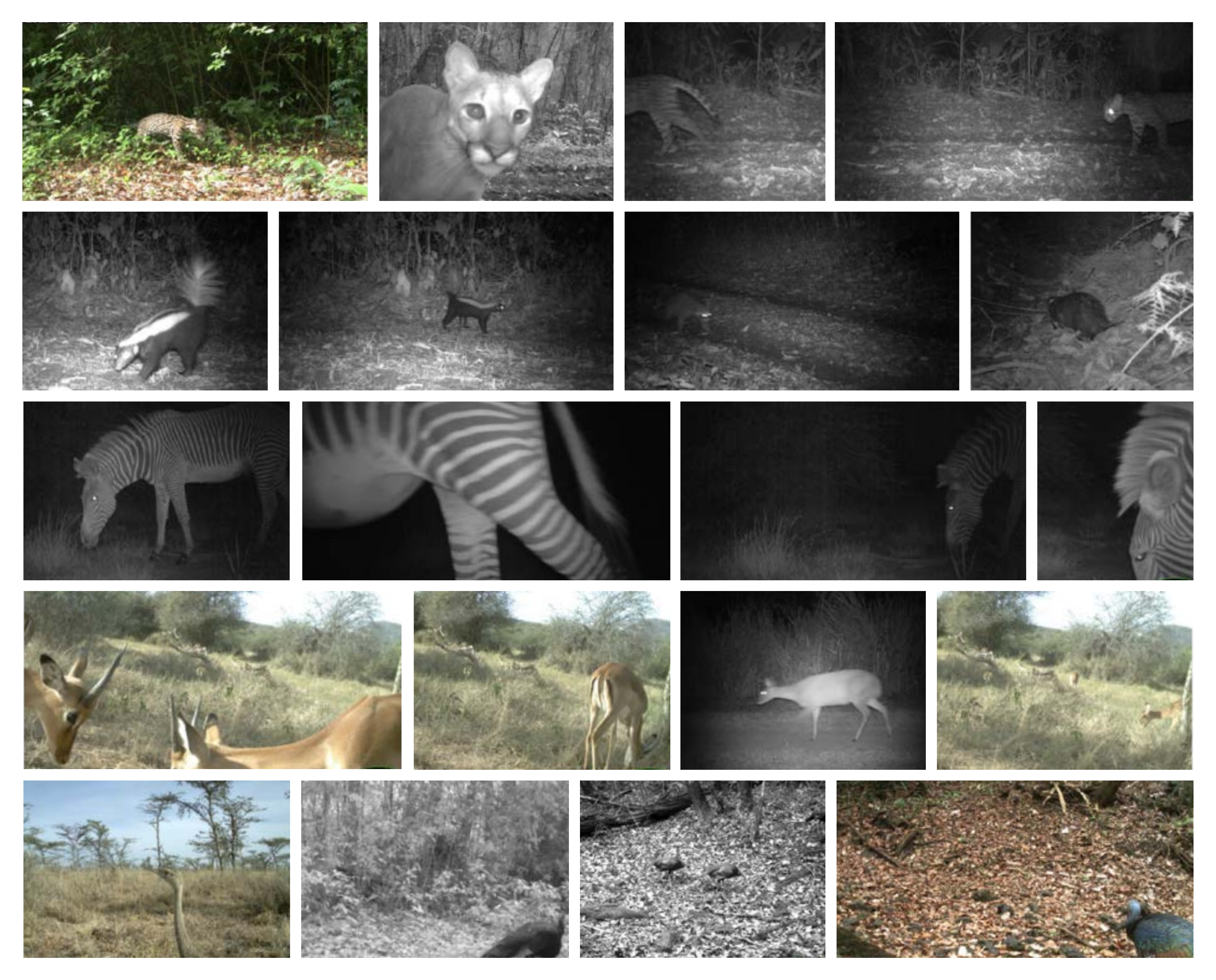}
  \caption{
    \textbf{Qualitative analysis of fine-grained performance on \textbf{\iwild}} as described in Table~\ref{tab:inat_results}. We show a few example images from \textbf{\iwild} where \seer demonstrates better performance than pre-training on \ImNet. Each row represent a different category of animals, feline predators, small mammals, zebras, gazelles and birds. \seer is better at identifying animals species across those categories across various challenges such as camouflage, blur, occlusion, motion, and unusual perspectives.
  }
  \label{fig:iwilds_viz}
\end{figure*}

\subsubsection{Fine-Grained Recognition}
\label{sec:inat_section}
We next evaluate the performance of our models on a challenging fine-grained image classification task of recognizing various animal special in the \inat~\cite{vanhorn2018inaturalist} dataset. We further evaluate how well the models generalize on a wildlife monitoring (and preservation as a result) task \iwild~\cite{koh2021wilds}. This dataset has \textit{real-world} geographic shift as the images are taken by camera traps all over the world and across different camera traps, there is drastic variation in illumination, color, camera angle, background, vegetation, and relative animal frequencies, which makes this dataset challenging~\footnote{We note that \iwild dataset enables us to test the practical application of computer vision research in wildlife preservation effort where the models are used to recognize animal species (if any) in the camera trap.}.

\paragraph{Datasets.}
\textit{\inat} dataset is composed of $437,513$ images of super-classes Mammalia, Aves, and Reptilia in train set representing a total of $8,142$ fine-grained species. 

\textit{\iwild}, adapted from iWildCam $2020$ competition dataset~\cite{beery2020iwildcam}, contains $129,809$ images of $182$ species (including ``no animal'') in the \texttt{train} set where the images are taken by $243$ camera traps deployed all over the world. The in-distribution \texttt{test} set comprises $8,154$ images taken by the same $243$ camera traps. The goal is to identify the animal species, \textit{if any}, within each photo and due to the data challenges such as camouflage, blur, occlusion, motion, perspective etc, the task is quite challenging.

\paragraph{Evaluation Protocol.} We evaluate the SEER and baseline models on these datasets using two protocols: linear and full-finetuning.

Following recent works~\cite{dosovitskiy2021image}, we perform finetuning at input image resolution $384$. Further, for \inat full-finetuning, we initialize the model weights with SEER models and attach a linear $8142$ dimensional MLP head. We finetune the full model using SGD momentum of $0.9$ for $48$ epochs. We use learning rate of $0.015$ for a batchsize of $256$ images, weight decay of $1e-4$ and use global \syncbn synchronizing the statistics across all GPU workers. We use cosine learning rate schedule decaying learning rate at every iteration to the final value of $5e-7$. We do not regularize \batchnorm and neither the bias in the model layers.

For \iwild full-finetuning, we initialize the model weights from the SEER model full-finetuned on \inat (following the guideline~\cite{beery2020iwildcam}) and attach a linear $182$ dimensional MLP head and full-finetune the model. We use SGD momentum of $0.9$ to finetune for $60$ epochs, weight decay of $1e-4$, step learning rate schedule with learning rate of $0.001$ decayed at epochs [$10$,$40$] by gamma $0.1$. We use global \syncbn to synchronize statistics across all GPU workers and also regularize \batchnorm and bias in the model layers.

For linear evaluation on \inat, we initialize models with respective model weights and attach an MLP classifier head on top. While keeping the model trunk fixed, we train the linear classifier head for $28$ epochs using SGD momentum 0.9, weight decay $0.0$, learning rate of $0.015$ for a batchsize of $256$ and step learning rate schedule with gamma factor $0.1$ with decay at steps [$8$, $16$, $24$]. 

For linear probe on \iwild, similar to full-finetuning, we initialize the model from the SEER model weights full-finetuned on \inat and follow the same linear probe strategy as for \inat in above paragraph.

\begin{table}[t]
  \centering
  
  \begin{tabular}{@{}llcc c c@{}}
  \toprule
  \textbf{Model} & \textbf{Arch.} & \textbf{dims} & \textbf{size}  & \textbf{\textit{mAP}} \\
  
  \midrule
  \multicolumn{5}{@{}l}{\textit{Supervised pretraining on ImageNet}} \\
  \multigrain~\cite{berman2019multigrain} & ResNet-50 & 2048 & long $800$ & 82.5  \\  
  
  \supervised~\cite{caron2021emerging} & \vit-B16 & 1536 & $224^2$ & 76.4  \\    
  \midrule
  \multicolumn{5}{@{}l}{\textit{Self-supervised pretraining on ImageNet}} \\
  \dino~\cite{caron2021emerging} & \vit-B/16 & 1536 & $224^2$ & 81.7  \\
  \dino~\cite{caron2021emerging} & \vit-B/8  & 1536 & $320^2$ & 85.5  \\
  
  \swav & ResNet-50     & 1024 & long 224 & 76.2  \\
  \swav & RG-128Gf & 2904 & long 224 & 83.0  \\
  
  \midrule
  \multicolumn{5}{@{}l}{\textit{Pretrained on random internet images}} \\
  \seer (ours) & RG-128Gf   & 2904 & long 224 & 86.5  \\
  \seer (ours) & RG-256Gf   & 4096 & long 224 & 87.8  \\
  
  \seer (ours) & RG-10B     & 4096 & long 384 & {88.8} \\
  \seer (ours) & RG-10B     & 9500 & long 384 & \textbf{90.6} \\
  \bottomrule
  \end{tabular}
  \caption{Image \textbf{Copy Detection} performance (\textit{mAP}) on the ``strong'' subset of the \copydays dataset as described in Sec.~\ref{sec:copy_detection_results}. We observe state-of-the-art performance using SEER models with the performance increasing with model size. We show qualitative results in Figure~\ref{fig:copydays_viz_seer_v_supervised_appendix}.}
  \label{tab:copy_detection_table}
\end{table} 
\paragraph{Results.}
We report the performance of (several variants) SEER models and baseline models from ~\ref{sec:models_baseline} in Table~\ref{tab:inat_results}. We observe that \textit{(i)} SEER models consistently achieve better visual representation for both linear and full-finetuning protocols on both \inat and \iwild datasets, \textit{(ii)} further, as the size of SEER models increases, the performance increases, \textit{(iii)} on \inat dataset which has many challenges (such as occlusion, camouflage, blur, motion), SEER outperforms other baseline models and current state-of-the-art model on leaderboard~\footnote{\url{https://wilds.stanford.edu/leaderboard/\#iwildcam}} indicating the visual quality of SEER features is more robust to these challenges, and \textit{(iv)} finally, we note that the SEER models achieve significantly better accuracy $+3.8\%$ on \inat for finetuning protocol. We hypothesize that since SEER models are trained on random Instagram data which is human-centric images and \inat contains fine-grained images of animals, mammal, aves species, full-finetuning helps to adapt the model better. We show qualitative result analysis in \inat and \iwild in Figure~\ref{fig:inat18_viz} and Figure~\ref{fig:iwilds_viz} respectively.

\subsubsection{Image Copy detection}
\label{sec:copy_detection_results}

We evaluate the performance of our models on Image Copy Detection~\cite{copydays2009} task which tests the robustness of models to adversarial attacks. This task has important practical applications in computer vision for various real-world problems such as content integrity, misinformation and user safety. This task involves identifying the source of an altered image within a large collection of unrelated images. The images are altered / manipulated by applying several data distortions such as blur, insertions, print and scan, etc making this a challenging task. \\

\noindent\textbf{Dataset.}
We use \copydays~\ dataset ``strong'' subset which has $157$ images in the \texttt{Database} and $3,212$ images as \texttt{Queries}. We augment the data with $10$K random distractor images from YFCC100M~\cite{yfcc2016} following previous works~\cite{berman2019multigrain,caron2021emerging} and denote this setting as CD10K. Image retrieval benefits from PCA whitening and thus we use an additional 20K images from YFCC100M following ~\cite{caron2021emerging,berman2019multigrain} to train PCA whitening. \\

\noindent\textbf{Evaluation Protocol.} We extract the features of our models on all images in \texttt{Database}, \texttt{Queries}, 10K distractors and $20$K whitening set. Following~\cite{tolias2016particular}, the features are pooled with regionalized pooling layer (R-MAC) with spatial level $3$~\footnote{We experimented with R-MAC and GeM both and found R-MAC to work best for SEER models} which by design also L2 normalizes the features. We train the PCA whitening on $20$K images and apply this whitening to the \texttt{Database} and \texttt{Queries} features. We then perform copy detection using cosine similarity between the database and query features and evaluate the performance using mean average precision (\textit{mAP}) metric. \\

\noindent\textbf{Results.}
We report the performance of our SEER models and baseline models in Table~\ref{tab:copy_detection_table}. We observe that \textit{(i)} self-supervised models achieve competitive performance on this task which corroborates the finding in previous work~\cite{caron2021emerging}, \textit{(ii)} We further observe that as model size increases, copy detection performance improves for the same features size, \textit{(iii)} we observe $90.6$\% mAP with best SEER model which is an improvement of +5.1\% over previous best results. We show some qualitative analysis and comparison in Appendix Figure~\ref{fig:copydays_viz_seer_v_supervised_appendix} and additional implementation details in Appendix \ref{sec:appendix_copy_detection_nuances}.

\begin{table*}[t]
  \centering
 
\begin{adjustbox}{max width=\linewidth}
  \begin{tabular}{@{}llll c ccc@{}}
  \toprule
    \textbf{Model} & \textbf{Arch.} & \textbf{Pretrain} & \textbf{Param.} && \textbf{INet-1K} & \textbf{\Places} & \textbf{\VOCseven} \\
  \midrule
  \multicolumn{8}{@{}l}{\textit{Supervised pretraining on ImageNet}} \\
  Supervised & RG-128Gf & INet-1K & 693M    && 80.6 & 56.0 & 89.4 \\
  
  Supervised & ViT-B/16 & INet-1K & 85M    && 81.6 & 53.6 & 90.5 \\
  
  \midrule
  \multicolumn{8}{@{}l}{\textit{Self-supervised pretraining on ImageNet}} \\
  \moco & VIT-B/16 & INet-1K & 85M         && 75.8 & 53.9 & 89.4 \\
  \dino & VIT-B/16 & INet-1K & 85M         && 78.2 & 55.2 & 90.6 \\
  \dino & VIT-B/8 & INet-1K & 85M          && 80.1 & 57.7 & 91.9 \\
  
  \simclr & RN152w3+SK & INet-1K & 794M    && 80.0 & 56.0 & -- \\
  \byol & RN200w2 & INet-1K & 250M         && 78.3 & 56.8 & 90.1 \\
  \dino & RN50 &     INet-1K & 25M         && 75.1 & 55.9 & 88.5 \\
  \swav & RN50 &     INet-1K & 25M         && 75.2 & 56.3 & 88.5 \\
  \swav & RN50w5   & INet-1K     & 585M    && 78.5 & 60.3 & 90.3 \\
  \swav & RG-128Gf & INet-1K     & 693M    && 78.4 & 60.1 & 91.4 \\
  
  \midrule
  \multicolumn{8}{@{}l}{\textit{Pretrained on random internet images}} \\
  \seer (ours) & RG-128Gf & IG-1B & 693M && 76.0 & 61.9 & 91.6 \\
  \seer (ours) & RG-10B   & IG-1B & 10B  && 79.8 & \textbf{62.9} & \textbf{91.8}\\
  \bottomrule
  \end{tabular}
  \end{adjustbox}
  \caption{
    \textbf{Representation learning using linear probe} on standard image classification datasets as described in Sec.~\ref{sec:summarize_representation_learning_main}. We compare performance of the several models on downstream classification tasks. We observe that, despite training our model on random internet images, our model achieves competitive or better results than ImageNet based supervised and self-supervised models.
  }
  \label{tab:linear_in1k_places_voc_results}
\end{table*}

\subsubsection{Representation learning using Linear-probe}
\label{sec:summarize_representation_learning_main}

One of the objectives of our work is to learn task-agnostic high-quality visual features from random internet image in the wild. To this end, we evaluate the quality of visual representations learned by our models during pretraining on a variety of datasets in computer vision~\cite{zhai2020largescale}. There are two widely used protocols for evaluating the visual features quality: \textit{linear-probe} and \textit{full-finetuning}. While it has been proven that fine-tuning exceeds the performance of linear classifiers~\cite{zhai2020largescale}, for our benchmarking, we choose linear-probe protocol. We make this choice because full-finetuning adapts visual features to each dataset and can compensate for and potentially mask the failures to learn general and robust representations during the pre-training. However, linear classifiers can highlight these failures which provides a better measure of the features quality.  

\paragraph{Datasets.}
We follow the previous work~\cite{zhai2020largescale} to select $25$ tasks (summarized in Appendix Table~\ref{tab:appendix_tasks_list}) that can be grouped into few categories based on the task domain. 
\textit{(i)} \textit{standard datasets} such as \ImNet, \Places and \VOCseven which haven been widely used for testing features quality in many previous works~\cite{caron2019unsupervised,caron2020unsupervised,yan2020clusterfit,wu2018unsupervised,goyal2019scaling}; 
\textit{(ii)} \textit{medical and satellite images} such as in RESISC45~\cite{resiscCheng2017}, EuroSAT~\cite{helber2019eurosat}, PatchCamelyon~\cite{veeling2018rotation}; 
\textit{(iii)} \textit{structured datasets} containing synthetic images and we select these datasets as even the best ImagNet representations fail to capture the aspects in these datasets such as counting and depth prediction tasks on \textit{CLEVR}~\cite{johnson2016clevr} which contains simple 3D shapes with two tasks, camera-elevation prediction on \textit{SmallNorb}~\cite{snorb2004} which has images of artificial objects viewed under varying conditions, location
and orientation prediction tasks on \textit{dSprites}~\cite{dsprites17} which contain 2D rendered black/white shapes; 
\textit{(iv)} \textit{activity recognition in videos} by taking the middle frame on datasets Kinetics-700~\cite{carreira2019short} and UCF-101~\cite{soomro2012ucf101} and \textit{scene recognition} on SUN397~\cite{sun397Chao2010}; 
\textit{(v)} \textit{self-driving} related tasks such as german traffic sign recognition in GTSRB~\cite{gtsrb2011}, measuring the distance of nearest vehicle in KITTI-Distance~\cite{kittidist2012}; 
\textit{(vi)} \textit{textures} on datasets such as DTD~\cite{cimpoi2013describing}; 
\textit{(vii)} \textit{natural} datasets such as STL-10~\cite{pmlr-v15-coates11a}, Oxford-IIT Pets~\cite{parkhi12a}, Oxford Flowers102~\cite{nilsback2008automated}, Caltech-101~\cite{caltech2006} and finally
\textit{(viii)} \textit{optical character recognition (OCR)} tasks such as on SVHN~\cite{netzer2011reading} which involves street number transcription on the distribution of Google Street View photos.

\begin{figure*}[t]
  \centering
  \includegraphics[width=0.33\textwidth]{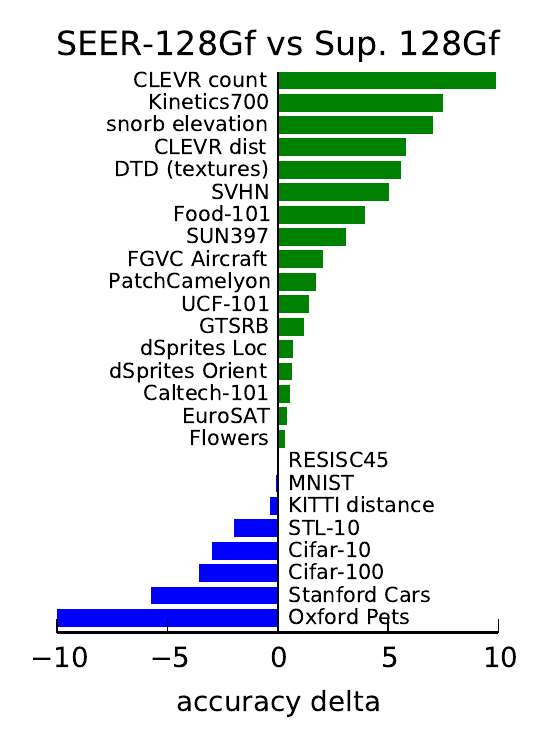}
  \includegraphics[width=0.33\textwidth]{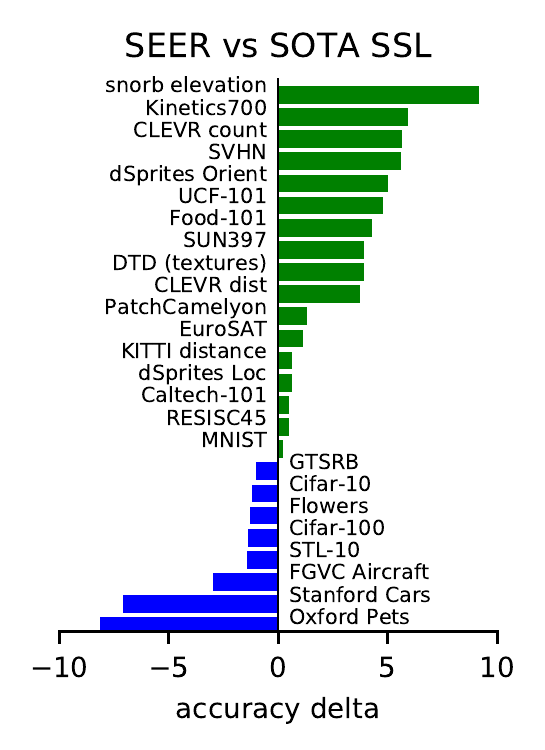}
  \includegraphics[width=0.33\textwidth]{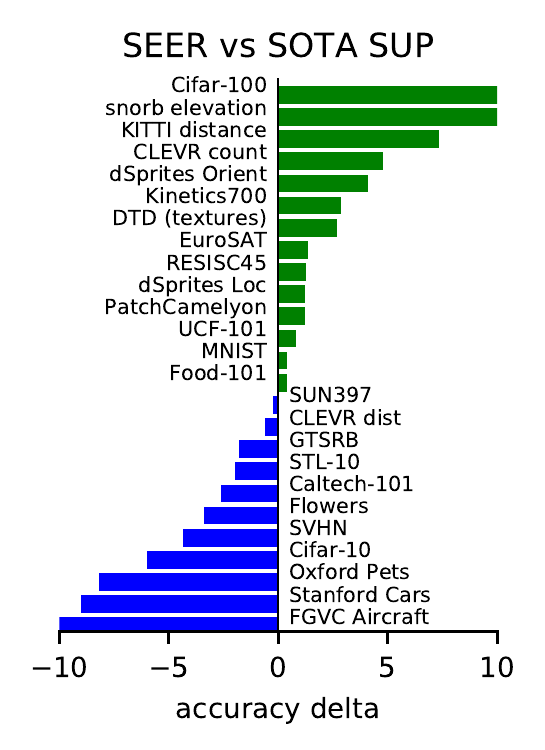}
  \caption{
     \textbf{Linear probe performance comparison} of our SEER model on $25$ linear benchmark tasks as described in Sec.~\ref{sec:summarize_representation_learning_main}. On the left, for the same architecture RG-128Gf, we show the delta in accuracy between SEER model (trained on Instagram) and supervised model trained on ImageNet. In the middle, we show the performance delta between best SEER model (trained on Instagram) and best self-supervised model (any approach, architecture, scale) trained on ImageNet. On the right, we show the delta but with the best supervised model trained on ImageNet. 
  }
  \label{fig:linear_probe_comparison_summary}
\end{figure*}

\paragraph{Evaluation Protocol.}
We train linear classifiers by learning a multinomial logistic regression on the visual features. We initialize models with respective model weight and attach a linear classifier head initialized from scratch\footnote{We follow ~\cite{goyal2019scaling} which uses a BatchNorm followed by linear layer. We found that this setting leads to robust hyperparameter choice and sweeping hyperparams such as learning rate, weight decay only leads to marginal (+/-0.1) change in performance.} on top. While keeping the model trunk fixed, we train the linear classifier head for $28$ epochs using SGD momentum 0.9, weight decay $0.0005$, learning rate of $0.01$ for a batchsize of $256$ and step learning rate schedule with gamma factor $0.1$ with decay at steps [$8$, $16$, $24$]. For majority of tasks above, we use the same settings and note any differences in Appendix~\ref{sec:appendix_reprensentation_learning}.

\paragraph{Results.}
We report the linear probe numbers for all our models in Appendix Table~\ref{tab:appendix_all_linear_detailed_results} and in Table~\ref{tab:linear_in1k_places_voc_results}, we report the performance of models on three standard tasks commonly used in computer vision. Further, in Figure~\ref{fig:linear_probe_comparison_summary}, we summarize the difference in visual features quality of our best SEER model (RegNet-$10$B) compared to the best ImageNet based supervised and self-supervised performance on respective tasks. We observe that \textit{(i)} For the same model size (RG-128Gf), our model trained on random images in the wild outperforms self-supervised models trained on ImageNet on 17 out of 25 tasks, \textit{(ii)} our best model (RG-10B) also surpassed the best state-of-the-art self-supervised models (any size, approach data, and architecture) on 17 out of 25 tasks and achieves competitive (within 1\% accuracy) on 5 out of 8 tasks, \textit{(iii)} we further note that our best model also surpasses the best supervised (fully supervised or weakly-supervised) models (any size, architecture) on 14 out of 25 tasks and achieves competitive accuracy on the remaining. \textit{(iv)} on tasks in datasets such as medical imaging, satellite images, structured images, OCR, activity recognition in videos, our model consistently outperforms ImageNet models. On the other datasets such as Oxford Pets, Cars etc which are highly object centric, training on object-centric datasets gives better results yet our model achieves competitive performance despite training on random images in wild.

\section{Salient Properties}
Motivated by the use of discriminative self-supervised approach for training on random group of internet images, we also evaluate if the model learns some salient properties present in the images and differentiates between images. Towards this, in Sec.~\ref{sec:geoloc_section} we probe our model for the ability to predicting the GPS coordinates from images taken from all over the world. Further, we also probe the model embeddings space for the ability to embed together similar concepts with variations all over the world (for example, ``wedding'' concept varies culturally across the globe). For this, we qualitatively study the embeddings of hashtags (all languages, regions) in the model space in Sec.~\ref{sec:multilingual_word_cloud}.

\begin{table*}[t]
  \centering
  \begin{adjustbox}{max width=\textwidth}
    \begin{tabular}{@{}lll c ccccc@{}}
      \toprule
      &&&& \multicolumn{5}{c}{\textbf{Accuracy within Distance (km)}} \\
      \cmidrule{5-9} 
      &&&& \textbf{Street} & \textbf{City} & \textbf{Region} & \textbf{Country} & \textbf{Continent} \\ 
     
      \textbf{Model} & \textbf{Data} & \textbf{Arch.} &&  1 km & 25 km & 200 km & 750 km & 2500 km \\
          \midrule
      
      Human        &  &         && --   & --   & $\phantom{0}$3.8 & 13.9 & 39.3 \\
      \midrule
      \multicolumn{9}{@{}l}{\textit{Comparison with specialized models using ImageNet pretraining}} \\
      ISNs         & & --        && 15.6 & 39.2 & 48.9 &  65.8 & 78.5 \\
      ISNs+HSC     & & --        && 15.2 & 40.9 & 51.5 &  65.4 & 78.5 \\
      ISNs+HSSC    & & --        && \textbf{16.9} & 43.0 & 51.9 & 66.7 & 80.2 \\
          CPlaNet      & & --        && 16.5 & 37.1 & 46.4 & 62.0 & 78.5 \\
          Deep-Ret+    & & --        && 14.4 & 33.3 & 47.7 & 61.6 & 73.4 \\
      PlaNet       & & --        && $\phantom{0}$8.4  & 24.5 & 37.6 & 53.6 & 71.3 \\
      \midrule 
      \supervised  & INet-1K & RG-128Gf  && 13.5 &  34.2 & 45.6 & 60.3 & 72.2 \\
      
      \midrule 
      \multicolumn{9}{@{}l}{\textit{Self-supervised pretraining on ImageNet}} \\
      \swav        & INet-1K & RG-128Gf  && 15.6  & 42.6 &  54.9 &  72.2 &  83.5 \\
      
      \midrule
      \multicolumn{9}{@{}l}{\textit{Pretrained on random internet images}} \\
      \seer         & IG-1B & RG-128Gf  && 16.0 & 42.6  & 54.9  & \textbf{73.4} & 83.5 \\
      \seer         & IG-1B & RG-256Gf  && 15.2 & \textbf{43.9} & \textbf{58.3} & \underline{73.0} & \textbf{83.5} \\ 
              \midrule
      
      \\ 
          
      \midrule
      \multicolumn{8}{@{}l}{\textit{Evaluation on Im2GPS3k test set}}\\
      ISNs+HSSC    & & --        && 10.5 & 28.0 & 36.6 & 49.7 & 66.0 \\
      \seer (ours)      & IG-1B   & RG-256Gf  && \textbf{12.6} & \textbf{33.9} & \textbf{45.3} & \textbf{61.0} & \textbf{76.0} \\

      \bottomrule
    \end{tabular}
  \end{adjustbox}
  \caption{
    \textbf{Geo Localization} results of several models on \textit{Im2GPS} \texttt{test} set (top) containing 237 images and \textit{Im2GPS3k} test set (bottom). 
    Metric is the fraction of images localized within the given radius using the GCD distance.
    +HSC = \textit{using hierarchical classification}
    +HSSC = \textit{with hierarchical and scene set classification}. 
  }
  \label{tab:im2gps_results}
\end{table*}

\subsection{Geo Localization}
\label{sec:geoloc_section}

In this task, we are interested in auditing if the model has learned some salient property allowing it to predict the gps coordinates of a given input image.
We do so by coping with the problem of geolocalization \ie predicting the GPS coordinates of images taken from all over the world.
Such images exhibit a wide range of variations, \ie picturing different objects, using different camera settings, taken at different daytime or seasons. 
Moreover, the images provide very few visual clues about the respective GPS location. 
Unlike previous works~\cite{planet2016,workman2015widearea,zamir2010,grant2007}, we neither make prior assumptions on the task nor simplify the problem by restricting the task to images from famous landmarks and cities, natural areas like deserts or mountains.
We want to test if the model works at a global scale without any assumptions on the data or task. 

We follow Muller \etal~\cite{muller2018geoloc} and treat this problem as a classification problem, sub-dividing the Earth into geographical cells.
There are three types of partitionings: coarse, middle and fine, with varying number of cells.
We visually illustrate the difference between those partitionings in Appendix Figure~\ref{fig:partitioning_figure}.
In our experiment, we use the \texttt{fine} partitioning which divides the globe into $12,893$ cells. 

We finetune our model on a subset of YFCC100M~\cite{yfcc2016} introduced for the MediaEval Placing Task MP-$16$~\cite{martha2017planet}.
This subset includes $4,219,225$ geo-tagged images from Flickr\footnote{Available at: \url{https://multimedia-commons.s3-website-us-west-2.amazonaws.com}}.
The dataset contains ambiguous photos of indoor environments, food, and humans for which the location is difficult to predict. 
During finetuning, we validate the performance on a validation set composed of $22,855$ geo-tagged images. 
We finetune for $15$ epochs using SGD wth a momentum of $0.9$, a weight decay of $1e-4$, and a learning rate of $0.05$.
We decay the learning rate by $0.01$ at epochs $[8, 12]$.
Finally, we evaluate the finetuned model on the im2gps~\cite{hays2008im2gps} \texttt{test} set, containing $237$ geo-tagged images. 
We perform inference and record the top predicted cell for each image in the test set. 
The predicted cells are mapped back to a geographical latitude and longitude and the great circle distance (GCD) is computed by comparing to the ground truth latitude/longitude. 

\begin{figure*}[t]
  \centering
  \includegraphics[width=\linewidth]{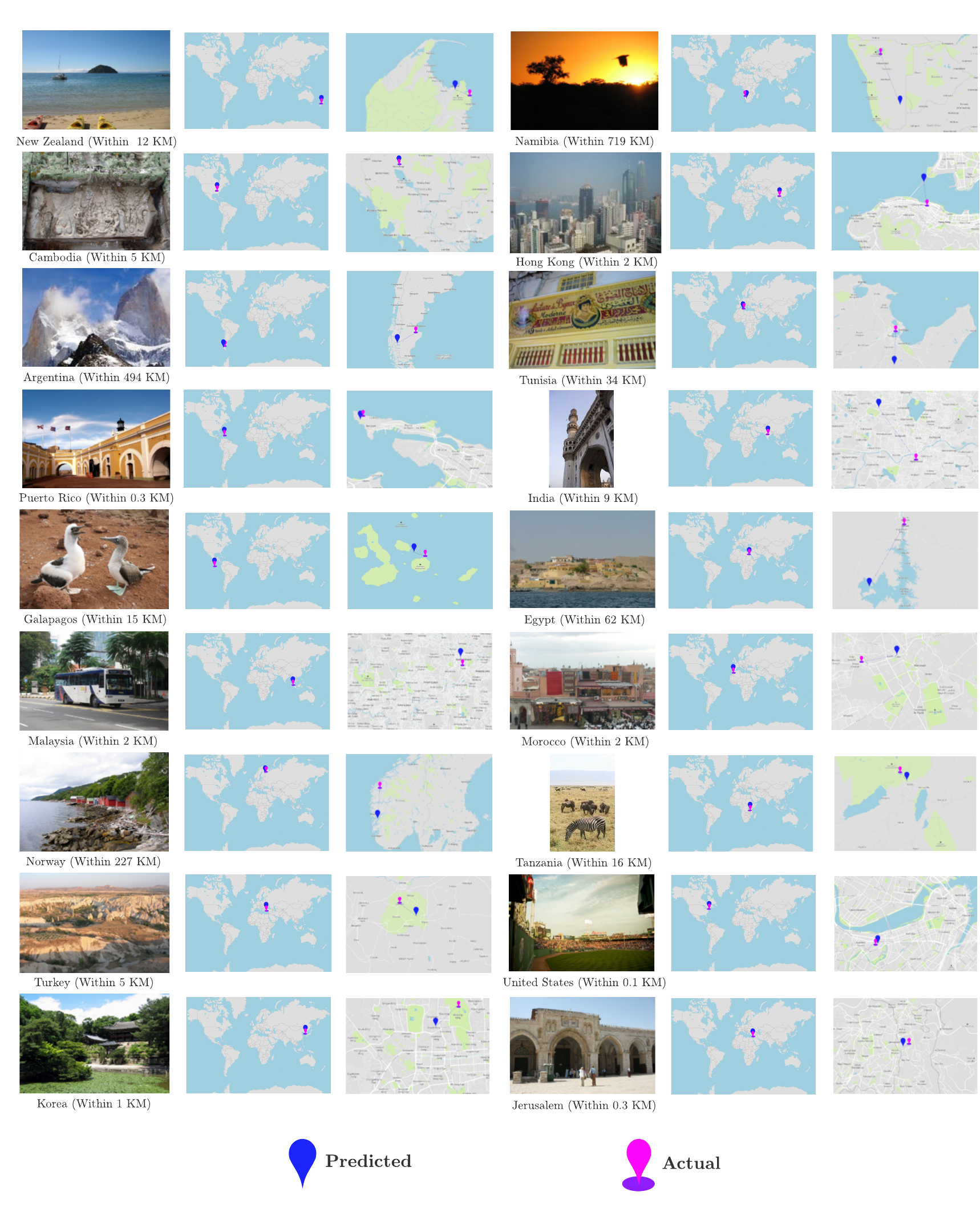}
  \caption{
        Visualization of diverse examples from the im2gps test sets. We chart RegNet-256Gf SEER predictions and targets, and display how many kilometers the SEER model predictions are off from the target.
  }
  \label{fig:im2gps_actual_v_predicted_grid}
\end{figure*}

Following Hays \etal~\cite{hays2008im2gps}, we report accuracy as the percentage of test images that are predicted within a certain distance to the ground-truth location. 
The results are presented in Table~\ref{tab:im2gps_results}. 
SEER models achieve the state-of-the-art geolocalization results for all different distance thresholds.
Moreover, as the size of models increases, geolocalization accuracy improves. 
We show qualitative results for this evaluation in Figure~\ref{fig:im2gps_actual_v_predicted_grid}.

\begin{figure*}[t]
  \centering
  \includegraphics{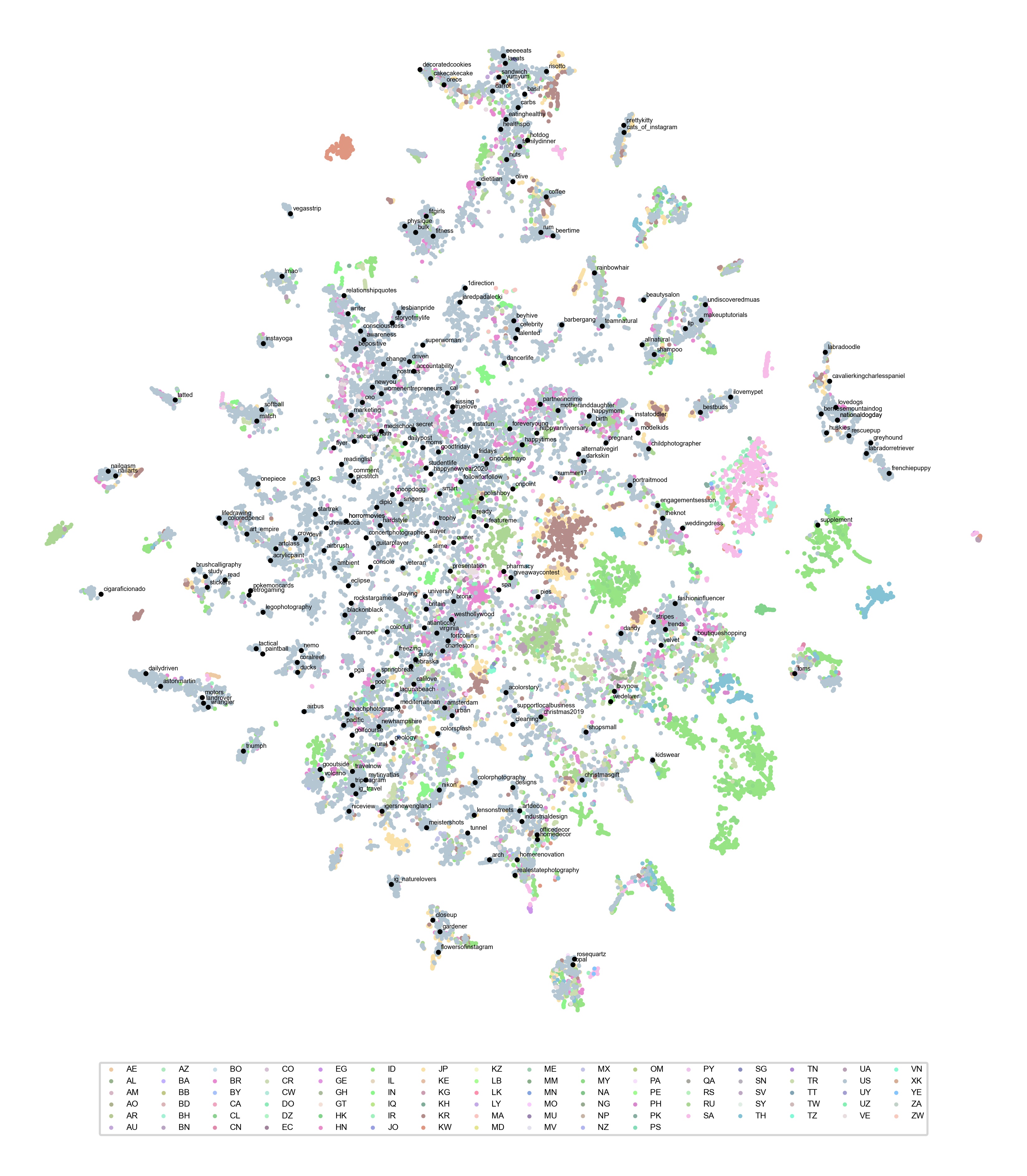}
  \caption{
    \textbf{T-SNE map of Hashtag representations.}
    We color-code hashtags from different countries.
    For easy readability, the hashtag text is provided for a subset of $250$ tags.
    We see clear patterns emerging, for examples the tags about food, sports, animals being grouped together.
  }
  \label{fig:tsne} \end{figure*}

\subsection{Multilingual Hashtag Embeddings}
\label{sec:multilingual_word_cloud}
In this experiment, we want to leverage our image encoder to get some qualitative understanding of our data and test if the model has learned some interesting salient properties.
Also, since our network is self-supervised, it can be effectively used for that purpose.
Indeed, our features show good representation properties without heavy finetuning.
The image encoder can directly be used as a proxy metric for the metadata associated with images.
Since we work on Instagram data, we propose to study whether our model allows to properly represent hashtags associated with the images. 

In order to construct hashtag embeddings, we took a random subsample of $6,000,480$ images with their associated hashtags.
We sorted hashtags by their frequency, and kept the $30,000$ most frequent ones.
The most frequent one appeared $220,345$ times and least frequent one $183$ times.
For each hashtag, we retrieved the set of images it is associated to, and computed their $256$-dimensional features / model embeddings using the \seer RG-128Gf model.
We represent the hashtag by simply taking the average of those features.

Given the hashtag features, we reduce the dimension to $50$ using PCA.
We represent the hashtags in a $2$D plane by computing a t-sne map~\cite{van2008visualizing}.
We use the scikit-learn~\cite{scikit-learn} implementation, with a perplexity of $40.0$, a learning rate of $100.0$ and running for $5000$ iterations.

Given that we have geo-diverse data i.e. our pretraining data represents images from all over the world as shown in Figure~\ref{fig:ig10m_data_distribution}, we represent this diversity by color-coding the hashtag features to represent the countries.
For each hashtag, we associate it with a country by taking a majority vote across images associated with that tag.
Because of the predominance of US-based data (see Figure~\ref{fig:ig10m_data_distribution}), this vote-based method leads to $14,962$ hashtags associated with the United States.
Nonetheless, more than half of the data is associated with other countries and we obtain a wide coverage, with hashtags from $91$ countries being represented.
We represent the hashtag embeddings in Figure~\ref{fig:tsne}.
For readability, we present the actual text associated with the features for $250$ US-based tags.

We observe that our model indeed embeds together the hashtags in different languages but corresponding to same concept from all over the world. For instance, the concept wedding has hashtags: ``shaadi'' (Indian wedding), ``nikah'', ``bridesmaid'' etc all embedded in close proximity and likewise many other multilingual clusters appear. Further, the clusters are fine-grained for example: within the concept ``wedding'' sub-clusters appear like one for the wedding photoshoot, wedding dress, wedding design/styles etc.

\section{Conclusion}
In this work, we have demonstrated the potential of using self-supervised training on random internet images to train models that are more fair and less harmful (less harmful predictions, improved and less disparate learned attribute representations and larger improvement in object recognition on images from low/medium income households and non-Western countries). We train a 10B parameters dense model and observe that fairness indicator results improve as model size increases. We also observe better robustness to distribution shift, SOTA image copy detection and new metadata information captured by model such as gps prediction and multilingual word embeddings. The model also captures semantic information better and outperforms SOTA models (supervised and self-supervised) trained on ImageNet on 20 out 25 image classification tasks in computer vision while achieving competitive performance on the rest. \\

{\small{\noindent\textbf{Acknowledgement}:
We would like to thank Laurens Van Der Maaten, Matthijs Douze, Matthew Muckley, Piotr Dollar, Mannat Singh for helpful discussions and feedback, and Min Xu, Giri Anantharaman, Myle Ott, Vittorio Caggiano for their help with FSDP for our model training. We are grateful to Lei Tian, Wenyin Fu, Sachin Lakharia and Richard Huang for their help in optimizing training speed and reliability.}}

\begin{figure}[t]
    \centering
    \includegraphics[width=\linewidth]{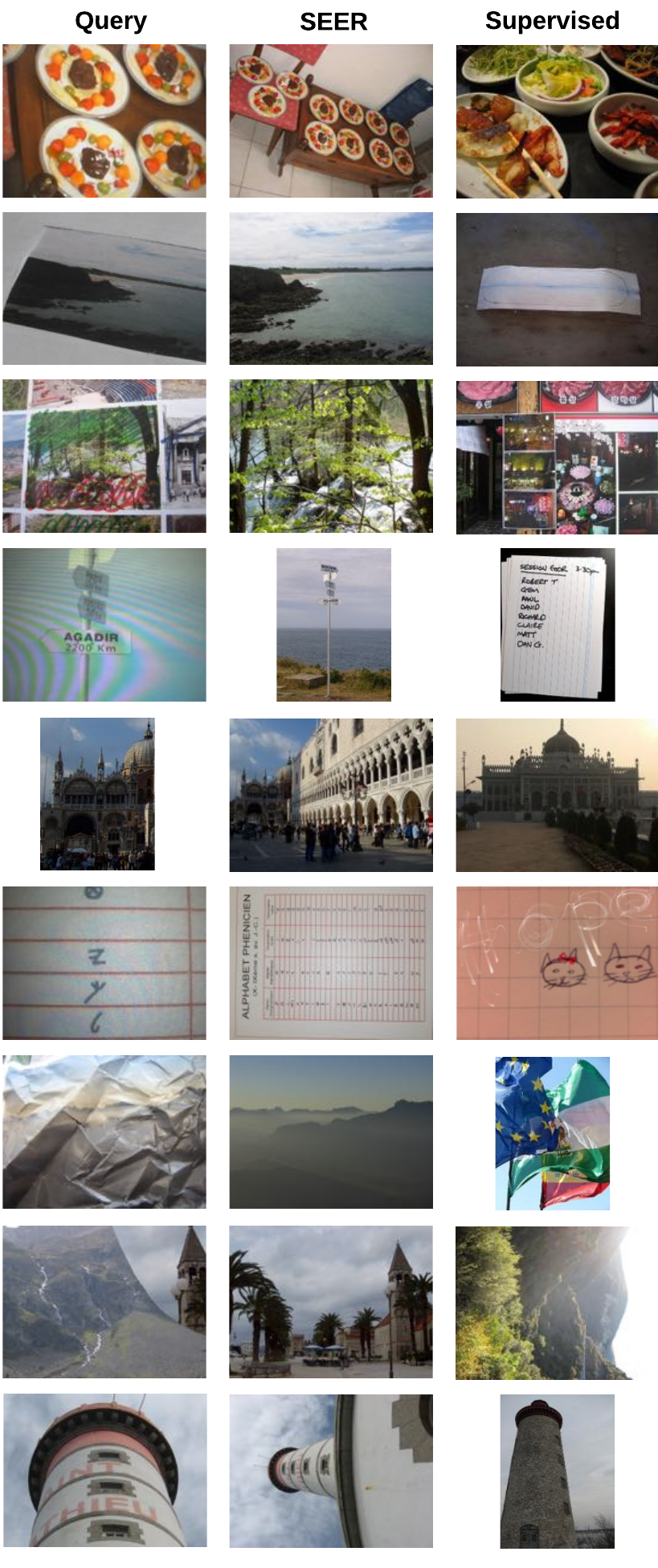}
    \caption{
    \textbf{Qualitative results of Copy Detection} results on ``strong'' subset of Copydays as described in Sec.~\ref{sec:copy_detection_results}. The objective is to find the original image based on a copy of that image. Shown are queries in which the SEER model finds the correct image, while the supervised model fails to do so. We define correct as the original image being ranked first among all 10,157 database and distractor images.
    }
    \label{fig:copydays_viz_seer_v_supervised_appendix} 
\end{figure}

{\small
\bibliographystyle{ieee_fullname}
\bibliography{egbib}
}

\clearpage

\section{Appendix}

\appendix

\section{Licence Information for Photos in the Paper}

\newcommand{\imagelicence}[3]{{\underline{"#1" #2 - Photo}: #3}}

\noindent The photos in Figure \ref{fig:front_page} in section "Geographical diversity" (in order from left to right, top to bottom) are from dollar street, and have the following licences:

{\small
\begin{itemize}
    \item \imagelicence{Kitchens}{South Korea}{Luc Forsyth (CC BY 4.0)}
    \item \imagelicence{Armchairs}{Romania}{Catalin Georgescu (CC BY 4.0)}
    \item \imagelicence{Everyday Shoes}{Bulgaria}{Boryana Katsarova (CC BY 4.0)}
    \item \imagelicence{Spices}{India}{AJ Sharma (CC BY 4.0)}
    \item \imagelicence{Necklaces}{Pakistan}{Hisham Najam (CC BY 4.0)}
    \item \imagelicence{Bathrooms}{Kenya}{Chris Dade (CC BY 4.0)}
\end{itemize}
}
\noindent The photos in Figure \ref{fig:dollar_street_samples} (in order left to right, top to bottom) are from dollar street, and have the following licences:

{\small
\begin{itemize}
  \item \imagelicence{Street View}{UK}{Jeny Garcia (CC BY 4.0)}
  \item \imagelicence{Street View}{Bolivia}{Zoriah Miller (CC BY 4.0)}
  \item \imagelicence{Street View}{Burundi}{Johan Eriksson (CC BY 4.0)}
  \item \imagelicence{Street View}{Brazil}{Leony Carvalho (CC BY 4.0)}
  \item \imagelicence{Street View}{India}{Zoriah Miller (CC BY 4.0)}
  \item \imagelicence{Street View}{Haiti}{Zoriah Miller (CC BY 4.0)}
  \item \imagelicence{Street View}{India}{Zoriah Miller (CC BY 4.0)}
  \item \imagelicence{Street View}{Philippines}{Luc Forsyth (CC BY 4.0)}
  
  \item \imagelicence{Spices}{India}{AJ Sharma (CC BY 4.0)}
  \item \imagelicence{Spices}{Nigeria}{Adeola Olagunju (CC BY 4.0)}
  \item \imagelicence{Spices}{Netherlands}{Global Exploration (CC BY 4.0)}
  \item \imagelicence{Spices}{UK}{Chris Dade (CC BY 4.0)}
  \item \imagelicence{Spices}{USA}{Elizabeth Barentine (CC BY 4.0)}
  \item \imagelicence{Spices}{India}{Zoriah Miller (CC BY 4.0)}
  \item \imagelicence{Spices}{India}{Kunal Apastamb (CC BY 4.0)}
  \item \imagelicence{Spices}{India}{Vanshika Sharma (CC BY 4.0)}
  
    \item \imagelicence{Everyday Shoes}{Jordan}{Zoriah Miller (CC BY 4.0)}
    \item \imagelicence{Everyday Shoes}{India}{Abhineet Malhotra (CC BY 4.0)}
    \item \imagelicence{Everyday Shoes}{China}{Jonathan Taylor (CC BY 4.0)}
    \item \imagelicence{Everyday Shoes}{Brazil}{Moises Morero (CC BY 4.0)}
    \item \imagelicence{Everyday Shoes}{Bulgaria}{Boryana Katsarova (CC BY 4.0)}
    \item \imagelicence{Everyday Shoes}{UK}{Chris Dade (CC BY 4.0)}
    \item \imagelicence{Everyday Shoes}{China}{Jonathan Taylor (CC BY 4.0)}
    \item \imagelicence{Everyday Shoes}{Kenya}{Johan Selin (CC BY 4.0)}
    
    \item \imagelicence{Necklaces}{Pakistan}{Hisham Najam (CC BY 4.0)}
    \item \imagelicence{Necklaces}{USA}{Elizabeth Barentine (CC BY 4.0)}
    \item \imagelicence{Necklaces}{India}{Akshay Jain (CC BY 4.0)}
    \item \imagelicence{Necklaces}{USA}{Isaiah Williams (CC BY 4.0)}
    \item \imagelicence{Necklaces}{Netherlands}{Global Exploration (CC BY 4.0)}
    \item \imagelicence{Necklaces}{Serbia}{Darko Rajkovic (CC BY 4.0)}
    \item \imagelicence{Necklaces}{India}{Kunal Apastamb (CC BY 4.0)}
    \item \imagelicence{Necklaces}{Romania}{Catalin Georgescu (CC BY 4.0)}

    \item \imagelicence{Armchairs}{USA}{Sarah Diamond (CC BY 4.0)}
    \item \imagelicence{Armchairs}{Cote d'Ivoire}{Zoriah Miller (CC BY 4.0)}
    \item \imagelicence{Armchairs}{Romania}{Catalin Georgescu (CC BY 4.0)}
    \item \imagelicence{Armchairs}{Vietnam}{Victrixia Montes (CC BY 4.0)}
    \item \imagelicence{Armchairs}{Kyrgyzstan}{Svetlana Lebedeva  (CC BY 4.0)}
    \item \imagelicence{Armchairs}{China}{Jonathan Taylor (CC BY 4.0)}
    \item \imagelicence{Armchairs}{Colombia}{Zoriah Miller (CC BY 4.0)}
    \item \imagelicence{Armchairs}{Nigeria}{Johan Eriksson (CC BY 4.0)}

    \item \imagelicence{Stoves}{India}{Preksha Panchamia (CC BY 4.0)}
    \item \imagelicence{Stoves}{Palestine}{Eman Jomaa (CC BY 4.0)}
    \item \imagelicence{Stoves}{Latvia}{Konstatins Sigulis (CC BY 4.0)}
    \item \imagelicence{Stoves}{India}{Akshay Jain (CC BY 4.0)}
    \item \imagelicence{Stoves}{Nepal}{Luc Forsyth (CC BY 4.0)}
    \item \imagelicence{Stoves}{China}{Jonathan Taylor (CC BY 4.0)}
    \item \imagelicence{Stoves}{Nepal}{Luc Forsyth (CC BY 4.0)}
    \item \imagelicence{Stoves}{Nepal}{Luc Forsyth (CC BY 4.0)}
\end{itemize}
}

\section{Evaluation Datasets}
We evaluate performance of SEER models on several tasks in computer vision. The list of tasks and the data distribution is presented in Table~\ref{tab:appendix_tasks_list}.

\begin{table*}[t]
  \centering
  
  \begin{tabular}{@{}l l c c c @{}}
  
  \toprule
    \textbf{Task Type} & \textbf{Dataset} & \textbf{Train size} & \textbf{Test size} & \textbf{Classes}  \\

    \midrule
    Standard & ImageNet-1K & 1281167 & 50000 & 1000 \\
    Standard & Places205 & 2448862  & 20500 & 205 \\
    Standard & PASCAL VOC07 & 5011 & 4952 & 20 \\
    Standard & Oxford-IIT Pets & 3680 & 3669 & 37 \\
    Standard & Oxford Flowers & 2040 & 6149 & 102 \\
    Standard & Caltech-101 & 3060 & 6085 & 102 \\
    
    \midrule
    Medical & PatchCamelyon & 262144 & 32768 & 2 \\
    Satellite & RESISC45 & 25200 & 6300 & 45 \\
    Satellite & EuroSAT & 10000 & 5000 & 10 \\

    \midrule
    Structured & CLEVR distance & 70000 & 15000 & 6 \\
    Structured & CLEVR counting & 70000 & 15000 & 8 \\
    Structured & Small Norb elevation & 24300 & 24300 & 9 \\
    Structured & dSprites Orientation & 589824 & 147456 & 16 \\
    Structured & dSprites Location & 589824 & 147456 & 16 \\
    
    \midrule
    Videos Activity Recognition & UCF-101 & 9537 & 3783 & 101 \\
    Videos Activity Recognition & Kinetics-700 & 536485 & 33966 & 700 \\
    
    \midrule
    Scene Recognition & SUN397 & 76129 & 21758 & 397 \\
    
    \midrule
    Self-driving & GTSRB & 26683 & 12630 & 43 \\
    Self-driving & KITTI-Distance & 5985 & 1496 & 4 \\
    
    \midrule
    textures & DTD & 1880 & 1880 & 47 \\
    
    \midrule
    OCR & SVHN & 73257 & 26032 & 10 \\

  \bottomrule
  \end{tabular}
  \caption{\textbf{List of downstream image classification datasets} with the data distribution and the type of task that we evaluate our models on.}
  \label{tab:appendix_tasks_list}
\end{table*}

\begin{table*}[t]
  \centering
  \begin{adjustbox}{max width=\textwidth}
    \begin{tabular}{@{}lccccc@{}}
      \toprule
      \textbf{Model} & \textbf{depth} & \textbf{group width} & \textbf{layer widths} & \textbf{resolution} & \textbf{FLOPs} \\
      \midrule
      \makecell[l]{\textbf{base model}: RegNetY-128gf} & [2, 7, 17, 1] & [264, 264, 264, 264] & [528, 1056, 2904, 7392] & 224 &  1.28E+11 \\
      \midrule
      \makecell[l]{\textbf{Variant1}: Narrow width + Deeper (alpha0.75)} & [3, 9, 23, 1] & [232, 232, 232, 232] & [464, 928, 2552, 6496] & 224 & 1.30E+11 \\
      \makecell[l]{\textbf{Variant2}: Narrow width + Deeper (alpha0.85} & [2, 8, 20, 1] & [240, 240, 240, 240] & [480, 960, 2640, 6720] & 224 & 1.20E+11  \\
      \midrule
      \makecell[l]{\textbf{Variant3}: Narrow width + HiRes} &  [2, 7, 17, 1] & [232, 232, 232, 232] & [464, 928, 2552, 6496] & 260 & 1.43E+11 \\
      \makecell[l]{\textbf{Variant4}: Narrow width + HiRes + Deeper} &  [2, 8, 20, 1] & [232, 232, 232, 232] & [464, 928, 2552, 6496] & 240 & 1.29E+11 \\
      \midrule
      \makecell[l]{\textbf{Variant5}: Wider + Deeper (alpha1.25)} & [2, 8, 19, 1] & [280, 280, 280, 280] & [560, 1120, 3080, 7840] & 224 & 1.58E+11  \\
      \midrule
      \makecell[l]{\textbf{Variant6}: RegNetZ-4gf (dWr scaling)} & [3, 8, 22, 3] &  [128, 128, 128, 128] & [384, 768, 2048, 4864] & 284 & 1.30E+11 \\
      \bottomrule
    \end{tabular}
  \end{adjustbox}
  \caption{
    Model size scaling dimensions and variants explored for scaling architecture to 10B parameters. 
    We chose a RegNetY-128gf model with 700M params as a base model and generated \textit{6 variants}.
  }
  \label{tab:model_scaling_variations}
\end{table*}

\begin{table}[t]
  \centering
  \begin{tabular}{@{}ll@{}}
  \toprule
    \textbf{Hyperparameter} & \textbf{Value} \\
    \midrule
    Batch size & 7936 \\
    Crops & 2x160+4x96 \\
    Head & \makecell[l]{[28280, 8192, 8192, 256] \\ (no BatchNorm)}  \\
    Training epochs & 1 \\
    Training Images & 1 Billion \\
    loss sinkhorn iterations & 10 \\
    loss epsilon & 0.03 \\
    loss temperature & 0.1 \\
    Weight decay & 1e-5 \\
    Warm-up iterations &  5500 \\
    SGD momentum & 0.9 \\
    SyncBatchNorm & yes  \\
  \bottomrule
  \end{tabular}
  \caption{\textbf{Hyperaprams of SEER 10B params} model pretraining.}
  \label{tab:seer_model_hyperparams}
\end{table}

\begin{table*}[t]
  \centering
  
  \begin{tabular}{@{}cccccccccc@{}}
  \toprule
Model & Arch. & Param. & w\_0 & w\_a & w\_m & depth & group width & Head & blocks \\
\thinline
SEER & RG-8gf & 42M &  192 &  76.82 &  2.19 &  27 &  56 & [2016,4096,4096,256] & (2, 7, 17, 1) \\
SEER & RG-16gf & 91M &  200 &  160.23 &  2.48 &  27 & 112 & [3024,4096,4096,256] & (2, 7, 17, 1) \\
SEER & RG-32gf & 156M &  232 &  115.89 &  2.53 &  27 & 232 &  [3712,4096,4096,256] & (2, 7, 17, 1) \\
SEER & RG-64gf & 250M &  352 &  147.48 &  2.4 &  27 & 328 &  [4920, 8192, 8192, 256] & (2, 7, 17, 1) \\
SEER & RG-128gf & 693M &  456 &  160.83 &  2.52 &  27 & 264 & [7392, 8192, 8192, 256] & (2, 7, 17, 1) \\
SEER & RG-256gf & 1.5B &  640 &  230.83 &  2.53 & 27 &  373 & [10444, 8192, 8192, 256] & (2, 7, 17, 1) \\
SEER & RG-10B & 10B &  1744 &  620.83 &  2.52 &  27 & 1010 & [28280, 8192, 8192, 256] & (2, 7, 17, 1) \\
  \bottomrule
  \end{tabular}
  
  \caption{Configuration of \textbf{all SEER models} with number of parameters varying from $40$ million to $10$ billion.}
  \label{tab:seer_model_variants}
\end{table*}

\section{SEER model architecture and training Hyperparams}
\label{sec:appendix_models_seer}

In Table~\ref{tab:model_scaling_variations}, we share in detail all the model architecture variants we explore to scale the SEER model size to $10$billion parameters. In Table~\ref{tab:seer_model_variants}, we summarize all the different sizes of SEER model and the configurations. For the 10Billion parameters SEER model, we describe the pretraining hyperparams in Table~\ref{tab:seer_model_hyperparams}.

\section{Adapting LARC implementation for FSDP training}
\label{sec:appendix_larc_adaptation}

LARC~\cite{you2017large} scales the learning rate of each layer $l$ based on the norm of the parameters $w^l$, the norm of the gradient of the parameters $\nabla w^l$, the weight decay $\beta$ and a trust coefficient $\eta$, following the formula:

\begin{equation}
  \lambda^l = \eta \frac{||w^l||}{|| \nabla w^l || + ||w^l|| * \beta}
\end{equation}

When training with FSDP, parameters and their gradients are sharded across GPUs. To avoid adding additional parameter consolidation across GPUs to compute the norms, we adapt the implementation of LARC to compute a distributed norm without exchanging the weights, by decomposing the computation of the norm into a sum of squares and a square root.

The sum of square can be computed on each shard separately and then all-reduced before taking the square root of the resulting sum. We also batch the all-reduce of all the layers together. As a result, enabling LARC only incurs the overhead of one single all-reduce on a tensor of size $2 L$ where $L$ is the number of layers of our model.

\section{Model State Dictionary checkpointing}
\label{sec:appendix_model_state_dict}

As typically done during training, we save our model state checkpoints allowing us to restart the training upon interruption as well as evaluate the model at intermediate training stages. For small models, this is typically done by saving the model weights and optimizer state in one file, in addition to information about the current training step and learning rate scheduler data.

For the 10 billion model, trained with FSDP, saving one checkpoint containing the model and optimizer state would require to consolidate the model and optimizer states, sharded across multiple GPUs during training, on a single GPU. This is impractical for memory reasons (it would account for 80GB of memory for FP32 weights) as well as communication reasons (consolidating mean communicating weights across GPUs).

Instead, for our 10 billion model, each GPU saves its own shard of the model weights and optimizer state, along with some metadata allowing to re-consolidate or re-shard the checkpoints offline. In addition, rank 0 GPU will save additional metadata allowing to locate the checkpoint shards.

During training, and after an interruption, we use the checkpoint shards to reload the model. Since training is resumed on the same number of GPU as the number number of shards, each GPU can simply load the shard corresponding to its rank and restart from there. This design allows to naturally exploit parallelism during model checkpointing and model reload. In practice, it makes state checkpointing very fast for our 10 billion model.

For evaluations such as linear evaluation, it is impractical to use as many GPUs as were used during training, so we cannot rely on the same mechanism to load our model. Instead, we transform the sharded checkpoints into evaluation checkpoints that are structured in such a way that they can be loaded by a model sharded on a arbitrary number of GPUs.

\begin{figure}[t]
    \centering
    \includegraphics[width=0.9\linewidth]{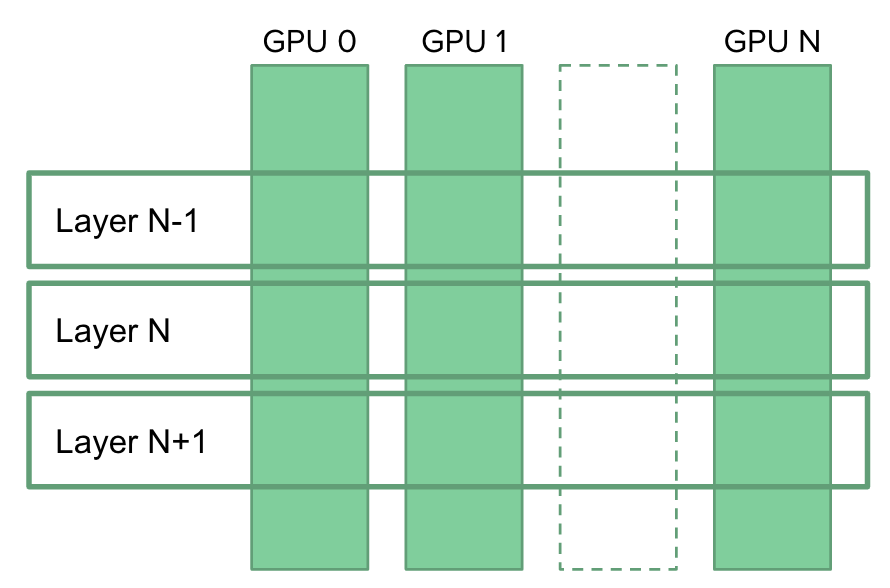}
    \caption{
      The two types of state checkpoints we use for our 10 billion model. For training, \texttt{sharded} checkpoints (weights sharded by GPU, vertical strips in the diagram) are used for efficiency. For evaluation, \texttt{sliced} checkpoints (each slice containing a layer, horizontal strips in the diagram) are used to be agnostic to the number of GPU. A script transforms our vertically split checkpoints to horizontally split checkpoints between training and evaluation.
    }
    \label{fig:fsdp_state_resharding} 
\end{figure}

 To do so, we transform our checkpoint shards into a \texttt{sliced} checkpoint, where each slice of the checkpoint consists of the full weights of a given layer of the model. Additional metadata is saved in order to keep track of all the slices. This design, which avoids the memory issues of saving and loading a single consolidated checkpoint, is illustrated in Figure-\ref{fig:fsdp_state_resharding}.

To load such a sliced checkpoint for evaluation, we build on the usual mechanism of \texttt{FSDP}, which consolidates layers one by one before each forward, only instead of calling forward, we initialize the weights.

\section{Representation Learning: Additional Results}
\label{sec:appendix_reprensentation_learning}

In this section, we provide additional results and details for Sec.~\ref{sec:summarize_representation_learning_main}. We also provide results on low-shot learning.

\subsection{Linear Probe: complete Results}
We present detailed results for our model and baseline models~\ref{sec:models_baseline} on image classification datasets  via linear probe in Table~\ref{tab:appendix_all_linear_detailed_results}.
\begin{table*}[t]
\centering
\setlength{\tabcolsep}{0.2em} \scalebox{0.63} {
\begin{tabular}{ccccc|ccccccccccccccccccccccccccccccc}
    & & & & &  
    \begin{turn}{90}CLEVR Count\end{turn} & 
    \begin{turn}{90}CLEVR Dist\end{turn} & 

    \begin{turn}{90}Food101\end{turn} & 
    \begin{turn}{90}Caltech101\end{turn} & 
    \begin{turn}{90}EuroSAT\end{turn} &
    \begin{turn}{90}DTD (textures)\end{turn} & 
    \begin{turn}{90}RESISC45\end{turn} & 
    \begin{turn}{90}SVHN\end{turn}  & 
    \begin{turn}{90}GTSRB\end{turn} & 
    \begin{turn}{90}dSprites Orient\end{turn} & 
    \begin{turn}{90}dSprites Loc\end{turn} & 
    \begin{turn}{90}Snorb elevation\end{turn} & 
    \begin{turn}{90}PatchCam\end{turn} & 
    \begin{turn}{90}Oxford Pets\end{turn} &
    \begin{turn}{90}Stanford Cars\end{turn} & 

    \begin{turn}{90}CIFAR10 \end{turn} & 
    \begin{turn}{90}CIFAR100\end{turn} & 
    \begin{turn}{90}FGVC Aircraft\end{turn} & 
    \begin{turn}{90}STL10\end{turn} &
    \begin{turn}{90}Oxford Flowers\end{turn} & 

    \begin{turn}{90}SUN397\end{turn} & 
    \begin{turn}{90}KITTI Distance\end{turn} & 

    \begin{turn}{90}MNIST\end{turn} & 
     
    \begin{turn}{90}UCF101\end{turn} & 
    \begin{turn}{90}Kinetics700\end{turn} & 
    
    \\

\hline
\multirow{8}{*}{\rotatebox{90}{SEER RG}}
& & & RG-8gf & & 86.06 & 75.5 & 76.19 & 88.67 & 96.6 & 73.88 & 93.42 & 85.6 & 87.43 & 64.35 & 95.32 & 53.23 & 84.22 & 76.04 & 53.62 & 82.7 & 71.14 & 42.0 & 92.53 & 92.75 & 71.03 & 73.6 & 99.28 & 79.13 & 42.31 \\
& & & RG-16gf & & 84.03 & 76.13 & 78.90 & 89.04 & 96.46 & 74.20 & 93.77 & 85.28 & 88.46 & 66.63 & 95.19 & 55.18 & 83.35 & 78.05 & 55.31 & 83.81 & 72.86 & 41.94 & 92.85 & 93.84 & 72.78 & 73.7 & 99.29 & 79.81 & 44.2 \\
& & & RG-32gf & & 84.49 & 75.49 & 80.32 & 89.83 & 97.08 & 77.18 & 93.39 & 84.66 & 87.81 & 70.04 & 95.51 & 53.76 & 85.02 & 78.0 & 58.21 & 84.24 & 74.57 & 44.31 & 94.03 & 93.95 & 72.74 & 74.67 & 99.45 & 81.53 & 46.36 \\
& & & RG-64gf & & 87.24 & 73.41 & 82.53 & 90.81 & 97.1 & 77.55 & 93.88 & 83.77 & 88.97 & 74.71 & 97.03 & 54.77 & 84.36 & 78.11 & 61.52 & 85.5 & 75.73 & 45.44 & 94.78 & 94.65 & 74.57 & 74.2 & 99.32 & 82.03 & 47.86 \\
& & & RG-128gf & & 87.98 & 72.67 & 84.57 & 91.10 & 97.52 & 78.62 & 94.34 & 84.14 & 89.61 & 77.43 & 95.45 & 55.15 & 87.15 & 79.71 & 64.51 & 86.83 & 77.39 & 47.96 & 95.24 & 95.27 & 76.28 & 74.33 & 99.3 & 81.95 & 49.05 \\
& & & RG-256gf & & 89.12 & 72.69 & 87.67 & 91.11 & 97.6 & 80.69 & 94.8 & 81.42 & 89.16 & 77.25 & 95.61 & 56.56 & 87.73 & 80.66 & 66.58 & 87.8 & 78.95 & 48.77 & 96.99 & 96.08 & 78.12 & 75.40 & 99.44 & 88.03 & 51.9 \\
& & & RG-10B & & 89.28 & 74.98 & 90.3 & 91.0 & 97.5 & 81.1 & 95.61 & 86.4 & 90.71 & 81.90 & 96.26 & 58.03 & -- & 85.3 & 68.03 & 90.0 & 81.53 & 54.82 & 97.3 & 96.3 & 80.0 & 78.34 & 99.42 & 82.4 & 51.13 \\
\hline

\multirow{3}{*}{\rotatebox{90}{SwAV}}
& & & RN50 & & 80.7 & 71.26 & 77.88 & 89.27 & 96.72 & 68.72 & 92.61 & 80.79 & 91.74 & 73.30 & 92.02 & 41.87 & 85.79 & 88.89 & 62.65 & 86.38 & 74.23 & 45.44 & 96.46 & 93.12 & 73.53 & 70.32 & 99.24 & 78.49 & 39.31 \\
& & & RG-128gf & & 83.64 & 69.85 & 82.22 & 89.18 & 97.26 & 76.22 & 94.32 & 74.05 & 91.11 & 75.92 & 95.68 & 48.87 & 85.00 & 89.05 & 68.79 & 89.85 & 79.48 & 45.95 & 97.48 & 95.5 & 76.11 & 75.67 & 99.24 & 80.81 & 41.46 \\
& & & RN50-w5 & & 81.87 & 69.67 & 82.05 & 88.57 & 97.48 & 75.90 & 94.73 & 74.42 & 91.7 & 82.28 & 91.74 & 46.83 & 85.11 & 81.17 & 70.18 & 88.79 & 78.28 & 45.98 & 97.41 & 94.72 & -- & 77.67 & 99.25 & 79.26 & 42.32 \\
\hline

\multirow{5}{*}{\rotatebox{90}{DINO}}
& & & RN50 & &81.79 & 70.18 & 78.68 & 89.47 & 96.88 & 70.16 & 93.48 & 79.41 & 91.23 & 68.59 & 89.35 & 41.64 & 84.12 & 89.98 & 60.64 & 85.42 & 75.21 & 47.66 & 96.49 & 94.2 & 73.39 & 71.73 & 99.26 & 78.70 & 39.69 \\
& & & DeiT-S/16 & & 52.63 & 53.12 & 80.95 & 89.60 & 96.98 & 73.78 & 93.05 & 70.37 & 83.73 & 29.48 & 23.72 & 39.44 & 85.47 & 92.21 & 59.83 & 87.17 & 78.69 & 46.83 & 97.45 & 94.52 & 72.86 & 68.32 & 96.94 & 79.36 & 40.02 \\
& & & DeiT-S/8 & & 55.08 & 53.35 & 83.67 & 90.61 & 96.52 & 73.67 & 92.37 & 72.03 & 86.53 & 29.56 & 23.29 & 36.69 & 86.32 & 93.95 & 67.73 & 89.19 & 81.21 & 53.36 & 98.1 & 95.48 & 74.67 & 68.18 & 97.61 & 83.25 & 42.43 \\
& & & DeiT-B/16 & & 54.06 & 53.25 & 82.96 & 90.19 & 97.3 & 74.79 & 93.97 & 70.53 & 87.27 & 31.06 & 25.22 & 40.76 & 85.50 & 93.93 & 71.51 & 89.01 & 82.2 & 52.1 & 98.3 & 96.72 & 74.15 & 69.65 & 97.6 & 82.69 & 42.5 \\
& & & DeiT-B/8 & & 55.15 & 53.29 & 85.96 & 90.18 & 97.4 & 76.54 & 93.67 & 71.97 & 87.57 & 32.58 & 24.45 & 40.21 & 86.87 & 92.67 & 75.17 & 91.15 & 82.9 & 57.79 & 98.73 & 97.59 & 75.6 & 71.79 & 97.87 & 82.8 & 44.67 \\
\hline

\multirow{3}{*}{\rotatebox{90}{MoCo}}
& & & v1-RN50~\cite{radford2021learning} & & 54.7 & -- & 65.9 & 78.1 & 97.1 & 70.0 & 82.9 & -- & 62.6 & -- & -- & -- & 85.7 & 70.4 & 35.9 & 85.0 & 63.1 & 43.5 & 85.6 & 85.4 & 52.6 & 60.2 & 97.6 & 64.2 & 40.7 \\
& & & v2-RN50~\cite{radford2021learning} & & 56.9 & -- & 72.2 & 89.9 & 97.2 & 75.1 & 85.4 & -- & 75.7 & -- & -- & -- & 85.6 & 84.4 & 48.3 & 93.4 & 76.3 & 51.1 & 95.7 & 90.7 & 60.2 & 75.4 & 98.4 & 72.7 & 47.8 \\
& & & v3-ViT-B/16 & & 57.03 & 56.03 & 81.09 & 90.44 & 96.90 & 73.09 & 93.35 & 73.76 & 84.98 & 30.77 & 25.51 & 44.63 & 86.74 & 91.94 & 63.01 & 90.67 & 82.52 & 44.04 & 97.89 & 94.44 & 73.60 & 70.25 & 97.69 & 79.41 & 41.17 \\
\hline

\multirow{2}{*}{{BYOL}}
& & & RN50~\cite{radford2021learning} & & 56.1 & -- & 74.0 & 93.7 & 97.6 & 77.0 & 88.2 & -- & 80.1 & -- & -- & -- & 84.8 & 88.3 & 61.6 & 93.6 & 79.1 & 62.3 & 96.4 & 94.3 & 63.7 & 71.4 & 98.7 & 77.3 & 49.3 \\
& & & RN200w2 & & 78.74 & 68.42 & 77.03 & 90.80 & 96.34 & 74.68 & 92.53 & 75.95 & 87.17 & 78.56 & 94.64 & 38.74 & 84.25 & 91.91 & 66.04 & 91.62 & 81.19 & 45.59 & 97.73 & 92.34 & 72.98 & 73.53 & 98.97 & 80.37 & 40.56 \\
\hline

\multirow{3}{*}{{SimCLRv2}}
& & & RN50~\cite{radford2021learning} & & 56.2 & -- & 76.4 & 91.8 & 97.5 & 77.0 & 85.8 & -- & 71.1 & -- & -- & -- & 84.8 & 88.3 & 56.3 & 93.2 & 77.9 & 51.7 & 96.7 & 92.9 & 64.1 & 69.1 & 97.6 & 78.4 & 51.0 \\
& & & RN101~\cite{radford2021learning} & & 53.6 & -- & 77.9 & 91.6 & 96.8 & 77.2 & 84.6 & -- & 65.7 & -- & -- & -- & 84.3 & 90.0 & 57.1 & 94.8 & 79.9 & 52.0 & 97.6 & 92.7 & 65.2 & 70.6 & 97.2 & 78.8 & 52.4 \\
& & & RN152w3+SK & & 56.57 & 48.76 & 75.0 & 87.63 & 94.32 & 70.32 & 89.77 & 55.37 & 80.07 & 50.12 & 64.58 & 41.25 & 81.53 & 88.15 & 60.75 & 68.86 & 59.23 & 41.22 & 93.45 & 91.87 & 67.31 & 70.66 & 95.74 & 70.8 & 38.83 \\
\hline

\multirow{8}{*}{\rotatebox{90}{Imagenet Supervised}}
& & & RN50 & & 65.06 & 74.13 & 72.96 & 88.52 & 94.98 & 68.09 & 88.56 & 89.75 & 92.29 & 71.83 & 90.61 & 39.58 & 83.42 & 92.21 & 64.22 & 87.49 & 75.43 & 44.04 & 95.42 & 93.40 & 67.55 & 68.38 & 98.95 & 73.26 & 35.13 \\
& & & RG-64gf & & 70.74 & 73.59 & 78.4 & 91.66 & 95.5 & 73.94 & 90.45 & 79.46 & 88.03 & 74.39 & 96.75 & 38.10 & 83.17 & 93.36 & 71.35 & 89.08 & 79.76 & 47.21 & 97.48 & 94.71 & 73.32 & 74.2 & 98.91 & 78.36 & 39.93 \\
& & & RG-128gf & & 73.65 & 73.35 & 78.52 & 90.54 & 95.82 & 73.03 & 91.07 & 79.11 & 88.40 & 76.79 & 95.06 & 39.73 & 83.6 & 93.06 & 70.24 & 89.81 & 80.96 & 45.89 & 97.28 & 94.95 & 74.16 & 71.32 & 99.03 & 76.64 & 40.15 \\
& & & DeiT-B/16 & & 53.37 & 54.25 & 82.13 & 90.29 & 96.6 & 72.08 & 92.48 & 69.43 & 83.28 & 33.33 & 24.28 & 34.11 & 84.73 & 93.41 & 67.28 & 90.83 & 81.6 & 43.62 & 98.14 & 93.56 & 73.78 & 69.05 & 97.99 & 77.56 & 42.10 \\o
& & & ViT-B/16 INet-22k & & 54.76 & 53.84 & 89.9 & 93.04 & 96.64 & 78.40 & 93.53 & 74.95 & 86.38 & 33.19 & 24.51 & 33.8 & 86.03 & 93.98 & 75.68 & 93.65 & 88.04 & 52.52 & 99.3 & 99.68 & 80.22 & 70.99 & 98.21 & 85.79 & 49.17 \\
& & & EN-B7~\cite{radford2021learning} & & 51.9 & -- & 84.5 & 94.7 & 96.3 & 76.8 & 86.8 & -- & 80.8 & -- & -- & -- & 85.2 & 95.2 & 77.1 & 94.9 & 80.1 & 72.3 & 99.1 & 95.9 & 69.0 & 75.8 & 98.6 & 81.9 & 56.8 \\
& & & EN-B7-Noisy~\cite{radford2021learning} & & 50.5 & -- & 88.4 & 95.5 & 96.2 & 80.5 & 88.5 & -- & 73.4 & -- & -- & -- & 83.8 & 95.5 & 72.2 & 96.0 & 82.0 & 71.2 & 99.4 & 96.6 & 72.6 & 73.0 & 98.5 & 86.6 & 63.2 \\
& & & EN-B8~\cite{radford2021learning} & & 51.4 & -- & 84.5 & 95.2 & 97.0 & 77.1 & 87.4 & -- & 80.4 & -- & -- & -- & 85.2 & 94.9 & 76.8 & 95.0 & 80.7 & 71.5 & 99.2 & 96.3 & 69.6 & 70.9 & 98.6 & 82.4 & 57.7 \\
\hline

\multirow{8}{*}{\rotatebox{90}{CLIP~\cite{radford2021learning}}}
& & & RN50 & & 53.6 & -- & 86.4 & 89.6 & 95.2 & 76.4 & 87.5 & -- & 82.4 & -- & -- & -- & 82.7 & 88.2 & 78.3 & 88.7 & 70.3 & 49.1 & 96.6 & 96.1 & 73.3 & 70.2 & 98.3 & 81.6 & 57.2 \\
& & & RN50x4 & & 52.5 & -- & 91.3 & 92.5 & 96.4 & 79.5 & 89.7 & -- & 85.5 & -- & -- & -- & 83.0 & 91.9 & 85.9 & 90.5 & 73.0 & 57.3 & 97.8 & 97.8 & 77.0 & 59.4 & 98.5 & 85.7 & 62.6 \\
& & & RN50x16 & & 53.8 & -- & 93.3 & 93.7 & 97.0 & 79.1 & 91.4 & -- & 89.0 & -- & -- & -- & 83.5 & 93.5 & 88.7 & 92.2 & 74.9 & 62.7 & 98.6 &  98.3  & 79.2 & 69.2 & 98.9 & 88.0 & 66.3 \\
& & & RN50x64 & & 55.0 & -- & 94.8 & 95.4 & 97.1 & 82.0 & 92.8 & -- & 90.2 & -- & -- & -- & 83.7 & 94.5 & 90.5 & 94.1 & 78.6 & 67.7 & 99.1 & 98.9 & 81.1 & 69.2 & 98.9 & 89.5 & 69.1 \\
& & & LM-RN50 & & 51.2 & -- & 81.3 & 85.5 & 93.4 & 71.5 & 84.0 & -- & 73.8 & -- & -- & -- & 82.9 & 82.8 & 74.9 & 82.8 & 61.7 & 44.9 & 95.3 &  91.1  & 69.6 & 70.2 & 96.6 & 76.4 & 51.9 \\
& & & ViT-B/16 & & 57.1 & -- & 92.8 & 94.7 & 97.1 & 79.2 & 92.7 & -- & 86.6 & -- & -- & -- & 83.5 & 93.1 & 86.7 & 96.2 & 83.1 & 59.5 & 99.0 & 98.1 & 78.4 & 67.8 & 99.0 & 88.4 & 66.1 \\
& & & ViT-L/14 & & 57.8 & -- & 95.2 & 96.5 & 98.2 & 82.1 & 94.1 & -- & 92.5 & -- & -- & -- & 85.8 & 95.1 & 90.9 & 98.0 & 87.5 & 69.4 & 99.7 & 99.2 & 81.8 & 64.7 & 99.2 & 91.5 & 72.0 \\
& & & ViT-L/14-336px & & 60.3 & -- & 95.9 & 96.0 & 98.1 & 83.0 & 94.9 & -- & 92.4 & -- & -- & -- & 85.6 & 95.1 & 91.5 & 97.9 & 87.4 & 71.6 & 99.7 & 99.2 & 82.2 & 69.2 & 99.2 & 92.0 & 73.0 \\
\hline

\end{tabular}
}
\caption{
    \textbf{Linear probe results} for all models on 25 different datasets. All models are evaluated by us unless otherwise indicated in which case, results are from Table 10 of ~\cite{radford2021learning}.
  }
  \label{tab:appendix_all_linear_detailed_results}
\end{table*}

\begin{table}[t]
  \centering
  \setlength{\tabcolsep}{0.3em}\scalebox{0.98}{
  \begin{tabular}{@{}ll c cc@{}}
    \toprule
    & & & \multicolumn{2}{c}{ImageNet-1K} \\
    \cmidrule{4-5}
    Method  & Arch. & Param. & 1\% & 10\%  \\

    \midrule
    \multicolumn{5}{@{}l}{\textit{Semi-supervised methods trained on full ImageNet}}\\
    FixMatch~\cite{sohn2020fixmatch} & RN50 & \phantom{0}24M & - & 71.5  \\
    CowMix~\cite{french2020milking}  & RN152 & 265M & - & 73.9  \\
    
    \midrule
    \multicolumn{5}{@{}l}{\textit{Self-supervised pretraining on full ImageNet}}\\
    SimCLR~\cite{chen2020simple}      & RN50   & \phantom{0}24M & 48.3 & 65.6  \\
    SwAV~\cite{caron2020unsupervised} & RN50   & \phantom{0}24M & 53.9 & 70.2  \\
    BYOL~\cite{grill2020bootstrap}    & RN200  & 250M   & 71.2 & 77.7  \\ 
    SimCLR v2~\cite{chen2020big}      & RN152w3+SK & 795M & 74.9  & 80.1  \\
    
    \midrule
     \multicolumn{5}{@{}l}{\textit{Pretrained on random internet images}}\\\
     SEER & RG128 & 693M & 57.5 & 76.7  \\
     SEER & RG256 & 1.5B & 60.5 & 77.9  \\
     SEER & RG10B & 10B & 62.4 & 78.8  \\
    \bottomrule
  \end{tabular}
  }
\vspace{.3em}
\caption{
    \textbf{Low-shot learning on ImageNet and Places205}
We compare our approach with semi-supervised approaches and self-supervised pretraining on low-shot learning. Our model is finetuned on either 1\% or 10\% of ImageNet, and \textit{does not access the rest of ImageNet images}. As opposed to our method, the other methods use all the images from ImageNet during pretraining or finetuning.
  }
  \label{tab:semi_sup_inet}
\end{table}

\subsection{Low-Shot Transfer}
We also evaluate the performance of our 10Billion parameters SEER model in the low-shot setting, i.e. with a fraction of data (1\% and 10\% ImageNet) on the downstream task similar to previous studies~\cite{goyal2021self,caron2020unsupervised}. The results are in Table~\ref{tab:semi_sup_inet}. We observe the performance consistently improves on these both tasks as the model size increases. Especially on the 1\% fraction, the improvement is the the most +2 pts.

\section{Copy Detection Nuances}
\label{sec:appendix_copy_detection_nuances}

For the CopyDays benchmark, we follow a typical retrieval pipeline. This consists of extracting features from the train, database, and query datasets, post-processing these features, training PCA on the train set, applying PCA on the query and database set, and finally ranking the database for each query. From this ranking we calculate and report the mean average precision (mAP).

This pipeline has many dials to turn. We ran experiments with the res4 and res5 layers, different post-processing methods including, Regional Maximum Activations of Convolutions (R-MAC)~\cite{tolias2016particular,caron2018deep}, Generalized-Mean (GeM) pooling~\cite{berman2019multigrain,caron2021emerging}, and Average and Max Pooling, which are special cases of GeM pooling. We also swept over different image sizes, PCA dimensions, and R-MAC spatial levels. Our best results typically used the res4 layer with R-MAC spatial level 3, while the best PCA dimension and image size depended on the model. When using R-MAC, we trained the PCA on the full max pool matrix of the different crops. When applying PCA on the database and query datasets, we apply PCA on the same full max pool matrix of the different crops before summing the crops and normalizing the output as in the R-MAC algorithm. All of the code used is available in the open source VISSL library \cite{goyal2021vissl}.

\begin{figure}[t]
  \centering
  \includegraphics[width=0.49\textwidth]{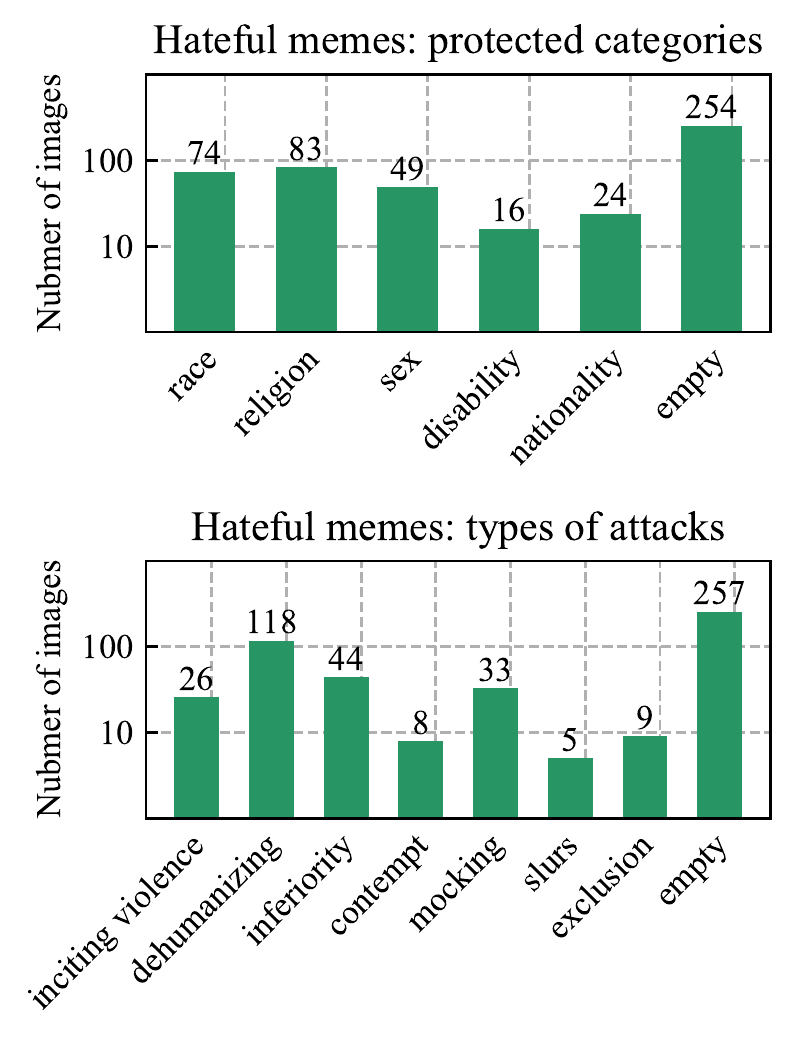}
  \caption{
    Data distribution of types of hate-speech in \textit{dev} set of \textbf{\hateful} dataset.
  }
  \label{fig:hateful_memes_distribution} \end{figure}

\section{Data distributions}

\begin{figure*}[t]
  \centering
  \includegraphics[width=0.45\textwidth]{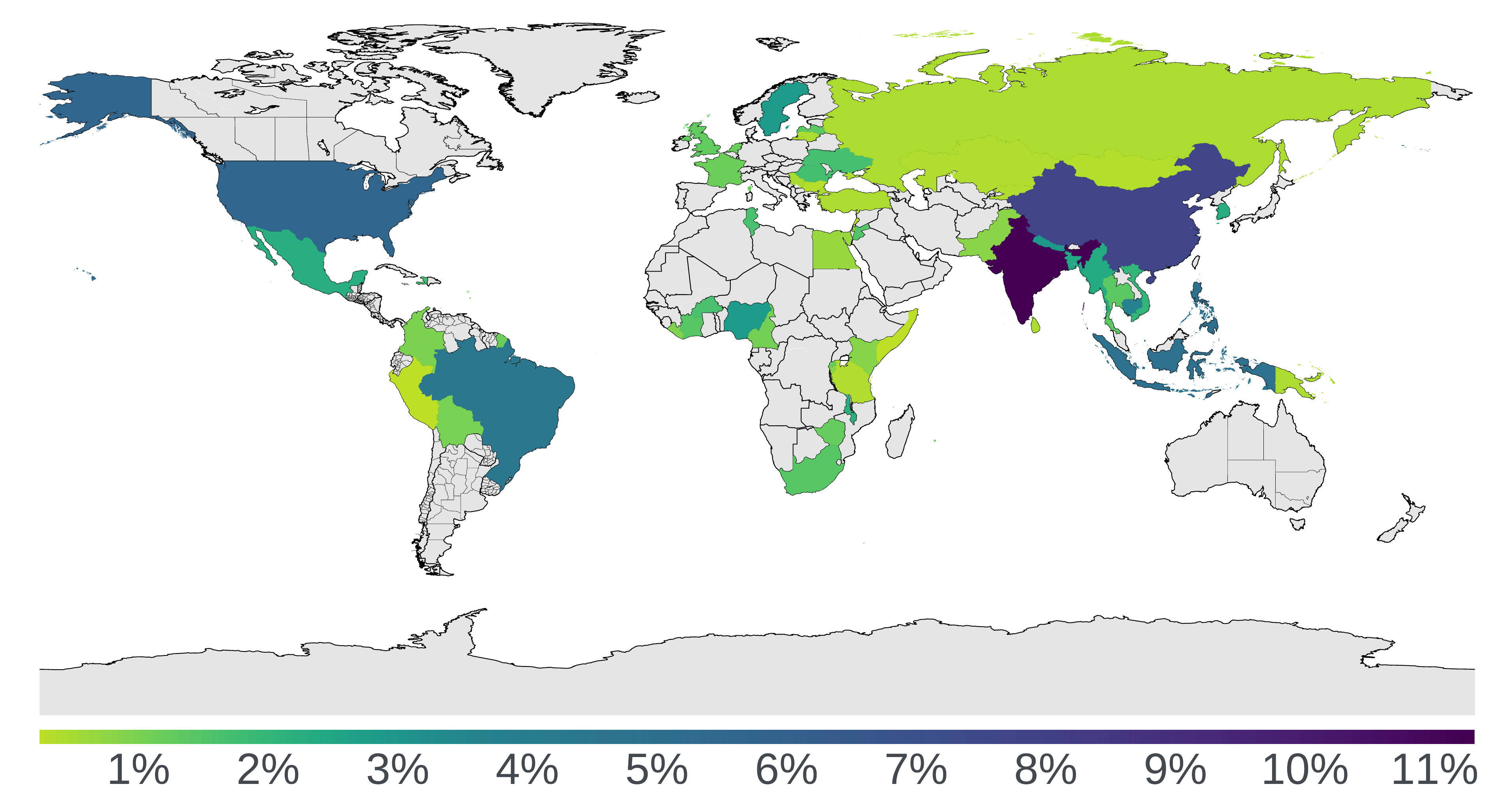}
  \hspace{0.5in}
  \includegraphics[width=0.45\textwidth]{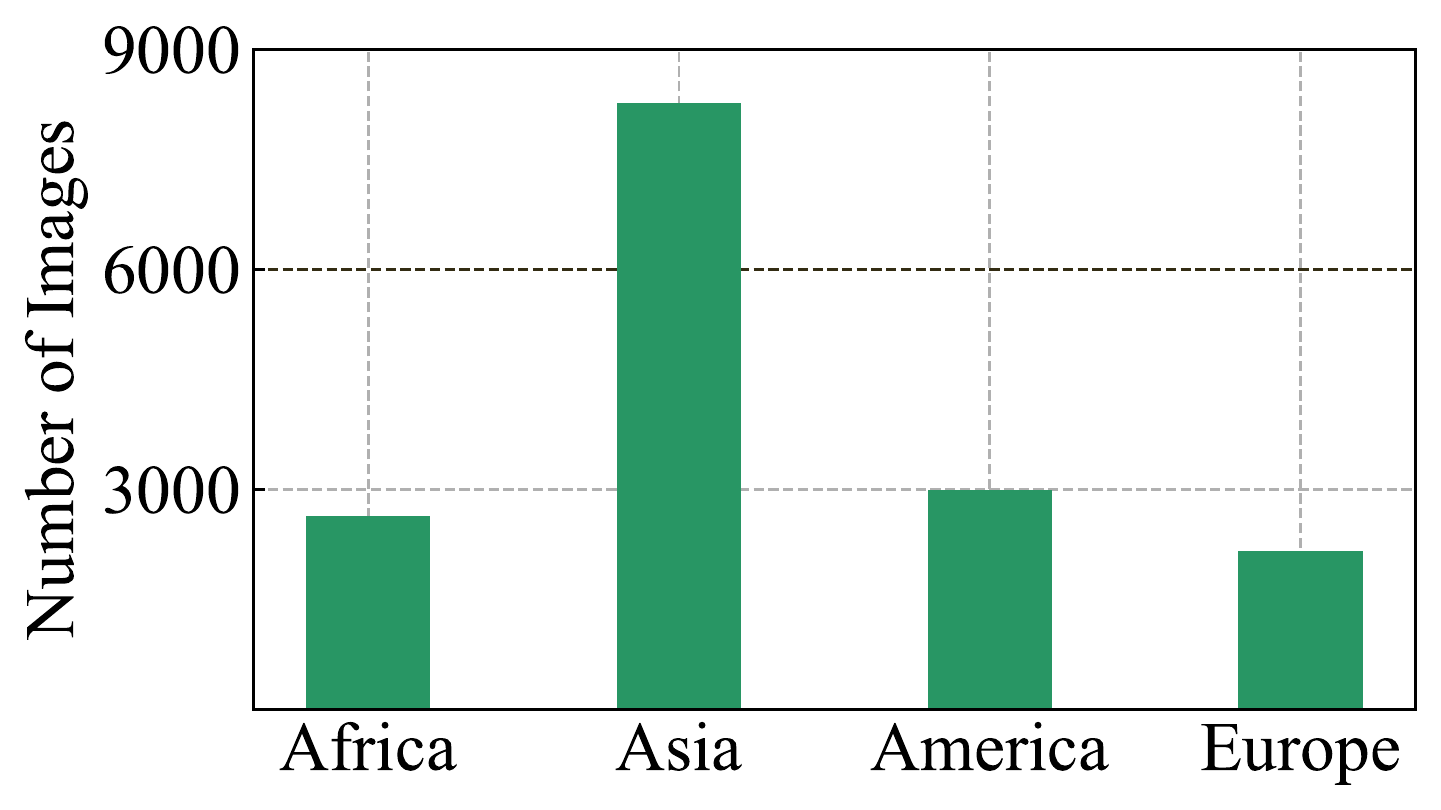}
  \caption{
    \textbf{Data Distribution of Dollar Street} dataset which features 94 concepts across 54 countries and 4 regions of the world.
  }
  \label{fig:dollar_street_data_distribution} \end{figure*}

\begin{figure*}[t]
  \centering
  \includegraphics[width=1\textwidth]{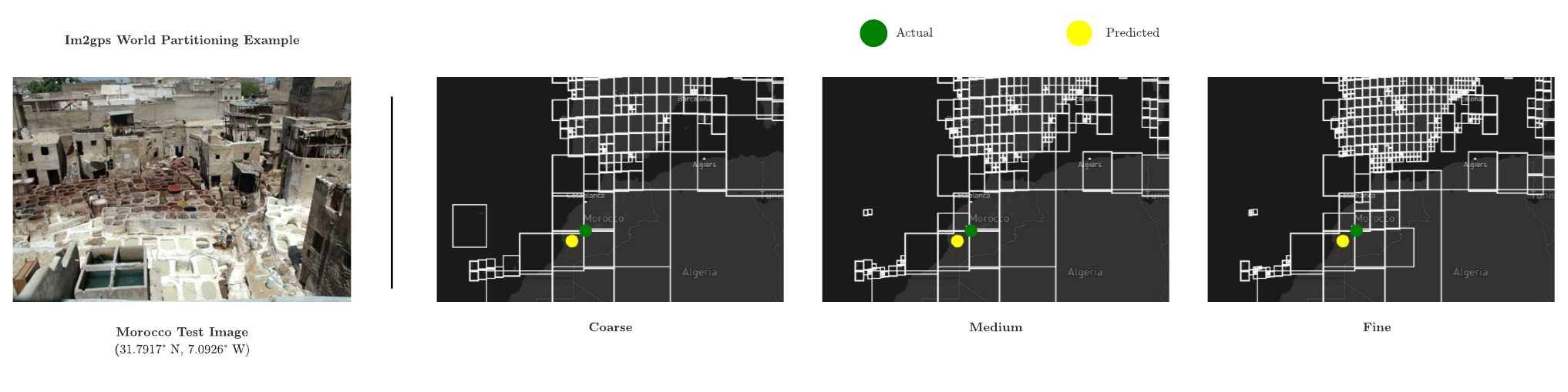}
  \caption{
    The im2gps evaluation requires finetuning a classification model on either coarse, medium, or fine partitionings of the world. The model outputs a probability distribution over these partitions, we predict the partition with the greatest probability, and choose the mean latitude and longitude of the predicted partition for our final prediction.
  }
  \label{fig:partitioning_figure}
\end{figure*}

For the fairness evaluations on DollarStreet and Hateful Memes challenge, we show the data distribution in Figure~\ref{fig:dollar_street_data_distribution} and Figure~\ref{fig:hateful_memes_distribution} respectively. Further, for studying the geolocalization salient property, we visually demonstrate the differences between various types of cell partitionings in Figure~\ref{fig:partitioning_figure}.
\end{document}